\documentclass[a4paper,11pt]{article}

\usepackage{fullpage}
\usepackage{courier}
\usepackage{authblk}
\usepackage{calc}

\usepackage[utf8]{inputenc} 
\usepackage[T1]{fontenc}    
\usepackage{lmodern}
\usepackage{hyperref}       
\usepackage{booktabs}       
\usepackage{amsfonts}       
\usepackage{nicefrac}       
\usepackage{microtype}      
\usepackage{dsfont}
\usepackage{algorithm}
\usepackage{paralist}
\usepackage[noend]{algorithmic}
\usepackage{caption,subcaption}
\usepackage{mathtools}
\usepackage{multirow}
\usepackage{enumitem}

\DeclarePairedDelimiter\floor{\lfloor}{\rfloor}

\usepackage{isa-definitions}

\newcommand{\X}{\ensuremath{\mathbf{X}}}

\newcommand{\SPN}{\mathcal{S}}
\newcommand{\cndbar}{\,|\,}

\newcommand{\Node}{\mathsf{N}}
\newcommand{\Leaf}{\mathsf{L}}
\newcommand{\SumNode}{\mathsf{S}}
\newcommand{\ProdNode}{\mathsf{P}}

\newcommand{\ch}{\mathsf{ch}}
\newcommand{\scope}{\mathsf{sc}}

\usepackage{xcolor}
\definecolor{wwine} {HTML} {cccaa9}
\definecolor{rwine} {HTML} {770a2d}

\usepackage{xcolor}

\usepackage{xcolor}
\definecolor{wwine} {HTML} {cccaa9}
\definecolor{rwine} {HTML} {770a2d}
\definecolor{gold2} {RGB} {255, 130, 0}
\definecolor{gold4} {RGB} {250, 100, 0}
\definecolor{gold6} {RGB} {245, 90, 0}
\definecolor{greentue} {RGB} {50, 152, 101}
\definecolor{lacamlilac} {RGB} {107,93,153}
\definecolor{lacamlilac2} {RGB} {93, 109, 152}
\definecolor{lacamlightlilac} {RGB} {174, 166, 201}
\definecolor{lacamdarklilac} {RGB} {51, 10, 102}
\definecolor{lacamdarklilac5} {RGB} {51, 10, 102}
\colorlet{lacamdarklilac4} {lacamdarklilac5!80!}
\colorlet{lacamdarklilac3} {lacamdarklilac5!60!}
\colorlet{lacamdarklilac2} {lacamdarklilac5!40!}
\colorlet{lacamdarklilac1} {lacamdarklilac5!20!}
\definecolor{pink3} {HTML} {F06292}
\definecolor{lacamgold5} {RGB} {255, 87, 0}
\colorlet{lacamgold4} {lacamgold5!80!}
\colorlet{lacamgold3} {lacamgold5!60!}
\colorlet{lacamgold2} {lacamgold5!40!}
\colorlet{lacamgold1} {lacamgold5!20!}
\definecolor{violet} {RGB} {119, 111, 178}
\definecolor{petroil2} {RGB} {36, 165, 175}
\definecolor{petroil4} {RGB} {30, 132, 149}
\definecolor{petroil6} {RGB} {23, 101, 115}
\definecolor{gold1} {RGB} {254, 176, 52}
\definecolor{gold2} {RGB} {255, 130, 0}
\definecolor{gold4} {RGB} {250, 100, 0}
\definecolor{gold6} {RGB} {245, 90, 0}
\definecolor{greentue} {RGB} {50, 152, 101}

\definecolor{lacamoil5}{rgb}{0.13, 0.67, 0.8}
\colorlet{lacamoil4} {lacamoil5!80!}
\colorlet{lacamoil3} {lacamoil5!60!}
\colorlet{lacamoil2} {lacamoil5!40!}

\definecolor{tomato0} {HTML} {EF9A9A}
\definecolor{tomato1} {HTML} {F44336}
\definecolor{tomato2} {HTML} {E53935}
\definecolor{tomato3} {HTML} {D32F2F}
\definecolor{tomato4} {HTML} {C62828}
\definecolor{tomato5} {HTML} {B71C1C}

\definecolor{peas1} {HTML} {009688}
\definecolor{peas2} {HTML} {00897B}
\definecolor{peas3} {HTML} {00796B}
\definecolor{peas4} {HTML} {00695C}
\definecolor{peas5} {HTML} {004D40}

\definecolor{bgrey0} {HTML} {78909C}
\definecolor{bgrey1} {HTML} {607D8B}
\definecolor{bgrey2} {HTML} {546E7A}
\definecolor{bgrey3} {HTML} {455A64}
\definecolor{bgrey4} {HTML} {37474F}
\definecolor{bgrey5} {HTML} {263238}

\definecolor{olive0} {HTML} {C5E1A5}
\definecolor{olive1} {HTML} {AED581}
\definecolor{olive2} {HTML} {9CCC65}
\definecolor{olive3} {HTML} {8BC34A}
\definecolor{olive4} {HTML} {7CB342}
\definecolor{olive5} {HTML} {689F38}

\definecolor{pink0} {HTML} {FCE4EC}
\definecolor{pink1} {HTML} {F8BBD0}
\definecolor{pink2} {HTML} {F48FB1}
\definecolor{pink3} {HTML} {F06292}
\definecolor{pink4} {HTML} {EC407A}
\definecolor{pink5} {HTML} {FF80AB}

\definecolor{brown0} {HTML} {D7CCC8}
\definecolor{brown1} {HTML} {BCAAA4}
\definecolor{brown2} {HTML} {A1887F}
\definecolor{brown3} {HTML} {8D6E63}
\definecolor{brown4} {HTML} {795548}
\definecolor{brown5} {HTML} {6D4C41}
\definecolor{brown6} {HTML} {5D4037}

\definecolor{yellow0} {HTML} {CDDC39}
\definecolor{yellow1} {HTML} {9E9D24}
\definecolor{yellow2} {HTML} {74741b}
\definecolor{yellow3} {HTML} {FFBD2A}
\definecolor{yellow4} {HTML} {FFB000}
\definecolor{yellow5} {HTML} {FFD600}
\definecolor{yellow6} {HTML} {D6C67B}

\definecolor{wwine} {HTML} {cccaa9}
\definecolor{rwine} {HTML} {770a2d}
\definecolor{abagreen} {HTML} {2ca02c}
\definecolor{abapink} {HTML} {e377c2}
\definecolor{abagray} {HTML} {7f7f7f}
\definecolor{abayellow} {HTML} {bcbd22}
\definecolor{ababrown} {HTML} {8c564b}

\colorlet{lacamoil1} {lacamoil5!20!}

\begin{document}

\title{Automatic Bayesian Density Analysis}

\author{Antonio Vergari$^{1}$}
\author{Alejandro Molina$^{2}$}
\author{Robert Peharz$^{3}$}
\author{\\\vspace{-10pt}Zoubin Ghahramani$^{3,4}$}
\author{Kristian Kersting$^{2}$}
\author{Isabel Valera$^{1}$\vspace{5pt}} 

\affil{%
  {$^{1}$Max-Planck-Institute for Intelligent Systems, Germany \hfill\texttt{first.last@tue.mpg.de}} \\
  {$^{2}$TU Darmstadt, Germany \hfill\texttt{last@cs.tu-darmstadt.de}}\\
  {$^{3}$University of Cambridge, UK \hfill\texttt{\{rp587,zoubin\}@cam.ac.uk}}\\
  {\hspace{-347pt}$^{4}$Uber AI Labs, USA\hfill}\\
}

\maketitle
\begin{abstract}
Making sense of a dataset in an automatic and unsupervised fashion is a challenging problem in statistics and AI.
Classical approaches for {exploratory data analysis} are usually not flexible enough to deal with the uncertainty inherent to real-world data: they are often restricted to fixed latent interaction models and homogeneous likelihoods;
they are sensitive to missing, corrupt and anomalous data; 
moreover, their expressiveness generally comes at the price of intractable inference.
As a result, supervision from statisticians is usually needed to find the right model for the data. 
However,
since domain experts are not necessarily also experts in statistics, we propose Automatic Bayesian Density Analysis (ABDA) to make exploratory data analysis accessible at large. 
Specifically, ABDA allows for automatic and efficient missing value estimation, statistical data type and likelihood discovery, anomaly detection and dependency structure mining, on top of providing accurate density estimation.
Extensive empirical evidence shows that ABDA is a suitable tool for automatic exploratory analysis of mixed continuous and discrete tabular data.
\end{abstract}

\frenchspacing
\section{Introduction}

``Making sense'' of a dataset---a task often referred to as \textit{data understanding} or \textit{exploratory data analysis}---is a fundamental step that precedes and guides a classical machine learning (ML) pipeline.
Without domain experts' background knowledge, a dataset might remain nothing but a list of numbers and arbitrary symbols.
On the other hand, without statisticians' supervision, processing the data and extracting useful models from it might go beyond the ability of domain experts who might not be experts in ML or statistics.
Therefore, in times of abundant data, but an insufficient number of statisticians, 
methods which can ``understand'' and ``make sense'' of a dataset \textit{with minimal or no supervision} are 
in high demand.

The idea of machine-assisted data analysis has been pioneered by The Automatic Statistician project~\cite{duvenaudLGTG13,lloydDGTG14} which proposed to automate model selection for regression 
and classification
tasks via compositional kernel search.
Analogously, but with a clear focus on performance optimization, AutoML frameworks~\cite{Guyon2016} automate the choice of supervised ML models for a task-dependent loss. 
In contrast, we address model selection for a \textit{fully unsupervised task}, with the aim of assisting domain experts in exploratory data analysis, 
providing them with a probabilistic framework to perform efficient inference and gain useful insights from the data in an automatic way. 

In principle, a suitable unsupervised learning approach \textit{to find out what is in the data} is \emph{density estimation} (DE).
In order to
perform DE on data in a tabular format, a practitioner would first try to heuristically infer the \textit{statistical types} of the data features, e.g., real, positive, numerical and nominal data types~\cite{Valera2017b,Valera2017a}.
Based on this, she would need to make assumptions about their distribution, i.e., selecting a \textit{parametric likelihood model} suitable for each marginal, e.g., Gaussian for real, Gamma for positive, Poisson for numerical and Categorical for nominal data.
Then, she could start investigating the global interactions among them, i.e., determining the \textit{statistical dependencies} across features.
In this process, she would also likely need to deal with \textit{missing values} and reason whether the data may be corrupted or contain \textit{anomalies}.

\begin{figure*}[!t]
\begin{subfigure}[b]{0.17\columnwidth}
    \includegraphics[width=1\textwidth]{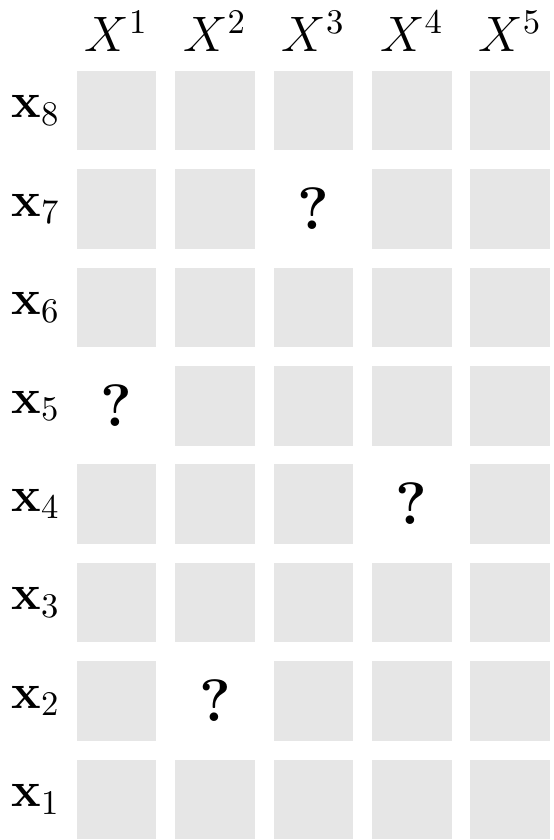}
    \caption{}
    \label{fig:abda-data}
  \end{subfigure}\hspace{10pt}
  \begin{subfigure}[b]{0.17\columnwidth}
    \includegraphics[width=1\textwidth]{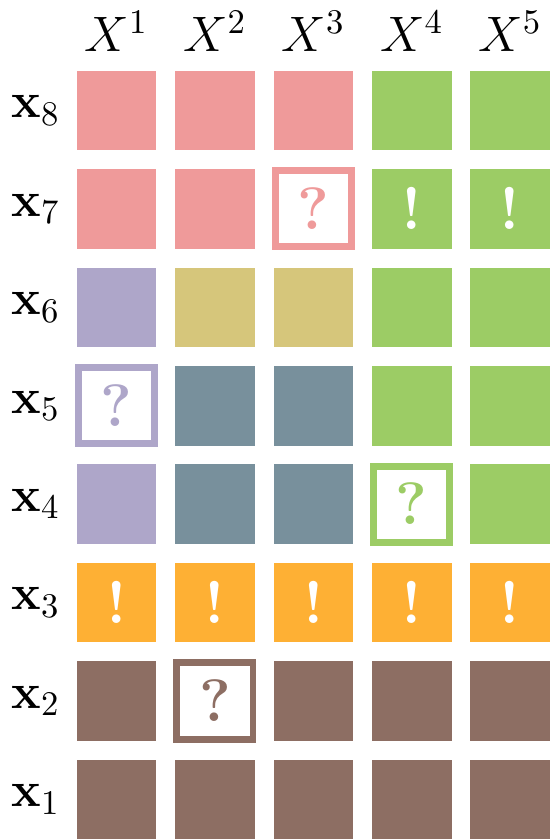}
    \caption{}
    \label{fig:abda-grid}
  \end{subfigure}\hspace{10pt}
  \begin{subfigure}[b]{0.25\columnwidth}
    \vspace{-5pt}
    \includegraphics[width=.8\textwidth]{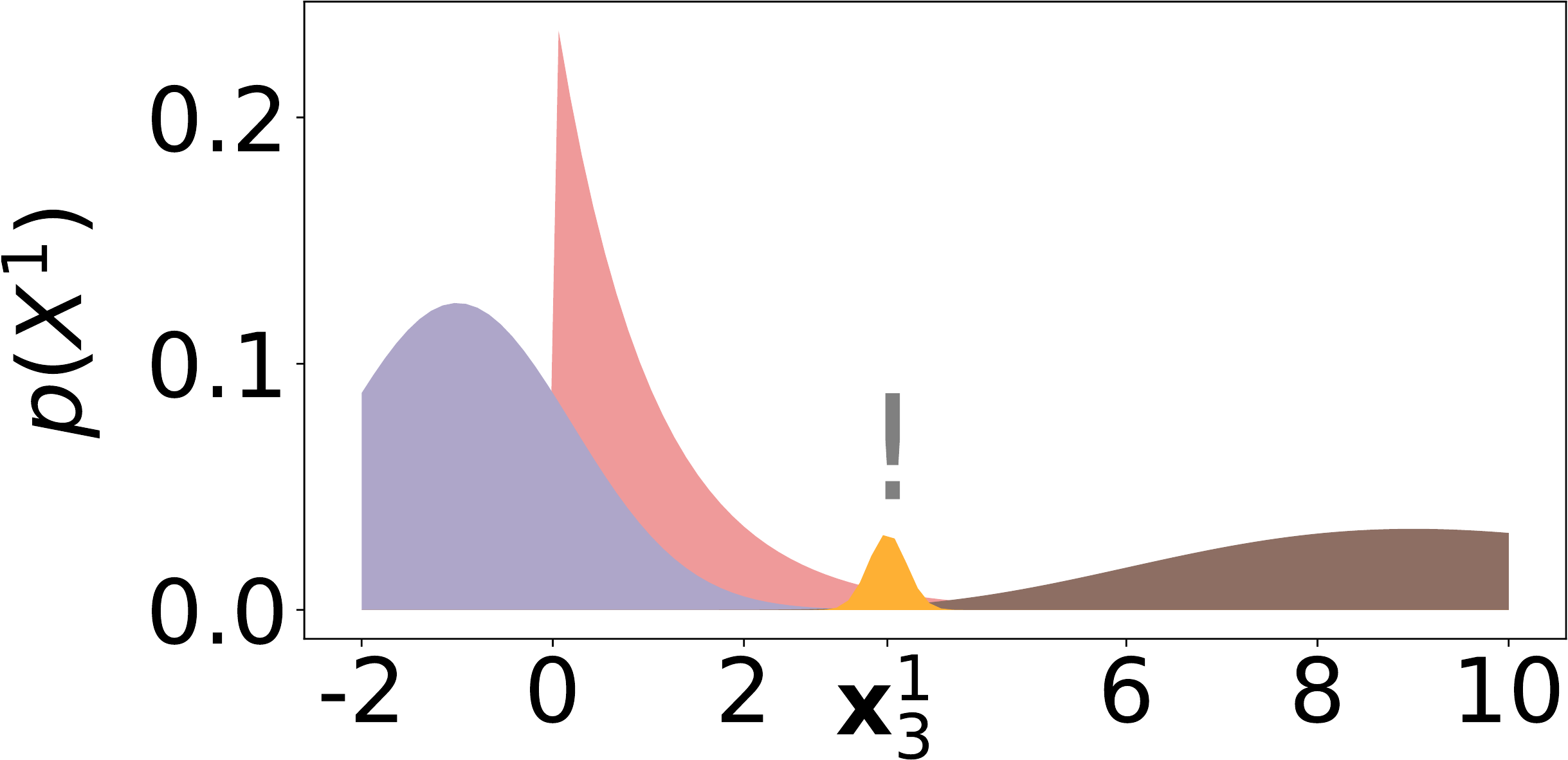}\\[10pt]
    \includegraphics[width=.8\textwidth]{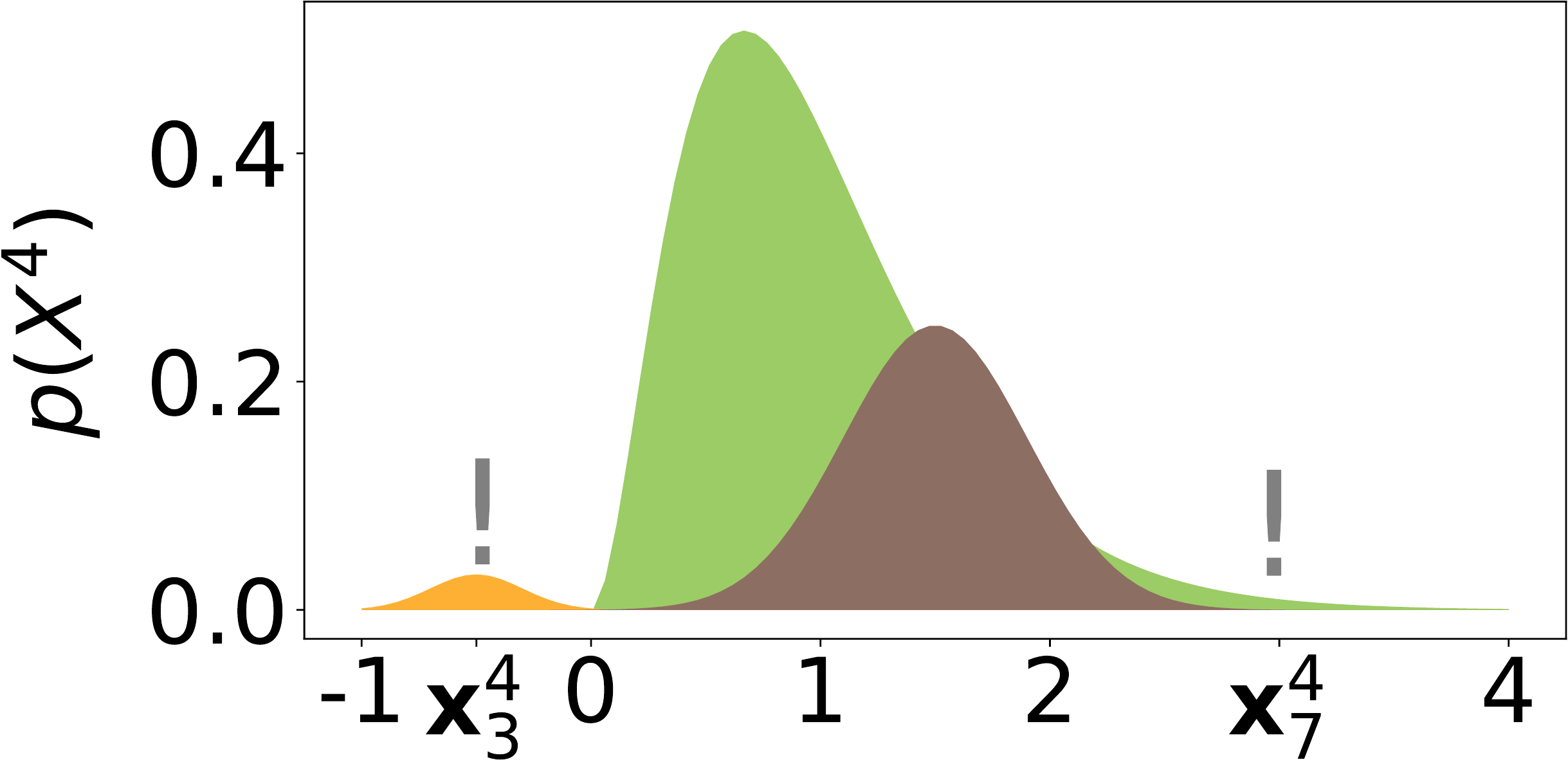}
    \caption{}
    \label{fig:abda-de}
  \end{subfigure}\hspace{-25pt}
  \begin{subfigure}[b]{0.25\columnwidth}
    \vspace{10pt}
    \includegraphics[width=.8\textwidth]{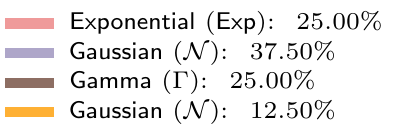}\\[22pt]
    \includegraphics[width=.8\textwidth]{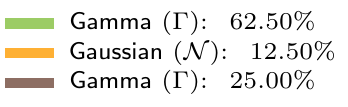}\\[10pt]
    \caption{}
    \label{fig:abda-lls}
  \end{subfigure}\hspace{-15pt}
  \begin{subfigure}[b]{0.10\columnwidth}
  \tiny
    \begin{align*}
    \mathcal{P}_{1}: 5.9\leq &\ {\color{brown3}X^{1}}< 10.0\ \wedge\\
    1.0\leq &\ {\color{brown3}X^{4}}< 2.01\\
    (\mathsf{supp}(&\mathcal{P}_{1})=0.25)\\
    \mathcal{P}_{2}: (0.0\leq &\ {\color{tomato0}X^{1}}< 3.5\ \vee\\ 
    -2.0\leq &\ {\color{lacamlightlilac}X^{1}}< 2.0)\ \wedge\\
    0.1\leq &\ {\color{olive2}X^{4}}< 3.0\\
    (\mathsf{supp}(&\mathcal{P}_{2})=0.49)\\[-1pt]  
\end{align*}
    \caption{}
    \label{fig:abda-pm}
  \end{subfigure}
    \caption{Automatic exploratory analysis with ABDA on a tabular dataset comprising samples from mixed continuous and discrete features $\mathbf{X}=\{X^{1},\ldots,X^{5}\}$, potentially containing missing values (denoted as ``$?$'') (\textbf{\ref{fig:abda-data}}). 
    The latent dependency structure inferred by ABDA induces a \textit{hierarchical partition} over samples and features (i.e., a hierarchical co-clustering) given a learned Sum-Product Network (SPN) structure (\textbf{\ref{fig:abda-grid}}).
    Statistical data types and likelihood models are discovered by estimating each feature distribution as a mixture model over a dictionary of suitable and interpretable parametric model, e.g., Gaussian for real data, Gamma, Exponential distributions for positive data in (\textbf{\ref{fig:abda-de}}-\textbf{\ref{fig:abda-lls}}).
    ABDA can efficiently impute missing entries as the most probable values \ref{fig:abda-grid} by SPN inference routines.
    Anomalous entries (denoted as ``$!$'' in \textbf{\ref{fig:abda-grid}}-\textbf{\ref{fig:abda-de}}), on the other hand, are presented to the user as low-likelihood samples that are  relegated 
    to micro-clusters (e.g., $\mathbf{x}_{3}$) or to the distribution tails (e.g., $\mathbf{x}_{7}^{4}$ and $\mathbf{x}_{7}^{5}$).
    Moreover, ABDA allows to automatically discover complex dependency patterns, e.g. conjunctions of confidence intervals, by exploiting the correlations over the induced hierarchical co-clustering (\textbf{\ref{fig:abda-pm}}).
    }
    \label{fig:abda-full}
\end{figure*}

Unfortunately, classical approaches to DE, even if ubiquitous in ML applications, are still far from delivering automatic tools suitable for performing \textit{all} these steps on \textit{real-world data}.
First, general-purpose density estimators usually assume the statistical data types and the likelihood models to be \textit{known} a priori and \emph{homogeneous} across random variables (RVs) or mixture model components~\cite{Valera2017a}.
Indeed, the standard approach is still to generally treat \textit{all} continuous data as (mixtures of) Gaussian RVs and discrete data as categorical variables.
Second, they either assume ``shallow'' dependency structures that might be too simplistic to capture real-world statistical dependencies or use ``deeper'' latent structures which cannot be easily learned.
As a result, they generally lack enough flexibility to deal with fine grain statistical dependencies, and to be robust to corrupt data and outliers, especially when a maximum likelihood learning approach is used.

A Bayesian treatment of DE, on the other hand, often struggles to scale up to high-dimensional data and approximate inference routines are needed~\cite{Ghahramani2000}.
CrossCat~\cite{Mansinghka2016} shows a clear example of this trade-off.
Even though CrossCat models sub-populations in the data with context-specific latent variables, 
this latent structure is limited to a two-layer hierarchy, and even so inference routines have to be approximated for efficiency.
Moreover, it is still limited to fixed and homogeneous statistical data types and therefore, likelihood models.

The latent variable matrix factorization model (ISLV) introduced in \cite{Valera2017a} is the first attempt to overcome this limitation by \textit{modeling uncertainty over the statistical data types} of the features.
However, like CrossCat, it can only perform inference natively in the transductive case, i.e., to data available during training.
While ISLV allows one to infer the data type of a feature, this approach still uses a single \emph{ad-hoc} likelihood function for each data type.

Recently, Mixed Sum-Product Networks (MSPNs) \cite{Molina2017b} have been proposed as deep models for heterogeneous data able to perform tractable inference also in the inductive scenario, i.e., on completely unobserved test data.
Indeed, MSPNs can exploit {context specific independencies} to learn latent variable hierarchies that are deeper than CrossCat.
However, MSPNs assume piecewise-linear approximations as likelihood models for both continuous and discrete RVs, which are not as interpretable as parametric distributions and also require continuous RVs to undergo a delicate discretization process.
As a result, MSPNs are highly prone to overfitting, and 
learning them via maximum likelihood results into a lack of robustness.

In this paper, we leverage the above models' advantages while addressing their shortcomings by proposing \emph{Automatic Bayesian Density Analysis} (ABDA).
Specifically, ABDA relies on sum-product networks (SPNs) to capture statistical dependencies in the data at different granularity through a hierarchical co-clustering. 
This rich latent structure is learned in an adaptive way, which automates the selection of adequate likelihood models for each data partition, and thus extends ISLV uncertainty modeling over statistical types.
As a result, ABDA goes beyond standard density estimation approaches, qualifying as the first approach to \textit{fully automate exploratory analysis for heterogeneous tabular data at large}.

As illustrated in Fig.~\ref{fig:abda-full}, ABDA allows for:  
\begin{description}[noitemsep,topsep=0pt,parsep=0pt,partopsep=0pt]
\item[\textbf{i)}]  inference 
for \emph{both} the statistical data types \emph{and} (parametric) likelihood models;
\item[\textbf{ii)}] robust estimation of missing values;  
\item[\textbf{iii)}] detection of corrupt or anomalous data; 
\item[\textbf{iv)}]  automatic discovery of the statistical dependencies and local correlations in the data.
\end{description}

ABDA relies on Bayesian inference through Gibbs sampling, allowing us to robustly measure uncertainties at performing all the above tasks.
In our extensive experimental evaluation, we demonstrate that ABDA effectively assists domain experts in both transductive and inductive settings.
Supplementary material and a reference implementation of ABDA are available at \url{github.com/probabilistic-learning/abda}.

\section{Sum-Product Networks (SPNs)}
\label{section:spns}

As SPNs provide the hierarchical latent backbone of ABDA, we will now briefly review them.
Please refer to~\cite{Peharz2017} for more details.

In the following, upper-case letters, e.g. $X$, denote RVs and lower-case letters their values, i.e.,~$x\sim X$. Similarly, we denote sets of RVs as~$\Xb$, and their combined values as $\mathbf{x}$.

\paragraph{Representation.} An SPN $\SPN$ over a random vector $\X= \{X^1, \ldots, X^D \}$ is a probabilistic model defined via a directed acyclic graph.
Each leaf node $\Leaf$ represents a probability distribution function over a single RV $X \in \X$, also called its \emph{scope}. 
Inner nodes represent either \emph{weighted sums} ($\SumNode$) or \emph{products} ($\ProdNode$).
For inner nodes, the scope is defined as the union of the scopes of its children.
The set of children of a node $\Node$ is denoted by $\ch(\Node)$.

A sum node $\SumNode$ encodes a mixture model $\SPN_{\SumNode}(\mathbf{x}) = \sum_{\Node \in \ch(\SumNode)} \omega_{\SumNode, \Node} \SPN_{\Node}(\mathbf{x})$ over sub-SPNs rooted at its children $\ch(\SumNode)$. 
We require that the all children of a sum node share the same variable scope---this condition is referred to as \textit{completeness} \cite{Poon2011}. 
The weights of a sum $\SumNode$ are drawn from the standard simplex (i.e., $\omega_{\SumNode,\Node} \geq 0$, $\sum_{\Node\in \mathsf{ch}(\SumNode)} \omega_{\SumNode,\Node} = 1$) and denoted as $\Omega^{\mathsf{S}}$. 
A product node $\ProdNode$ defines a factorization $\SPN_{\ProdNode}(\mathbf{x}) = \prod_{\Node \in \ch(\ProdNode)} \SPN_{\Node}(\mathbf{x})$ over its children distributions defined over disjoint scopes---this condition is referred to as \textit{decomposability} \cite{Poon2011}.
The parameters of $\SPN$ are the set of sum weights $\boldsymbol\Omega=\{\Omega^{\SumNode}\}_{\SumNode\in\SPN}$ and the set of all leaf distribution parameters $\{\boldsymbol H_{\Leaf}\}_{\Leaf\in\SPN}$. 

Complete and decomposable SPNs are highly expressive deep models and have been successfully employed in several ML domains~\cite{Peharz2015a,Molina2017a,Pronobis2017,Vergari2018} to capture highly complex dependencies in real-world data.
Moreover, they also allow to \textit{exactly} evaluate complete evidence, marginal and conditional probabilities in \textit{linear time} w.r.t.~their representation size~\cite{Darwiche2003}.

\paragraph{Latent variable representation.} Since each sum node defines a mixture over its children, one may associate a categorical latent variable (LV) 
$Z^{\mathsf{S}}$ to each sum $\mathsf{S}$ indicating a component.
This results in a \textit{hierarchical} model over the set of all LVs $\mathbf{Z}=\{Z^{\mathsf{S}}\}_{\mathsf{S}\in\SPN}$.
Specifically, an assignment to $\mathbf{Z}$ selects an \emph{induced
  tree} in $\SPN$~\cite{Zhao2016b,Peharz2017,Vergari2017}, i.e., a
tree path $\mathcal{T}$ starting from the root and comprising exactly
one child for each visited sum node and all child branches for each
visited product node (in green in Fig.~\ref{fig:abda-model:spn}).

It follows from completeness and decomposability that $\mathcal{T}$ selects a subset of $D$ leaves, in a one-to-one correspondence with $\X$. 
When conditioned on a particular $\mathcal{T}$, the SPN distribution factorizes as $\prod_d \Leaf^d_{j^d}$, where $\Leaf^d_k$ is the $k^\text{th}$ leaf for the $d^\text{th}$ RV, and $\mathbf{j}=\{j^{1},\dots,j^{D}\}$ are the indices of the leaves selected by $\mathcal{T}$.
The overall SPN distribution can be written as a mixture of such factorized distributions, running over all possible induced trees \cite{Zhao2016b}.

We will use this hierarchical LV structure to develop an efficient Gibbs sampling scheme to perform Bayesian inference for ABDA.

\paragraph{SPN learning.}

Existing SPN learning works focus on learning the SPN parameters given a structure \cite{Gens2012,Trapp2017,Zhao2016b} or jointly learn both the structure and the parameters \cite{Dennis2012,Dennis2015}. 
The latter approach has recently gained more attention since it automatically discovers the hierarchy over the LVs $\mathbf{Z}$ and admits greedy heuristic learning schemes leveraging the probabilistic semantics of nodes in an SPN~\cite{Gens2013,Rooshenas2014,Vergari2015}. 
These approaches recursively partition a data matrix in a top-down fashion, performing hierarchical co-clustering (partitioning over data samples and features).
As the base step, they learn a univariate likelihood model for single features.
Otherwise, they alternatively try to partition either the
RVs (or features) into independent groups, inducing
a product node; or the data samples (clustering), inducing a sum
node.
%

%
ABDA resorts to the SPN learning approach employed by MSPNs~\cite{Molina2017b}, as it is the only one able to build an SPN structure in a likelihood-agnostic way, therefore being suitable for our heterogeneous setting.
It performs a partitioning over mixed continuous and discrete data by exploiting a randomized approximation of the Hirschfeld-Gebelein-R\'{e}nyi Maximum Correlation Coefficient (RDC) \cite{LopezPaz2013}.
Note that, ABDA employs it to automatically select an \textit{initial}, global, LV structure, to be later provided to its Bayesian inference routines.

%

\section{Automatic Bayesian Density Analysis}
\label{section:abda}

For a general discussion, we extend $\Xb$ to a whole data matrix containing $N$ samples (rows) and $D$ features (columns).
RVs which are local to a row (column) receive now a sub-script $n$ (super-script $d$).

\begin{figure*}[!t]
  \centering
  \begin{subfigure}[b]{0.21\textwidth}
    \includegraphics[width=1\textwidth]{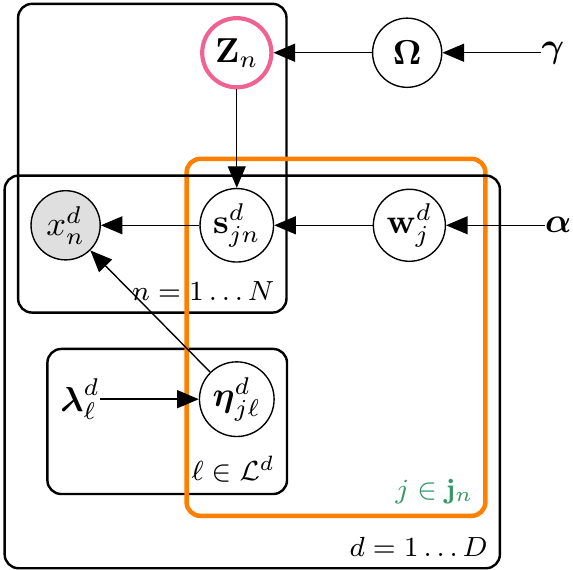} 
    \caption{Graphical model}
    \label{fig:abda-model:plate}
  \end{subfigure}\hspace{15pt}
  \begin{subfigure}[b]{0.18\textwidth}
    \includegraphics[width=1\textwidth]{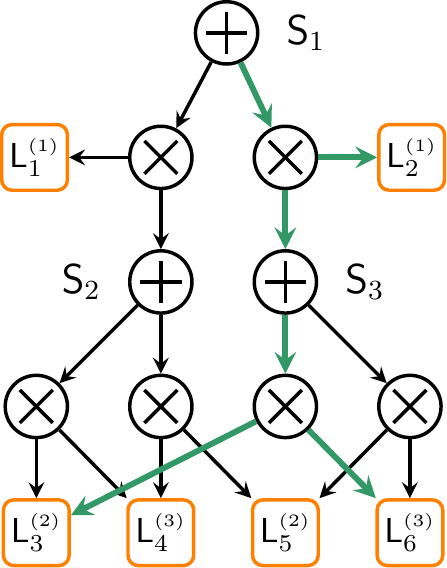}
    \caption{SPN}
    \label{fig:abda-model:spn}
  \end{subfigure}\hspace{20pt}
  \begin{subfigure}[b]{0.18\textwidth}
    \includegraphics[width=1\textwidth]{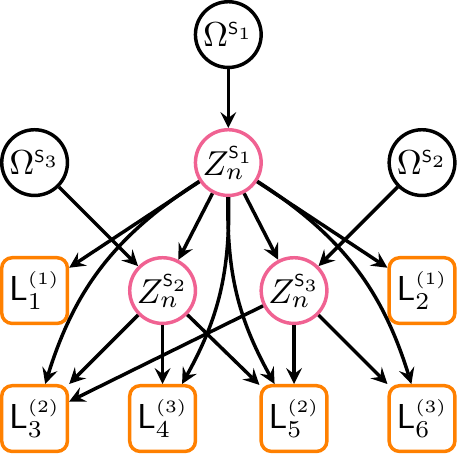}
    \caption{LV interpretation}
    \label{fig:abda-model:spn-lv}
  \end{subfigure}\hspace{20pt}
  \begin{subfigure}[b]{0.25\textwidth}
    \includegraphics[width=1\textwidth]{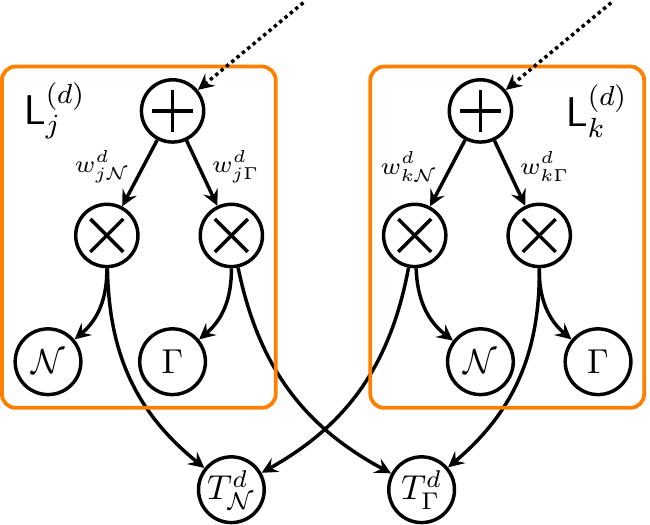}
    \caption{Type-augmented SPN}
    \label{fig:abda-model:global_types}
  \end{subfigure}
  \caption{\textbf{(\subref{fig:abda-model:plate})} Plate representation of ABDA, showing the \textit{global} LV (pink) and \textit{local} (orange) levels. \textbf{(\subref{fig:abda-model:spn})} The SPN representation with likelihood-modeling leaves $L_{j}^{(d)}$ over features $d$=1,2,3 selected by an induced tree (green), \textbf{(\subref{fig:abda-model:spn-lv})} and its corresponding hierarchy over LVs $\mathbf{Z}$. 
  \textbf{(\subref{fig:abda-model:global_types})} A global mixture model over Gaussian ($\mathcal{N}$) and Gamma ($\Gamma$) likelihoods interpreted as sub-SPNs sharing global auxiliary LVs $T^{d}$ across the local likelihood mixture models for feature $d$.}
  \label{fig:abda-model}
\end{figure*}

Our proposed Automatic Bayesian Density Analysis (ABDA) model can be thought as being organized in two levels: 
a \emph{global} level, capturing dependencies and correlations among the features, and a level which is \emph{local} w.r.t. each feature, consisting of dictionaries of likelihood models for that feature.
This model is illustrated Fig.~\ref{fig:abda-model}, via its graphical model representation, and corresponding SPN representations.
In this hierarchy, the global level captures context specific dependencies via recursive data partitioning, leveraging the LV structure of an SPN.
The local level represents context-specific uncertainties about the variable types, conditioned on the global context.

In contrast to classical works on DE, where typically fixed likelihood models are used as mixture components, e.g., see \cite{Poon2011,Vergari2015}), ABDA assumes a \emph{heterogeneous} mixture model combining several likelihood models from a user-provided \emph{likelihood dictionary}.
This dictionary may contain likelihood models for diverse types of discrete (e.g. Poisson, Geometric, Categorical,\ldots) and continuous (e.g., Gaussian, Gamma, Exponential,\ldots) data. 
It can be built in a generous automatic way, incorporating arbitrary rich collections of domain-agnostic likelihood functions.
Alternatively, its construction can be limited to a sensible subset of
likelihood models reflecting domain knowledge, e.g. a Gompertz and
Weibull distributions might be included by a user specifically dealing
with demographics data.

In what follows, assume that for each feature $d$ we have readily selected a dictionary $\{p^d_{\ell}\}_{\ell\in\mathcal{L}^{d}}$ of likelihood models, indexed by some set $\mathcal{L}^{d}$.
We next describe the process generating $\X$, also depicted in Fig.~\ref{fig:abda-model:plate}, in detail.

\paragraph{Generative model.}
The global level of ABDA contains latent vectors $\mathbf{Z}_n$ for each sample $\mathbf{x}_{n}$, associated with the SPN sum nodes (see previous section).
Each LV $Z^{\SumNode}_n$ in $\mathbf{Z}_n$ is drawn according to the sum-weights $\Omega^{\SumNode} \in \boldsymbol \Omega$, which are associated a Dirichlet prior, parameterized with hyper-parameters $\boldsymbol\gamma$: $$Z^\SumNode_n \sim \mathsf{Cat}(\Omega^{\mathsf{S}}),\quad\quad \Omega^{\mathsf{S}} \sim \mathsf{Dir}(\boldsymbol\gamma)$$.

As previously discussed, an assignment of $\mathbf{Z}_n$ determines an induced tree through the SPN, selecting a set of indices $\mathbf{j}_n = (j^1_n, \dots, j^D_n)$, such that the joint distribution of $\mathbf{x}_{n} = (x_n^1, \ldots, x_n^D)$ factorizes as 
\begin{equation}
    p(\mathbf{x}_{n} \cndbar \mathbf{j}_n, \boldsymbol\eta) 
    = \prod\nolimits_{d=1}^D \Leaf^d_{j^d_n}(x_n^d \cndbar \boldsymbol\eta^d_{j^d_n}),
\end{equation}
where $\Leaf^d_{j}$ is the $j^\text{th}$ leaf for feature $d$ and 
$\boldsymbol\eta^d_{j} = \{\boldsymbol\eta^d_{j,\ell}\}_{\ell\in\mathcal{L}^{d}}$
is the set of parameters belonging to the likelihood models associated to it.
More precisely, the $j^\text{th}$ leaf distribution $\Leaf^d_{j}$ is modeled as a mixture over the likelihood dictionaries provided by the user, i.e.,
\begin{equation}   \label{eq:leaf_mixture}
\Leaf^d_{j}(x^d_n \cndbar \boldsymbol\eta^d_{j}) = \sum\nolimits_{\ell\in\mathcal{L}^{d}} w^d_{j,\ell} \, p^d_{\ell}(x^d_n \cndbar \boldsymbol\eta^d_{j,\ell}).
\end{equation}
Note that the likelihood models $p^d_\ell$ are \emph{shared} among all leaves with the same scope $d$, but each leaf has its private parameter $\boldsymbol\eta^d_{j,\ell}$ for it.
Moreover, each $\boldsymbol\eta^d_{j,\ell}$ is also equipped with a suitable prior distribution parameterized with hyper-parameters $\boldsymbol{\lambda}^d_\ell$.
For a discussion on the selection of suitable prior distributions, refer to Appendix B in the supplementary material.

The likelihood mixture coefficients ${w}_{j,\ell}^{d}$ are drawn
from a Dirichlet distribution with hyper-parameters $\boldsymbol\alpha$, i.e., $\mathbf{w}^{d}_{j} \sim \mathsf{Dir}(\boldsymbol\alpha)$.
For each entry $x_n^d$, the likelihood model is selected by an additional categorical LV $s_{j,n}^{d} \sim \mathsf{Cat}(\mathbf{w}^{d}_{j})$.
Finally, the observation entry $x_n^d$ is sampled from the selected likelihood model: $x_{n}^{d}\sim p^d_{s_{j,n}^{d}}$.

%
%
%

\section{Bayesian Inference}
\label{section:inference}
The hierarchical LV structure of SPNs allows ABDA to perform Bayesian inference via a simple and effective Gibbs sampling scheme.
To initialize the global structure, 
we use the likelihood-agnostic SPN structure learning algorithm proposed for MSPNs~\cite{Molina2017b}.
Specifically, 
we apply the RDC to split samples and features while extending it to deal with missing data (see supplementary for details).
The local level in ABDA is constructed by equipping each leaf node in $\SPN$ with the dictionaries of likelihood models as described above.
Additionally, one can introduce global type-RVs, responsible for selecting a specific likelihood model (or data type) for each feature $d$ (see Fig.~\ref{fig:abda-model:global_types} and further explanations below).

Furthermore note that, in contrast to MSPNs, ABDA is not constrained to the LV structure provided by structure learning. 
Indeed, inference in ABDA accounts for \textit{uncertainty also on the underlying latent structure}.
As a consequence, a wrongly overparameterized LV structure provided to ABDA can still be turned into a simpler one by our algorithm by using a sparse prior on the SPN weights (see Appendix H).

To perform inference, we draw samples $\mathcal{D}$ from the posterior distribution $p(\mathbf{Z}, \mathbf{s}, \boldsymbol\Omega, \mathbf{w}, \boldsymbol\eta \cndbar \Xb)$ via Gibbs sampling, where $\mathbf{Z}$ is the set of the SPN's LVs, $\mathbf{s}$ is the set of all local LVs selecting the likelihood models, $\boldsymbol \Omega$ is the set of all sum-weights, $\mathbf{w}$ are the distributions of  $\mathbf{s}$, and $\boldsymbol \eta$ is the set collecting all parameters of all likelihood models.
Next, we describe each routine involved to sample from the conditionals for each of these RVs in turn.
Algorithm~\ref{algo:gibbs} summarizes the full Gibbs sampling scheme. 
Improved mixing via a Rao-Blackwellised version of the proposed sampler is discussed in Appendix A.

\paragraph{Sampling LVs $\mathbf{Z}$.}
\label{section:spn-samp}
Given the hierarchical LV structure of $\SPN$, 
it is easy to produce a sample for $\mathbf{Z}_{n}$ by ancestral sampling, 
i.e.~by sampling an induced tree $\mathcal{T}_{n}$.
To this end we condition on a sample $\mathbf{x}_n$, and current $\boldsymbol \Omega$, $\boldsymbol \eta$ and $\mathbf{w}$.
Starting from the root of $\SPN$, for each sum node $\SumNode$
we encounter, we sample a child branch $c$ from
\begin{equation}
p(Z_{n}^{\SumNode} = c \cndbar, \mathbf{x}_{n}, \boldsymbol\Omega, \boldsymbol\eta, \mathbf{w}) 
\propto  \omega_{\SumNode,c} \, \mathcal{S}_{c}(\mathbf{x}_{n} \cndbar \boldsymbol\Omega, \boldsymbol\eta, \mathbf{w}).
\end{equation}
Note that we are effectively conditioning on the states of the ancestors of $\SumNode$.
Moreover, we have marginalized out all $\boldsymbol Z$ below $\SumNode$ and all $\boldsymbol s$,
which just amounts to evaluating $\SPN_c$ bottom-up, for given
parameters $\boldsymbol\Omega$, $\boldsymbol\eta$, and $\mathbf{w}$
\cite{Poon2011}.

Note that, in general, sampling a tree $\mathcal{T}_n$ does not reach all sum nodes.
Since these sum nodes are ``detached'' from the data, we need to sample their LVs from the prior \cite{Peharz2017}.

\paragraph{Sampling likelihood model assignments $\mathbf{s}$.}
Similarly as for LVs $\mathbf{Z}_{n}$, we sample $s_{j,n}^d$ from the posterior distribution $p(s_{j,n}^d= \ell \cndbar \mathbf{w}, \mathbf{j}, \boldsymbol\eta) \propto w_{j,\ell}^d  \, p^d_\ell(x_n^d \cndbar \boldsymbol\eta)$ if $j = j^d_n$. 

\paragraph{Sampling leaf parameters $\boldsymbol\eta$.}
Sampling $\mathbf{Z}_n$
gives rise to the leaf indices $\mathbf{j}_n$, which assign samples to leaves.
Within leaves, $\mathbf{s}$ further assign samples to likelihood models.
For parameters $\boldsymbol\eta^d_{j,\ell}$ let $\mathbf{X}^d_{j,\ell} = \{x^d_n \cndbar \forall n\colon j^d_n = j\wedge  s^d_{j,n} = \ell \}$, i.e., the samples in the $d^\text{th}$ column of $\mathbf{X}$ which have been assigned to the $\ell^\text{th}$ model in leaf $\Leaf^d_j$.
Then, $\boldsymbol\eta^d_{j,\ell}$ is updated according to 
$$
\boldsymbol\eta^d_{j,\ell} \sim 
\prod_{x \in \mathbf{X}^d_{j,\ell}} p^d_\ell(x \cndbar \boldsymbol\eta^d_{j,\ell}) \,
p(\boldsymbol\eta^d_{j,\ell} \cndbar \boldsymbol\lambda_{\ell}^{d}).
$$
When the likelihood models $p^d_\ell$ are equipped with conjugate priors, these updates are straightforward.
Moreover, also for non-conjugate priors they can easily be approximated using numerical methods, since we are dealing with single-dimensional problems.

\paragraph{ Sampling weights $\boldsymbol\Omega$ and $\mathbf{w}$.}
For each sum node $\SumNode$ we sample its associated weights from the
posterior $p(\Omega^{\SumNode} \cndbar \{\mathbf{Z}_{n}\}_{n=1}^{N})$, which is a Dirichlet distribution with parameters 
$\boldsymbol\gamma+\sum_{n=1}^{N}\mathds{1}\{(\SumNode,c)\in\mathcal{T}_{{n}}\}$. %
Similarly, we can sample the likelihood weights $\mathbf{w}^d_j$ from a Dirichlet distribution with parameters 
$\boldsymbol \alpha + 
[\sum_{n=1}^{N} \mathds{1}\{j^d_n = j \wedge s_{j,n}^d=\ell \}]_{\ell \in \mathcal{L}^d}$. 

\section{Automating exploratory data analysis}
\label{section:glocal}
In this section, we discuss how inference in ABDA can be exploited to perform  common exploratory data analysis tasks in an automatic way. 
For all inference tasks (e.g., computing $\log p(\x_{n})$), one can either condition on the model parameters, e.g., by using the maximum likelihood parameters within posterior samples $\mathcal{D}$, or perform a Monte Carlo estimate over $\mathcal{D}$.
While the former allows for efficient computations, the latter allows for quantifying the model uncertainty.
We refer to Appendixes E-G in the supplementary for more details.
\paragraph{Missing value imputation.}
Given a sample $\x_{n}=(\x_{n}^{o},\x_{n}^{m})$ comprising observed $\x_{n}^{o}$ and missing $\x^{m}_{n}$ values, 
ABDA can efficiently impute the latter as the \textit{most probable explanation} $\tilde{\x}_{n}^{m}=\argmax_{\x_{n}^{m}}\SPN(\x_{n}^{m}|\x_{n}^{o})$ via efficient approximate SPN routines~\cite{Peharz2017}.
\paragraph{Anomaly detection.}
 ABDA is robust to outliers and corrupted values since, during inference, it will tend to assign anomalous
 samples  into low-weighted mixture components, i.e., sub-networks of the underlying SPN or leaf likelihood models.
 Outliers will tend to be either grouped into anomalous \textit{micro-clusters}~\cite{Chandola2009} or assigned to the tails of a likelihood model.

 Therefore, $\log p(\x_{n})$ can be used as a strong signal to indicate $\x_{n}$ is an outlier (in the transductive case) or a \textit{novelty} (in the inductive case) \cite{Goldstein2016}.

\paragraph{Data type and likelihood discovery.}
ABDA automatically estimates \textit{local} uncertainty over likelihood models and statistical types by inferring the dictionary coefficients $w^{d}_{j,\ell}$ for a leaf $\Leaf^d_{j}$.

However, we can extend ABDA to reason about data type also on a \textit{global} level by explicitly introducing a \emph{type variable} $T^d$ for feature $d$.
This type variable might either represent a \emph{parametric type}, e.g., Gaussian or Gamma, or a \emph{data type}, e.g., real-valued or positive-real-valued, in a similar way to ISLV~\cite{Valera2017a}. 
To this end, we introduce state-indicators for each type variable $T^d$, and connect them into each likelihood model in leaf $\Leaf^d_j$, as shown in Fig.~\ref{fig:abda-model:global_types}.
In this example, we introduced the state-indicators $T^d_{\mathcal{N}}$ and $T^d_{\Gamma}$ to represent a global type $T^d$ which distinguishes between Gaussian and Gamma distribution.

Note that this technique is akin to the \emph{augmentation} of SPNs \cite{Peharz2017}, i.e., explicitly introducing the LVs for sum nodes.
Here, however, the introduced $\{T^d\}_{d=1}^{D}$ have an explicit intended semantic as global data type variables.
Crucially, after introducing $\{T^d\}_{d=1}^{D}$, the underlying SPN is still complete and decomposable, i.e., we can easily perform \emph{inference} over them.
In particular, we can estimate the posterior probability for a feature $d$ to have a particular type as:
\begin{equation}
p(T^{d} \cndbar \Xb) \approx
\frac{1}{|\mathcal{D}|} 
\sum_{\{\boldsymbol \Omega, \mathbf{w}, \boldsymbol \eta\} \in \mathcal{D}}
\SPN(T^{d}  \cndbar  \Omega, \mathbf{w}, \boldsymbol \eta).
\label{eq:global_weights}
\end{equation}
The marginal terms $\SPN(T^{d}  \cndbar  \Omega, \mathbf{w}, \boldsymbol \eta)$ are easily obtained via SPN inference -- we simply need to set all leaves and all indicators for $\{T^{d'}\}_{d' \not= d}$ equal to $1$ \cite{Poon2011}.

\paragraph{Dependency pattern mining.}
ABDA is able to retrieve \textit{global dependencies}, e.g., by
computing pairwise hybrid mutual information, in a similar way to
MSPNs~\cite{Molina2017b}.

Additionally, ABDA can provide users \emph{local} patterns in the form of dependencies  
within a data partition $\Xb^{\Node}\subseteq\mathbf{X}$ associated with any node $\Node$ in $\SPN$.
In particular, let $\Xb^{\Node}$ contain all entries $x^{d}_{n}$ such that $d$ is in the scope of $\Node$
and $n$ such that $\mathbf{Z}_{n}$ yield an induced tree from the SPN's root to $\Node$.
Then, for each leaf $\Leaf^{d}_{j}$ and likelihood model $\ell\in\mathcal{L}^{d}$
one can extract a pattern of the form $\mathcal{P}\colon \pi_{l}^{d}\leq X^{d}<\pi_{h}^{d}$, where $[\pi_{l}^{d},\pi_{h}^{d})$ is an interval in the domain of $X^{d}$.
The pattern can be deemed as present, when its probability exceeds a user-defined threshold $\theta$: $p_{\Leaf^d_{j}}(\mathcal{P}) \geq \theta$.

A conjunction of patterns $\mathcal{P}^{\Node}=\mathcal{P}_{1}\wedge\ldots\wedge\mathcal{P}_{|\scope(\Node)|}$ represents the correlation among features in $\Xb^{\Node}$, and its \textit{relevance} can be quantified as $p_{\SPN}(\mathcal{P}^{\Node})$.
This technique relates to the notion of \textit{support} in association rule mining~\cite{Agrawal1994}, whose binary patterns are here generalized to also support continuous and (non-binary) discrete RVs.

%
\section{Experimental evaluation}
\label{section:exps}

We empirically evaluate ABDA on synthetic and real-world datasets both as a density estimator and as a tool to perform several exploratory data analysis tasks.
Specifically, we investigate the following questions:
\begin{description}[noitemsep,topsep=0pt,parsep=0pt,partopsep=0pt]
\item[\textbf{(Q1)}] How does ABDA estimate likelihoods and statistical data types when a ground truth is available?
\item[\textbf{(Q2)}] How accurately does ABDA perform density estimation and imputation over unseen real-world data?
\item[\textbf{(Q3)}] How robust is ABDA w.r.t. anomalous data?
\item[\textbf{(Q4)}] How can ABDA be exploited to unsupervisedly extract dependency patterns?
\end{description}

\paragraph{Experimental setting}
In all experiments, we use a symmetric Dirichlet prior with $\gamma=10$ for sum weights $\boldsymbol\Omega$ and a sparse symmetric prior with $\alpha=0.1$ for the leaf likelihood weights $\mathbf{w}_{j}^{d}$.
We consider the following likelihoods for continuous data: Gaussian distributions ($\mathcal{N}$) for $\mathsf{REAL}$-valued data; Gammas ($\Gamma$) and exponential ($\mathsf{Exp}$) for $\mathsf{POS}$itive real-valued data; and, for discrete data, 
we consider Poisson ($\mathsf{Poi}$) and Geometric ($\mathsf{Geo}$) distributions for $\mathsf{NUM}$erical data, and Categorical ($\mathsf{Cat}$) for $\mathsf{NOM}$inal data, while a Bernoulli for Binary data in the outlier detection case.
For details on the prior distributions employed, for the likelihood parameters and their hyper-parameters, please refer to the Appendix.

\begin{figure*}[!t]
    \centering
    \begin{subfigure}[t]{0.2\textwidth}
        \includegraphics[width=1\columnwidth]{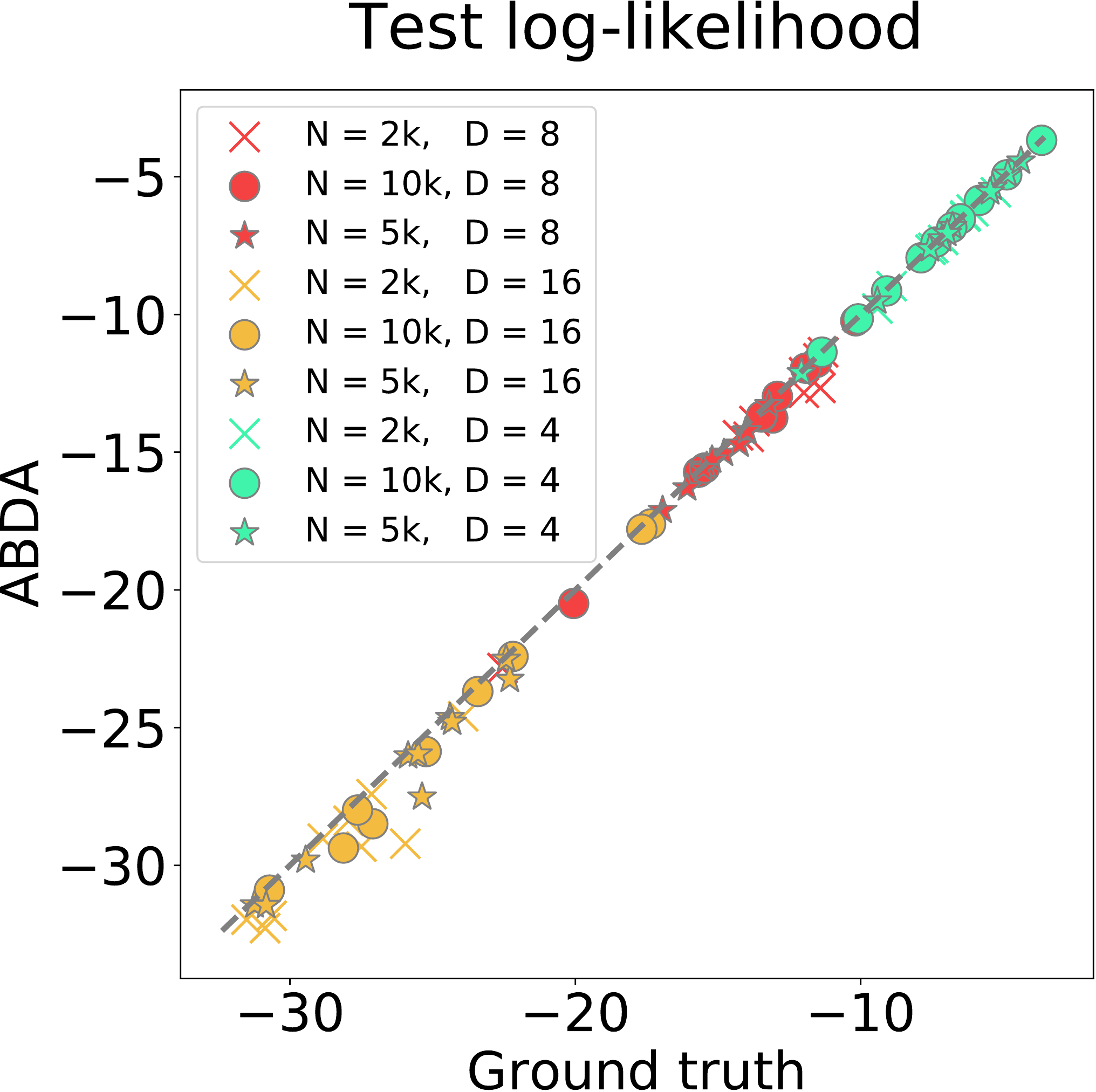} 
        \caption{}
        \label{fig:synth:ll}
    \end{subfigure}\hspace{10pt}
    \begin{subfigure}[t]{0.14\textwidth}\vspace{-93pt}
        \includegraphics[width=.98\columnwidth]{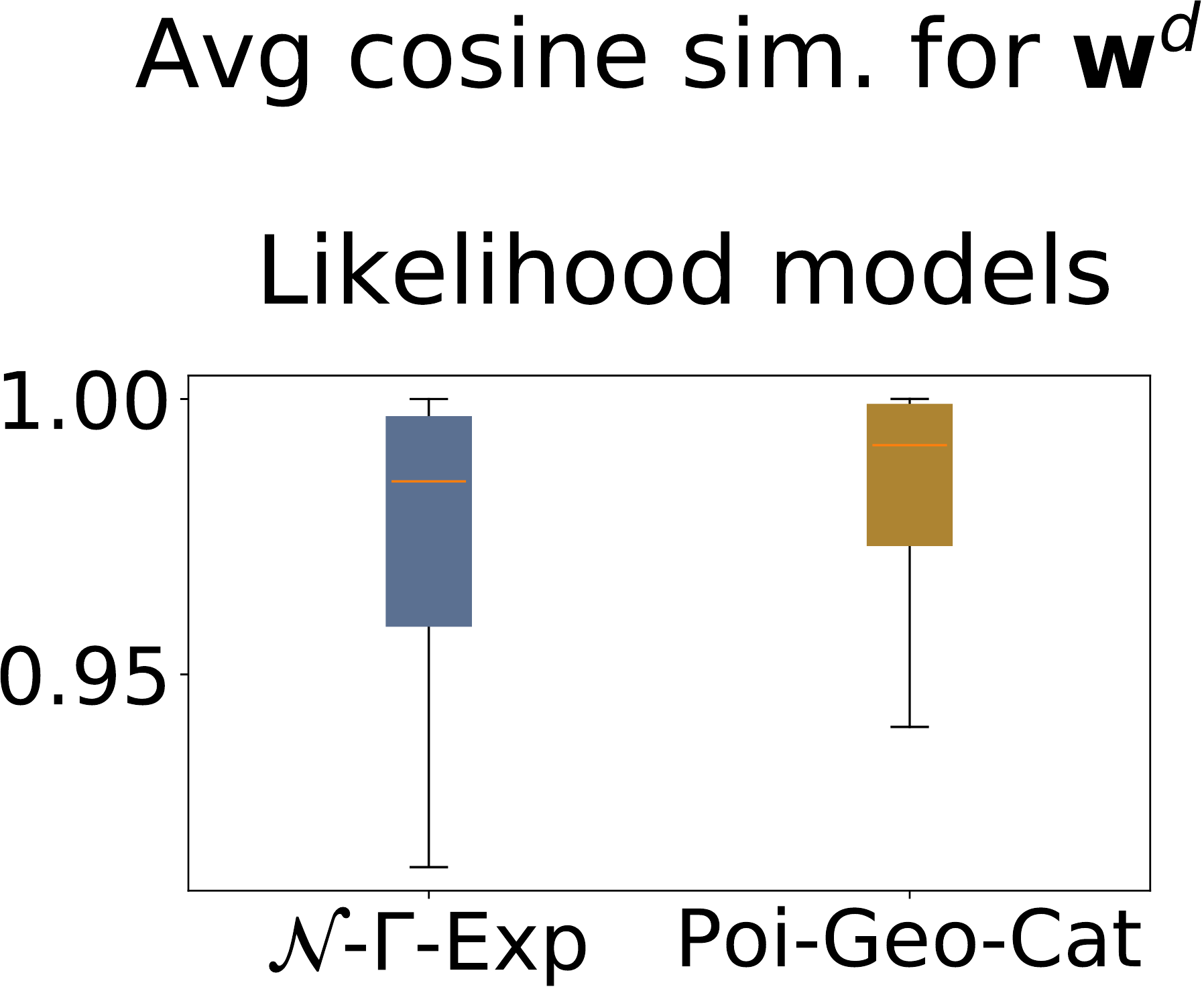}\\[4pt]
        \includegraphics[width=.95\columnwidth]{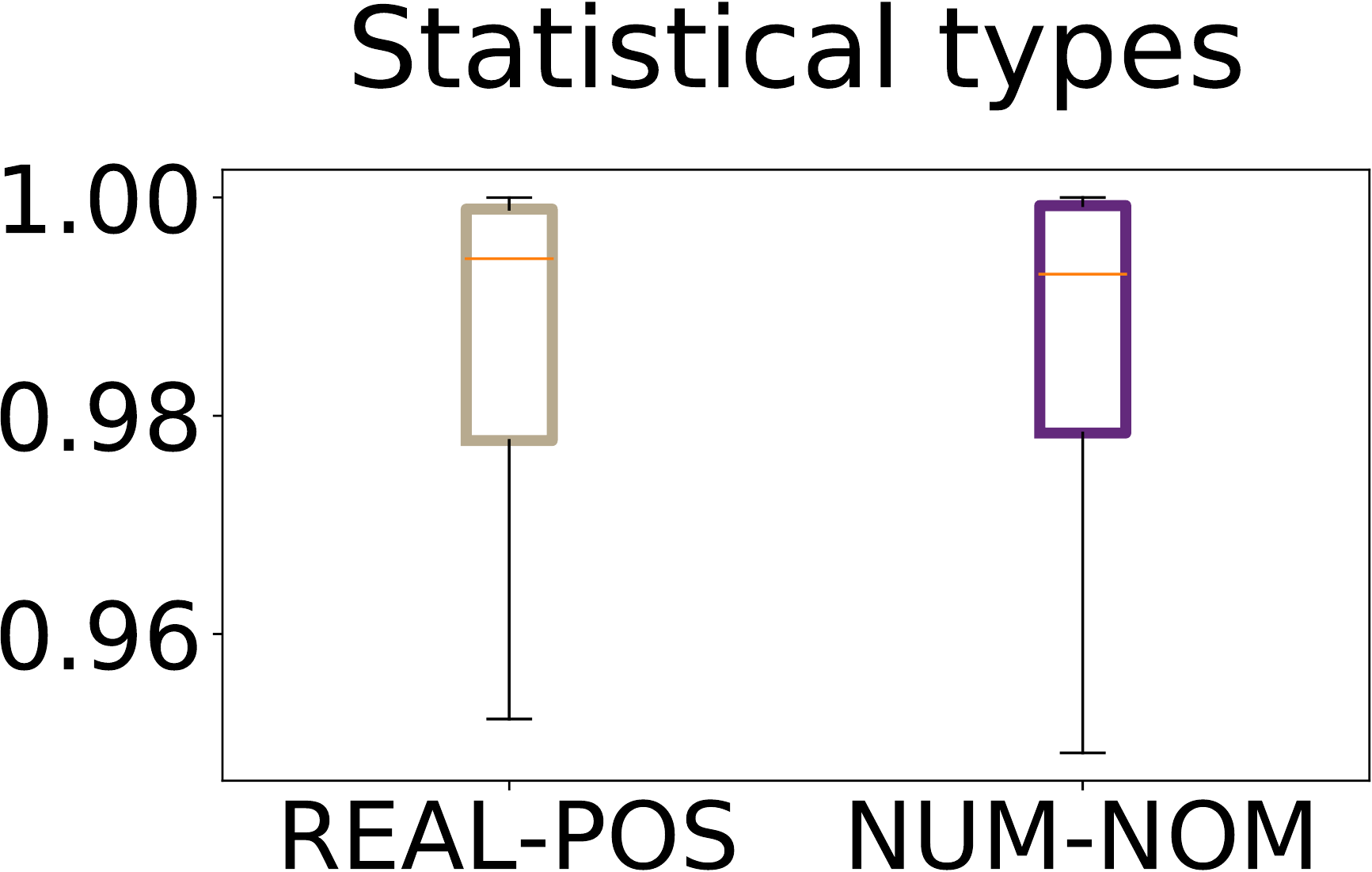}
        \caption{}
        \label{fig:synth:bp}
    \end{subfigure}\hspace{10pt}\begin{subfigure}[t]{0.32\textwidth}\vspace{-93pt}
        \includegraphics[width=0.31\columnwidth]{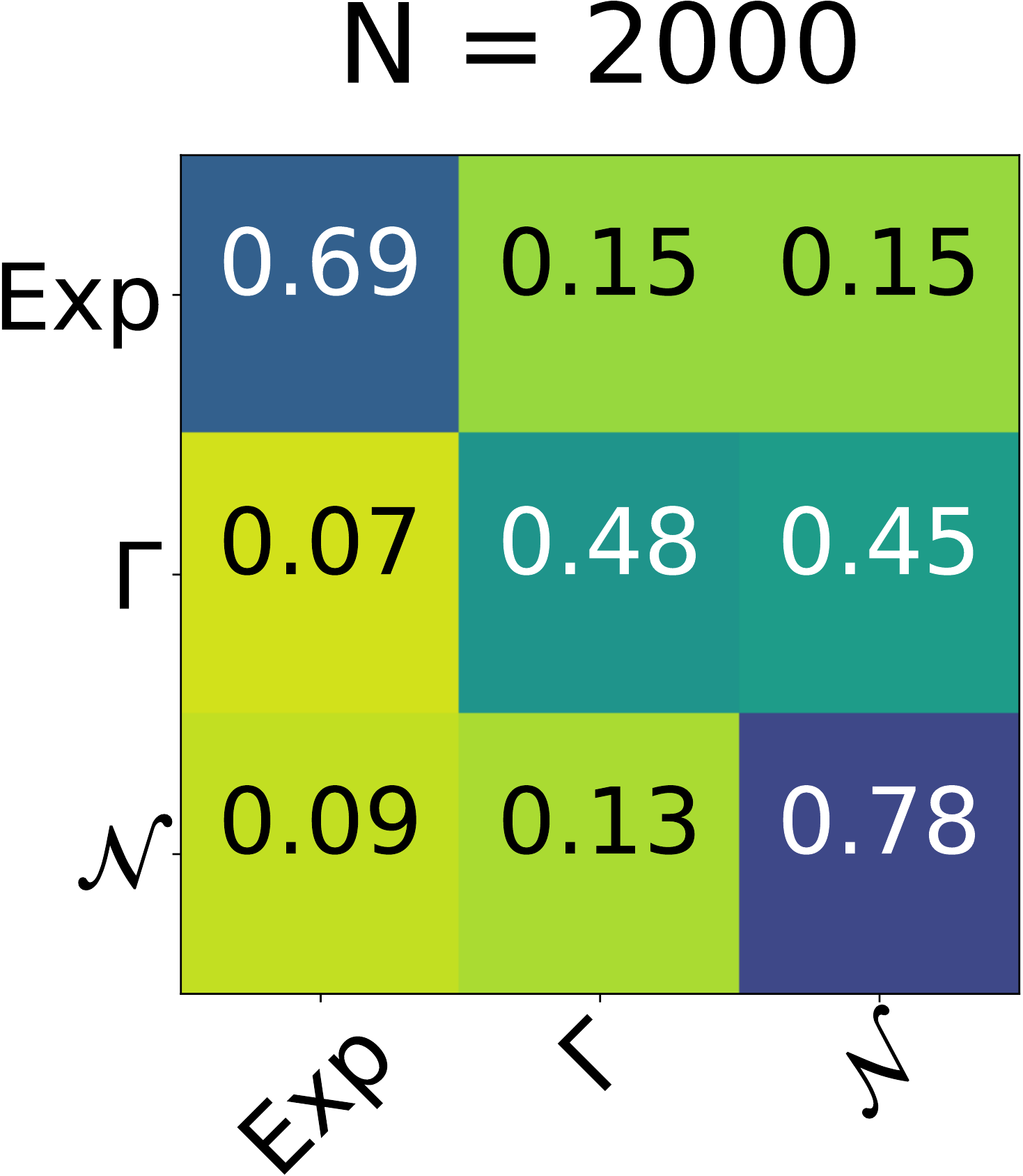}\hspace{-2pt}
        \includegraphics[width=0.32\columnwidth]{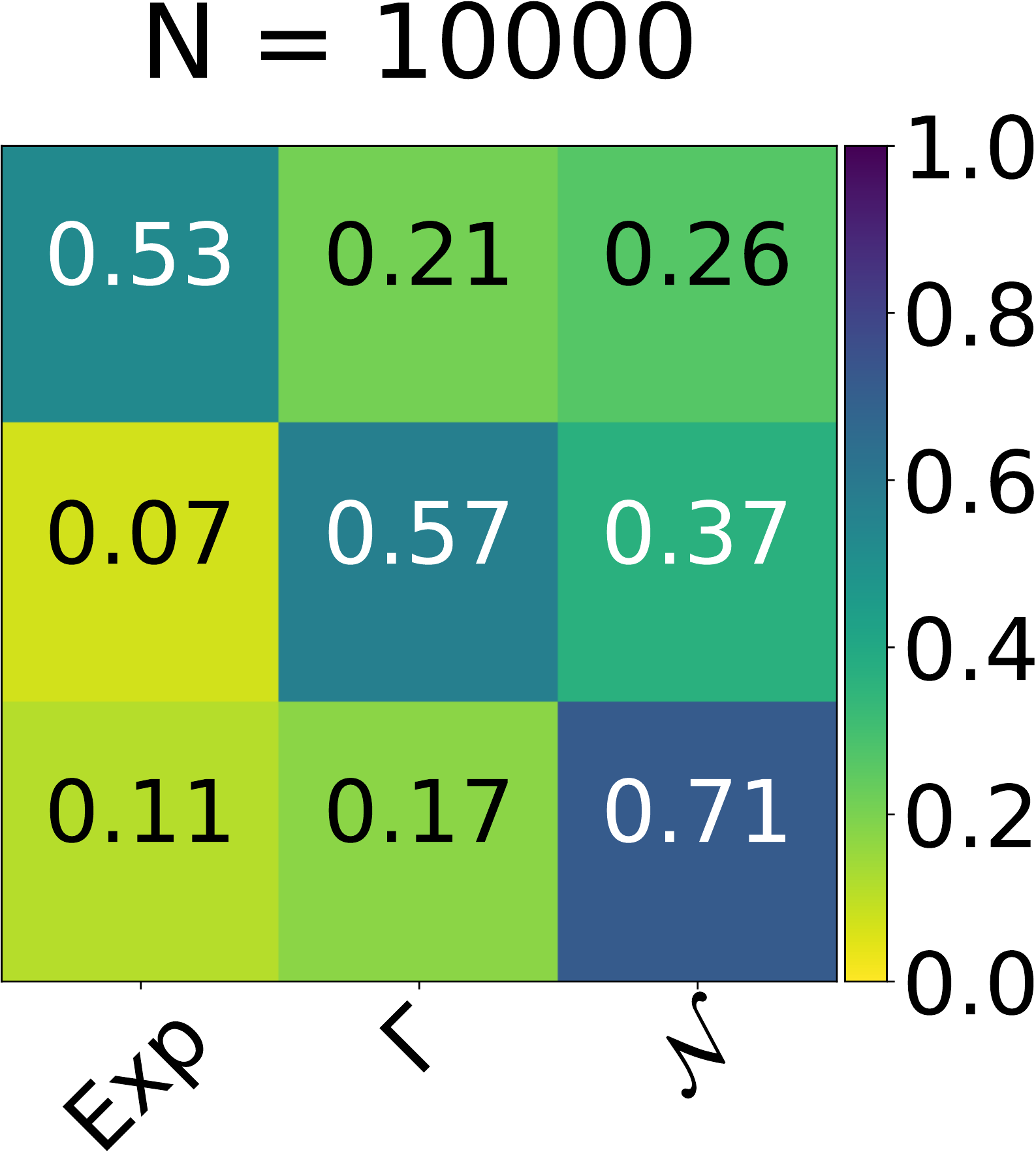}\hspace{0pt}
        \includegraphics[width=0.315\columnwidth]{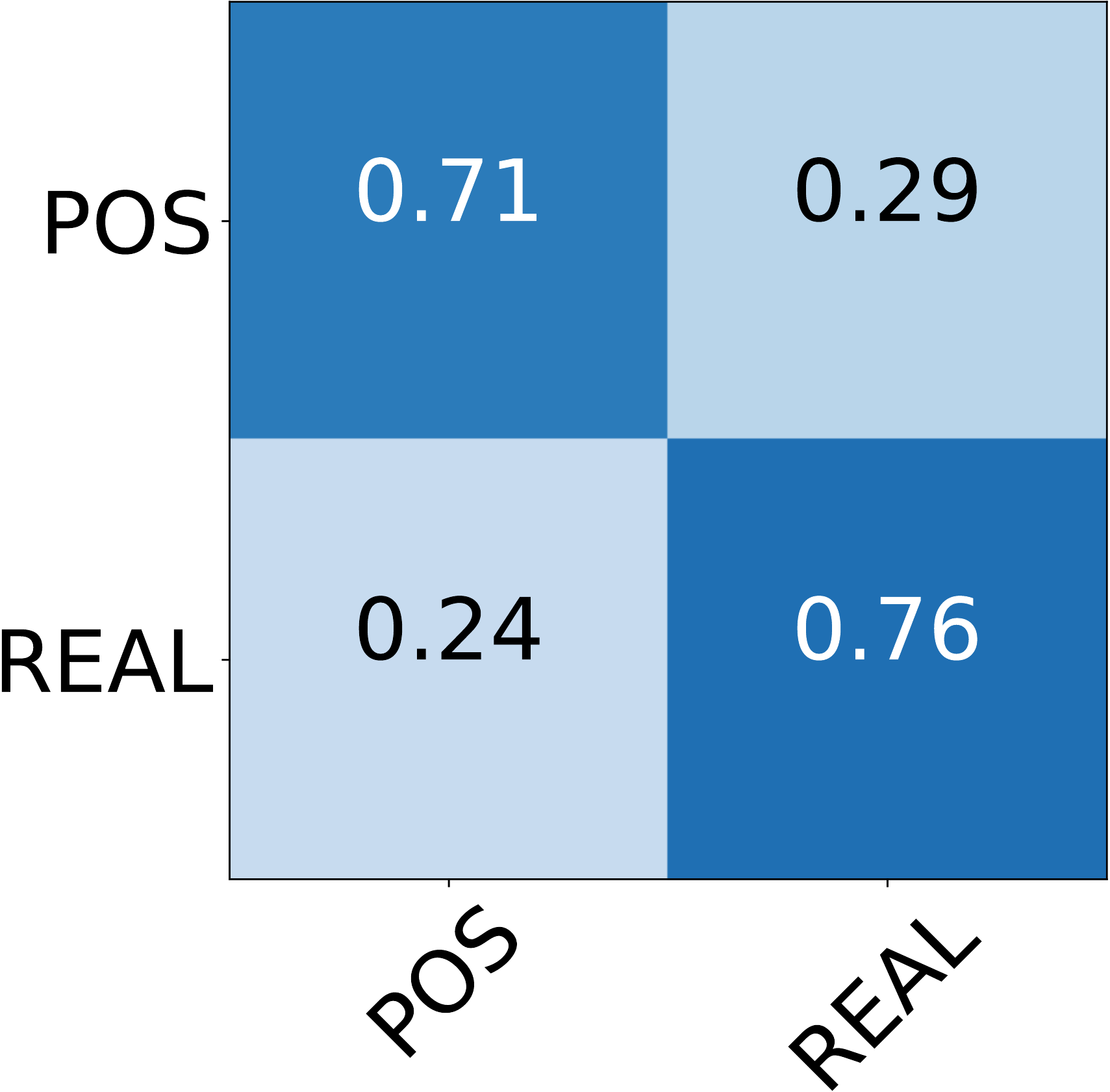}\\
        \includegraphics[width=0.31\columnwidth]{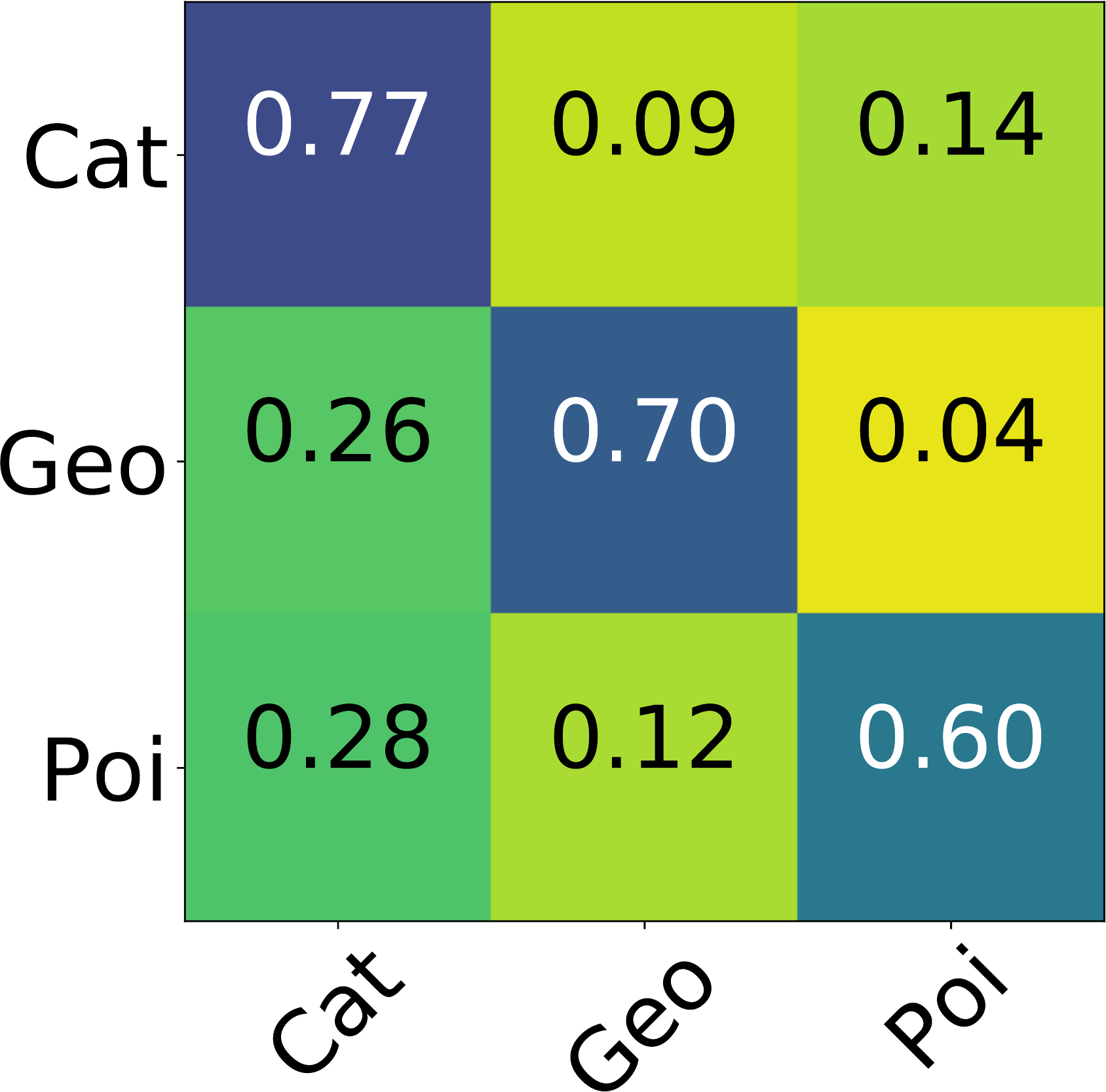}\hspace{-2pt}
        \includegraphics[width=0.315\columnwidth]{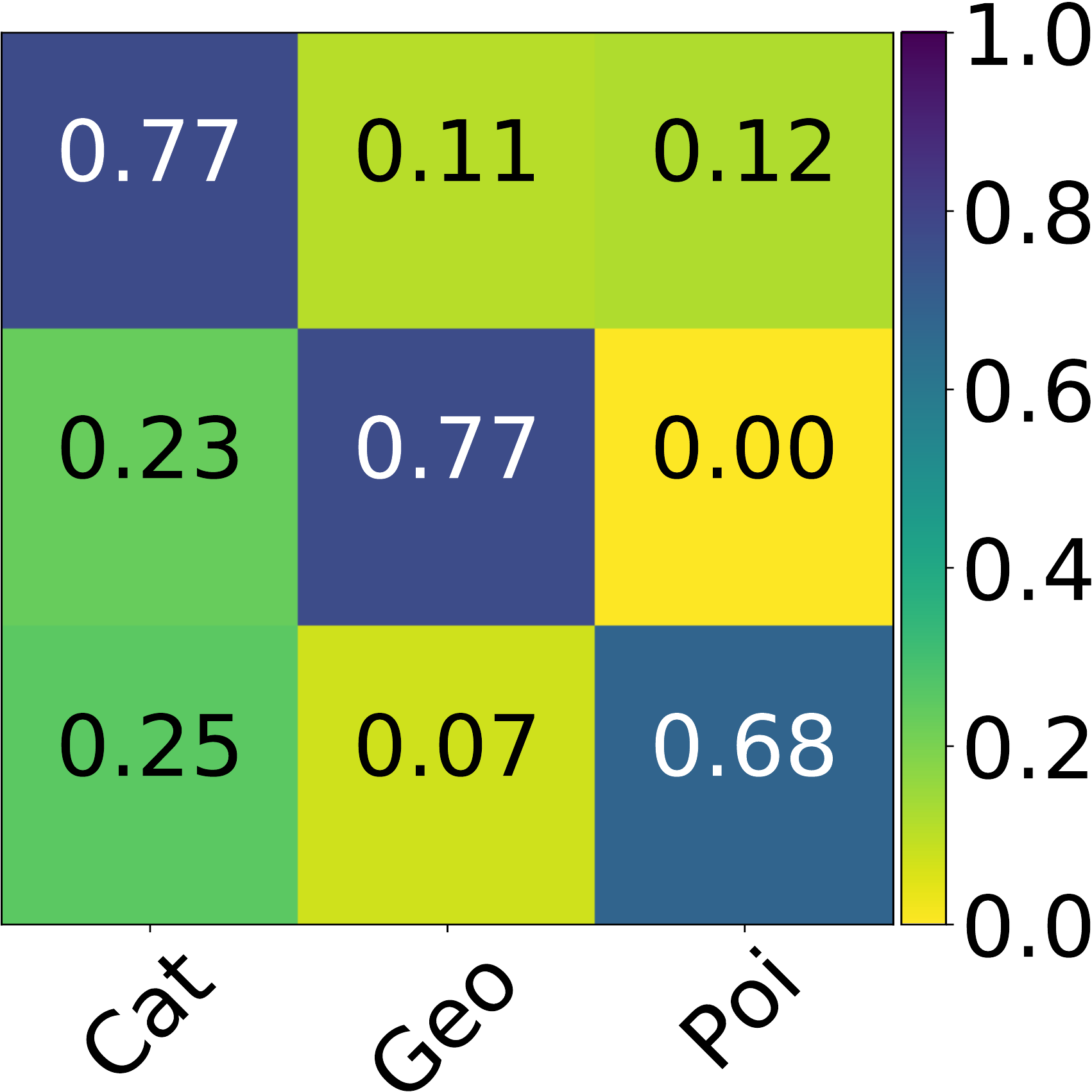}\hspace{0pt}
        \includegraphics[width=0.315\columnwidth]{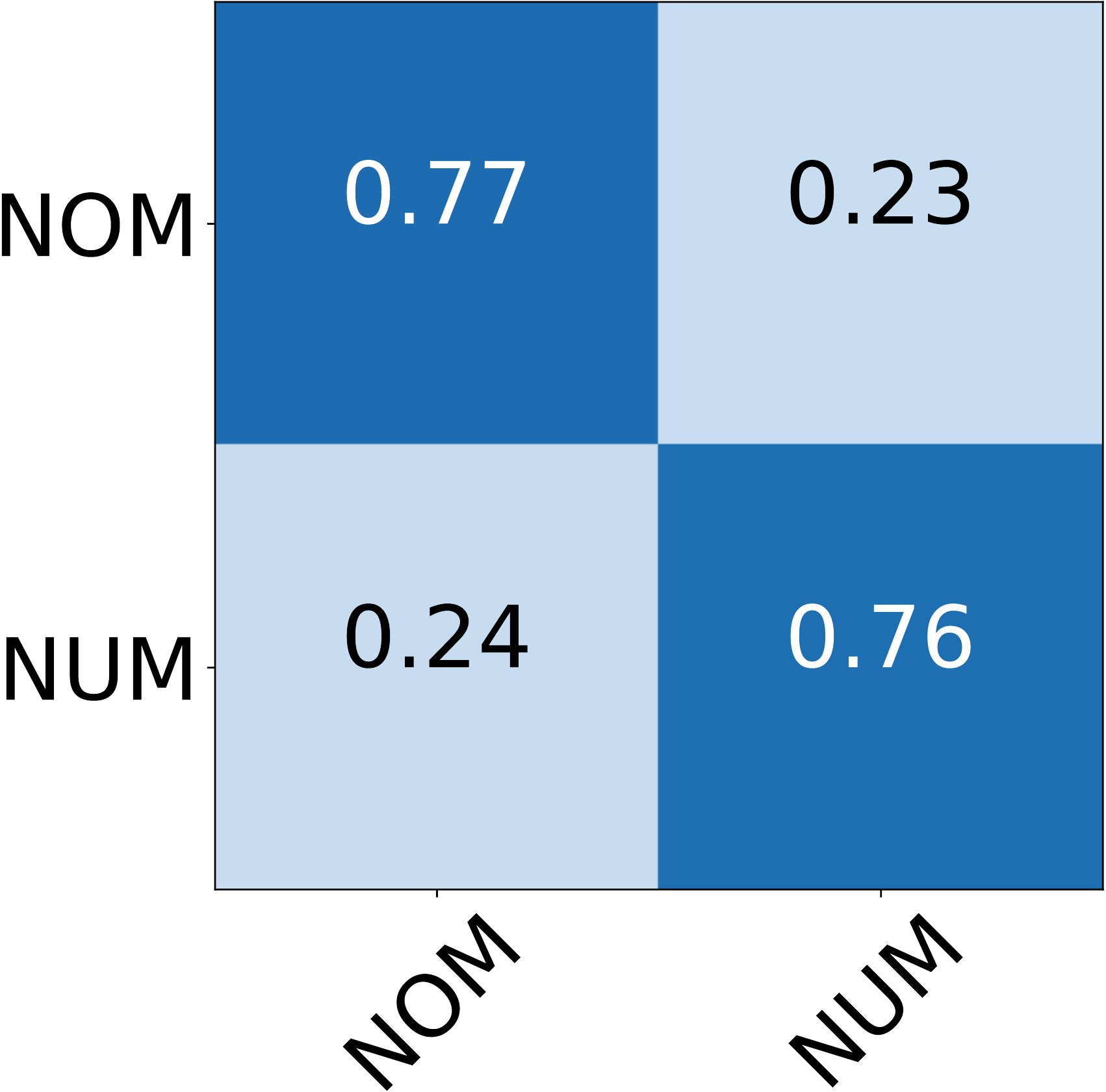}
        \caption{}
        \label{fig:synth:cm}
    \end{subfigure}\hspace{10pt}\begin{subfigure}[t]{0.2\textwidth}
        \includegraphics[width=0.99\columnwidth]{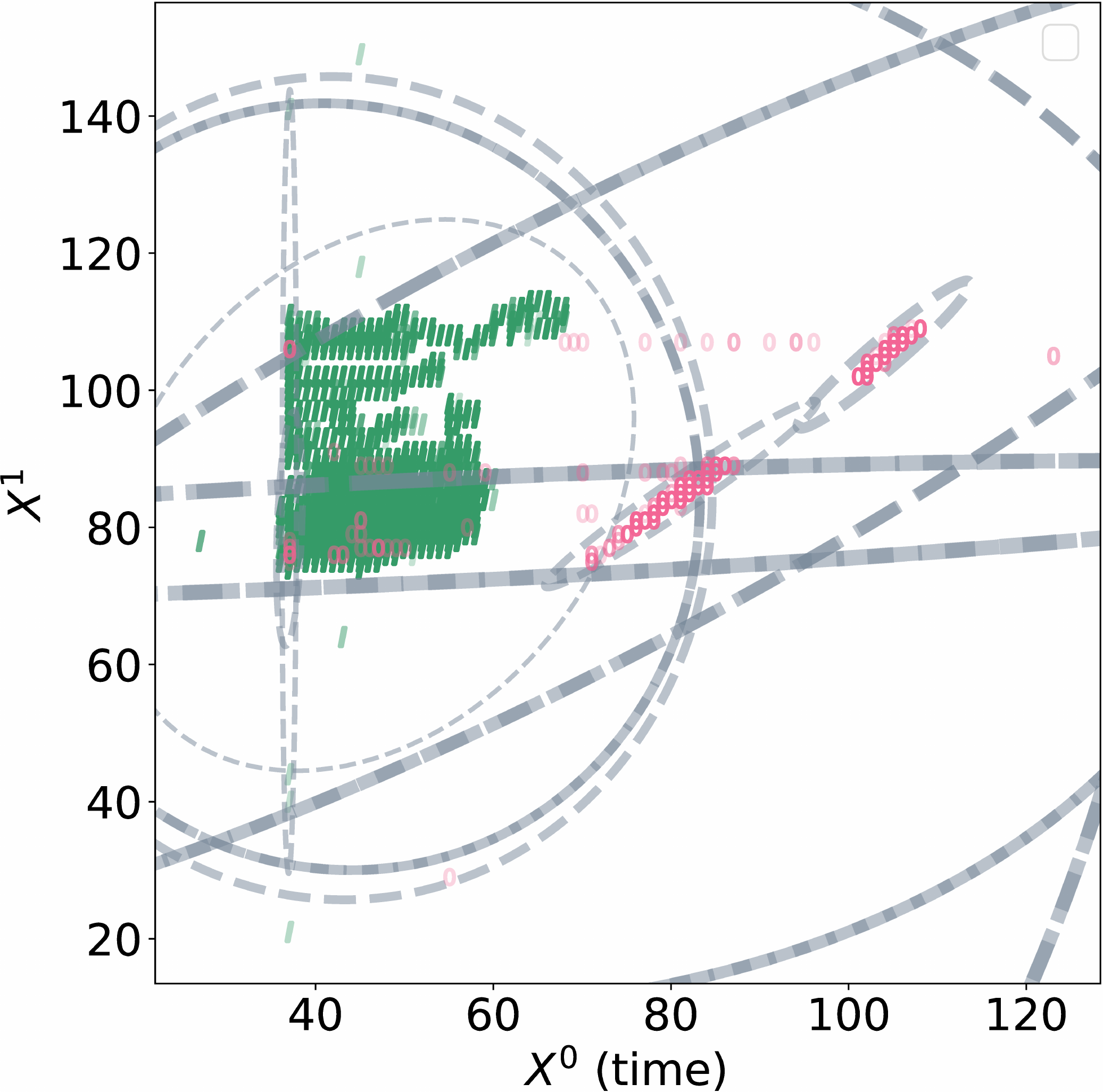}
        \caption{}
        \label{fig:od}
    \end{subfigure}

    \caption{\textbf{(\subref{fig:synth:ll})} Mean test log-likelihood on synthetic data w.r.t.~the ground truth.  \textbf{(\subref{fig:synth:bp})} Distributions of the mean cosine similarity between true and retrieved uncertainty weights over likelihood models (top) and statistical types (bottom). \textbf{(\subref{fig:synth:cm})} Confusion matrices for the most likely likelihood model resp.~statistical type. \textbf{(\subref{fig:od})} ABDA separates outliers (pink `O') from inliers (green `I') via hierarchical partitioning on \textsf{Shuttle} data, see Appendix F. Best viewed in colors.}
    \label{fig:synth}
    
      \vspace{-6pt}
\end{figure*}

\begin{table}[!t]
  \caption{Density estimation. Mean test log-likelihood on real-world benchmark datasets for trans-/inductive scenarios. Best values are bold.}
    \label{tab:trans}
\scriptsize
    \setlength{\tabcolsep}{5pt}
    \begin{tabular}{r r r r r r r r r r r r r r r r r}
    &\multicolumn{3}{c}{transductive setting (10\% mv)}&\multicolumn{3}{c}{transductive setting (50\% mv)}&\multicolumn{2}{c}{inductive setting}\\
    \cmidrule(l{1em}){2-4}\cmidrule(l{1em}){5-7}\cmidrule(l{1em}){8-9}
    &\textsf{ISLV}&\textsf{ABDA}&\textsf{MSPN}&\textsf{ISLV}&\textsf{ABDA}&\textsf{MSPN}&\textsf{ABDA}&\textsf{MSPN}\\
         \textsf{Abalone} & -1.15$\scriptstyle\scriptstyle\pm 0.12$ & -0.02$\scriptstyle\pm0.03$& \textbf{0.20}& -0.89$\scriptstyle\pm 0.36$& -0.05$\scriptstyle\pm0.02$& \textbf{0.14}& 2.22$\scriptstyle\pm0.02$& \textbf{9.73}& \\
         \textsf{Adult} & - & \textbf{-0.60}$\scriptstyle\pm\mathbf{0.02}$ & -3.46& -& \textbf{-0.69}$\scriptstyle\pm\mathbf{0.01}$&  -5.83& \textbf{-5.91}$\scriptstyle\pm\mathbf{0.01}$&-44.07&\\
         \textsf{Austral.} & -7.92$\scriptstyle\pm0.96$ & \textbf{-1.74}$\scriptstyle\pm\mathbf{0.19}$&  -3.85& -9.37$\scriptstyle\pm0.69$& \textbf{-1.63}$\scriptstyle\pm\mathbf{0.04}$&-3.76& \textbf{-16.44}$\scriptstyle\pm\mathbf{0.04}$&-36.14&\\
         \textsf{Autism} & -2.22$\scriptstyle\pm0.06$ &  \textbf{-1.23}$\scriptstyle\pm\mathbf{0.02}$& -1.54& -2.67$\scriptstyle\pm0.16$& \textbf{-1.24}$\scriptstyle\pm\mathbf{0.01}$& -1.57& \textbf{-27.93}$\scriptstyle\pm\mathbf{0.02}$&-39.20\\
         \textsf{Breast} & -3.84$\scriptstyle\pm0.05$ &  -2.78$\scriptstyle\pm0.07$& \textbf{-2.69}& -4.29$\scriptstyle\pm0.17$&\textbf{-2.85}$\scriptstyle\pm\mathbf{0.01}$& -3.06&  \textbf{-25.48}$\scriptstyle\pm\mathbf{0.05}$&-28.01\\
         \textsf{Chess} & -2.49$\scriptstyle\pm0.04$ &\textbf{-1.87}$\scriptstyle\pm\mathbf{0.01}$ & -3.94& -2.58$\scriptstyle\pm0.04$&\textbf{-1.87}$\scriptstyle\pm\mathbf{0.01}$ & -3.92&\textbf{-12.30}$\scriptstyle\pm\mathbf{0.00}$&-13.01\\
         \textsf{Crx} & -12.17$\scriptstyle\pm1.41$ &  \textbf{-1.19}$\scriptstyle\pm\mathbf{0.12}$& -3.28& -11.96$\scriptstyle\pm1.01$ & \textbf{-1.20}$\scriptstyle\mathbf\pm\mathbf{0.04}$& -3.51&\textbf{-12.82}$\scriptstyle\pm\mathbf{0.07}$ &-36.26\\
         \textsf{Dermat.} &-2.44$\scriptstyle\pm 0.23$& \textbf{-0.96}$\scriptstyle\pm\mathbf{0.02}$& -1.00& -3.57$\scriptstyle\pm 0.32$& \textbf{-0.99}$\scriptstyle\pm\mathbf{0.01}$& -1.01  &\textbf{-24.98}$\scriptstyle\pm\mathbf{0.19}$&-27.71\\
         \textsf{Diabetes} & -10.53$\scriptstyle\pm1.51$ &  \textbf{-2.21}$\scriptstyle\pm\mathbf{0.09}$& -3.88& -12.52$\scriptstyle\pm0.52$& \textbf{-2.37}$\scriptstyle\pm\mathbf{0.09}$& -4.01&\textbf{-17.48}$\scriptstyle\pm\mathbf{0.05}$ &-31.22\\
         \textsf{German} & -3.49$\scriptstyle\pm 0.21$ & \textbf{-1.54}$\scriptstyle\pm\mathbf{0.01}$& -1.58 & -4.06$\scriptstyle\pm 0.28$& \textbf{-1.55}$\scriptstyle\pm\mathbf{0.01}$& -1.60&  \textbf{-25.83}$\scriptstyle\pm\mathbf{0.05}$&-26.05\\
         \textsf{Student} & -2.83$\scriptstyle\pm 0.27$ &\textbf{-1.56}$\scriptstyle\pm\mathbf{0.03}$ & -1.57& -3.80$\scriptstyle\pm 0.29$& \textbf{-1.57}$\scriptstyle\pm\mathbf{0.01}$ & -1.58& \textbf{-28.73}$\scriptstyle\pm\mathbf{0.10}$&-30.18 &\\
         \textsf{Wine} &  -1.19$\scriptstyle\pm 0.02$&-0.90$\scriptstyle\pm 0.02$ & \textbf{-0.13}& -1.34$\scriptstyle\pm 0.01$ & -0.92$\scriptstyle\pm 0.01$& \textbf{-0.41}&-10.12$\scriptstyle\pm0.01$&\textbf{-0.13}\\
         \cmidrule(l{1em}){2-4}\cmidrule(l{1em}){5-7}\cmidrule(l{1em}){8-9}
    \end{tabular}
\end{table}

\begin{table}[!t]
  \caption{Anomaly detection with ABDA. Mean test log-likelihood on real-world benchmark datasets for trans-/inductive scenarios. Best values are bold.}
    \label{tab:anon}
\scriptsize
    \setlength{\tabcolsep}{5pt}
    \begin{tabular}{r r r r r r }
    \multicolumn{5}{c}{outlier detection}\\
    \cmidrule(l{1em}){2-5}
    
    &\textsf{1SVM}&\textsf{LOF}&\textsf{HBOS}&\textsf{ABDA}&\\
 \textsf{Aloi}&51.71$\scriptstyle\scriptstyle\pm 0.02$&\textbf{74.19}$\scriptstyle\scriptstyle\pm0.70$&52.86$\scriptstyle\scriptstyle\pm
                                          0.53$&47.20$\scriptstyle\scriptstyle\pm 0.02$&\\
         \textsf{Thyroid}&46.18$\scriptstyle\scriptstyle\pm0.39$&62.38$\scriptstyle\scriptstyle\pm1.04$&62.77$\scriptstyle\scriptstyle\pm3.69$&\textbf{84.88}$\scriptstyle\scriptstyle\pm0.96$&\\
         \textsf{Breast}&45.77$\scriptstyle\scriptstyle\pm11.1$&98.06$\scriptstyle\scriptstyle\pm0.70$&94.47$\scriptstyle\scriptstyle\pm0.79$&\textbf{98.36}$\scriptstyle\scriptstyle\pm0.07$&\\
         \textsf{Kdd99}&53.40$\scriptstyle\scriptstyle\pm3.63$&46.39$\scriptstyle\scriptstyle\pm1.95$&87.59$\scriptstyle\scriptstyle\pm4.70$&\textbf{99.79}$\scriptstyle\scriptstyle\pm0.10$&\\
         \textsf{Letter}&63.38$\scriptstyle\scriptstyle\pm17.6$&\textbf{86.55}$\scriptstyle\scriptstyle\pm2.23$&60.47$\scriptstyle\scriptstyle\pm1.80$&70.36$\scriptstyle\scriptstyle\pm0.01$&\\
         \textsf{Pen-glo}&46.86$\scriptstyle\scriptstyle\pm1.02$&87.25$\scriptstyle\scriptstyle\pm1.94$&71.93$\scriptstyle\scriptstyle\pm1.68$&\textbf{89.87}$\scriptstyle\scriptstyle\pm2.87$&\\
         \textsf{Pen-loc}&44.11$\scriptstyle\scriptstyle\pm6.07$&\textbf{98.72}$\scriptstyle\scriptstyle\pm0.20$&64.30$\scriptstyle\scriptstyle\pm2.70$&90.86$\scriptstyle\scriptstyle\pm0.79$&\\
         \textsf{Satellite}&52.14$\scriptstyle\scriptstyle\pm3.08$&83.51$\scriptstyle\scriptstyle\pm11.98$&90.92$\scriptstyle\scriptstyle\pm0.16$&\textbf{94.55}$\scriptstyle\scriptstyle\pm0.68$&\\
         \textsf{Shuttle}&89.37$\scriptstyle\scriptstyle\pm5.13$&66.29$\scriptstyle\scriptstyle\pm1.69$&\textbf{98.47}$\scriptstyle\scriptstyle\pm0.24$&78.61$\scriptstyle\scriptstyle\pm0.02$&\\
         \textsf{Speech}&45.61$\scriptstyle\scriptstyle\pm3.64$&\textbf{49.37}$\scriptstyle\scriptstyle\pm0.87$&47.47$\scriptstyle\scriptstyle\pm0.10$&46.96$\scriptstyle\scriptstyle\pm0.01$&\\
          \cmidrule(l{1em}){2-5}\\[-11pt]
    \end{tabular}
\end{table}

\subsubsection{(Q1) Likelihood and statistical type uncertainty}
We use synthetic data in order to have control over the ground-truth distribution of the data.
To this end, we generate 90 synthetic datasets with different combinations of likelihood models and dependency structures with different numbers of samples $N\in\{2000, 5000, 10000\}$ and numbers of features $D\in\{4, 8, 16\}$.
For each possible combination of values, we create ten independent datasets of $N$ samples (reserving $20\%$ of them for testing), yielding $90$ data partitionings randomly built by mimicking the SPN learning process of~\cite{Gens2013,Vergari2015}.
The leaf distributions have been randomly drawn from the aforementioned likelihood dictionaries.
We then perform density estimation with ABDA on these datasets. 
See Appendix C for details on ABDA inference and the data generation process.

Fig.~\ref{fig:synth} summarizes our results.
In Fig.~\ref{fig:synth:ll} we see that ABDA's likelihood matches the true model closely in all settings, indicating that ABDA is an accurate density estimator.
Additionally, as shown in Fig.~\ref{fig:synth:bp}, ABDA is able to capture the uncertainty over data types, as it achieves high average cosine similarity between the ground truth type distribution and the inferred posterior over data type $p(T^d \cndbar \Xb)$ (see Eq.~\eqref{eq:global_weights}), both for 
i) 
likelihood functions (using $\mathcal{N}$, $\Gamma$ and $\mathsf{Exp}$ for continuous and $\mathsf{Pos}$, $\mathsf{Geo}$ and $\mathsf{Cat}$ for discrete features); and 
ii)
the corresponding statistical data types (using $\mathsf{POS}$ and $\mathsf{REAL}$ for continuous and  $\mathsf{NUM}$ and $\mathsf{NOM}$ for discrete features).

Furthermore, when forcing a hard decision on distributions and data types, ABDA delivers accurate predictions.
As shown in the confusion matrices in Fig.~\ref{fig:synth:cm}, selecting the most probable likelihood (data type) based on ABDA inference matches the ground truth up to the expected indiscernibility due to finite sample size.
For further discussions, please refer to \cite{Valera2017a}.

\begin{figure}[!t]
    \centering
    \begin{subfigure}[t]{0.23\columnwidth}
        \includegraphics[width=1\linewidth]{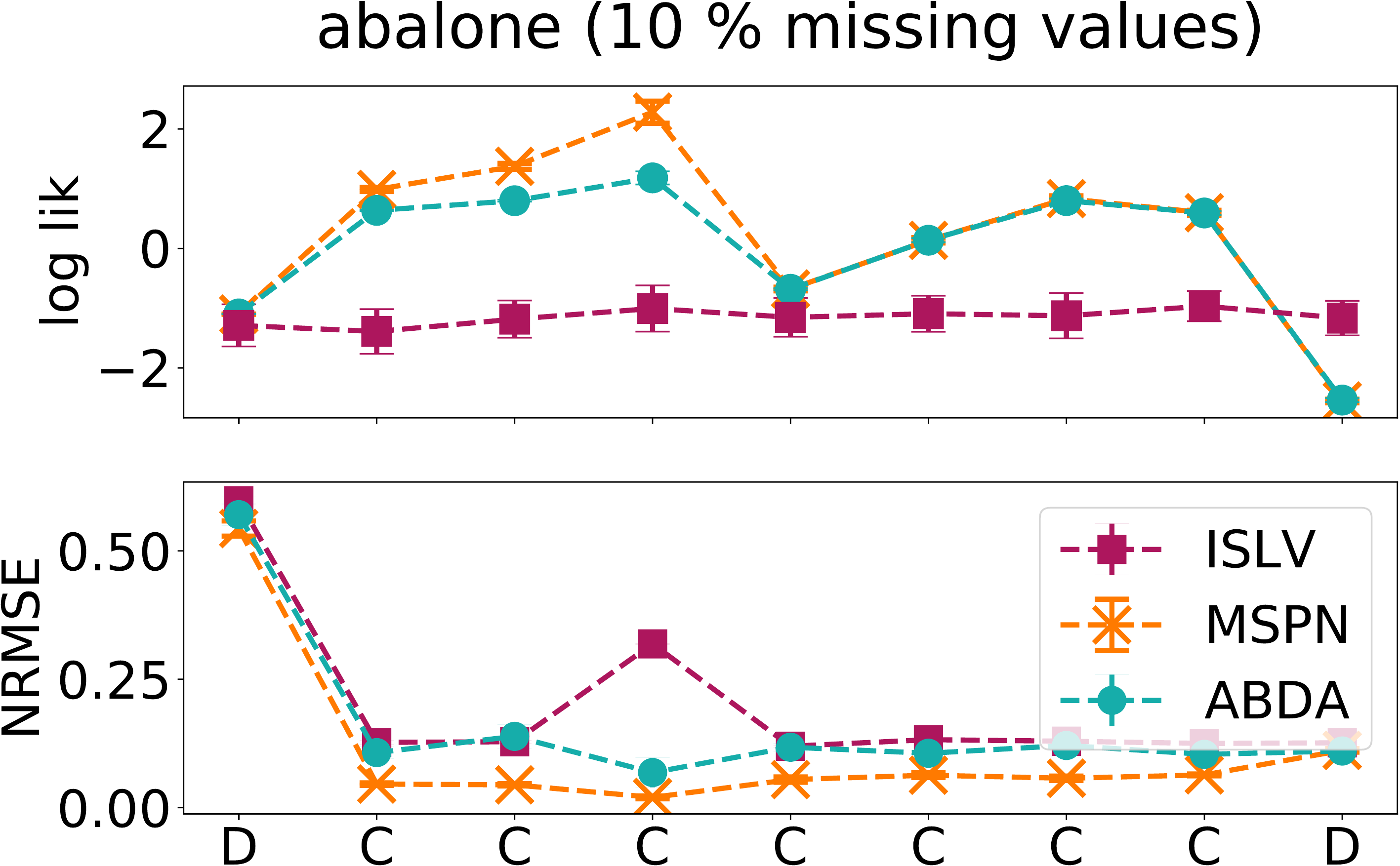}
        \caption{Abalone}
        \label{fig:miss:abalone}
    \end{subfigure}\hspace{3pt}
    \begin{subfigure}[t]{0.23\columnwidth}
        \includegraphics[width=1\linewidth]{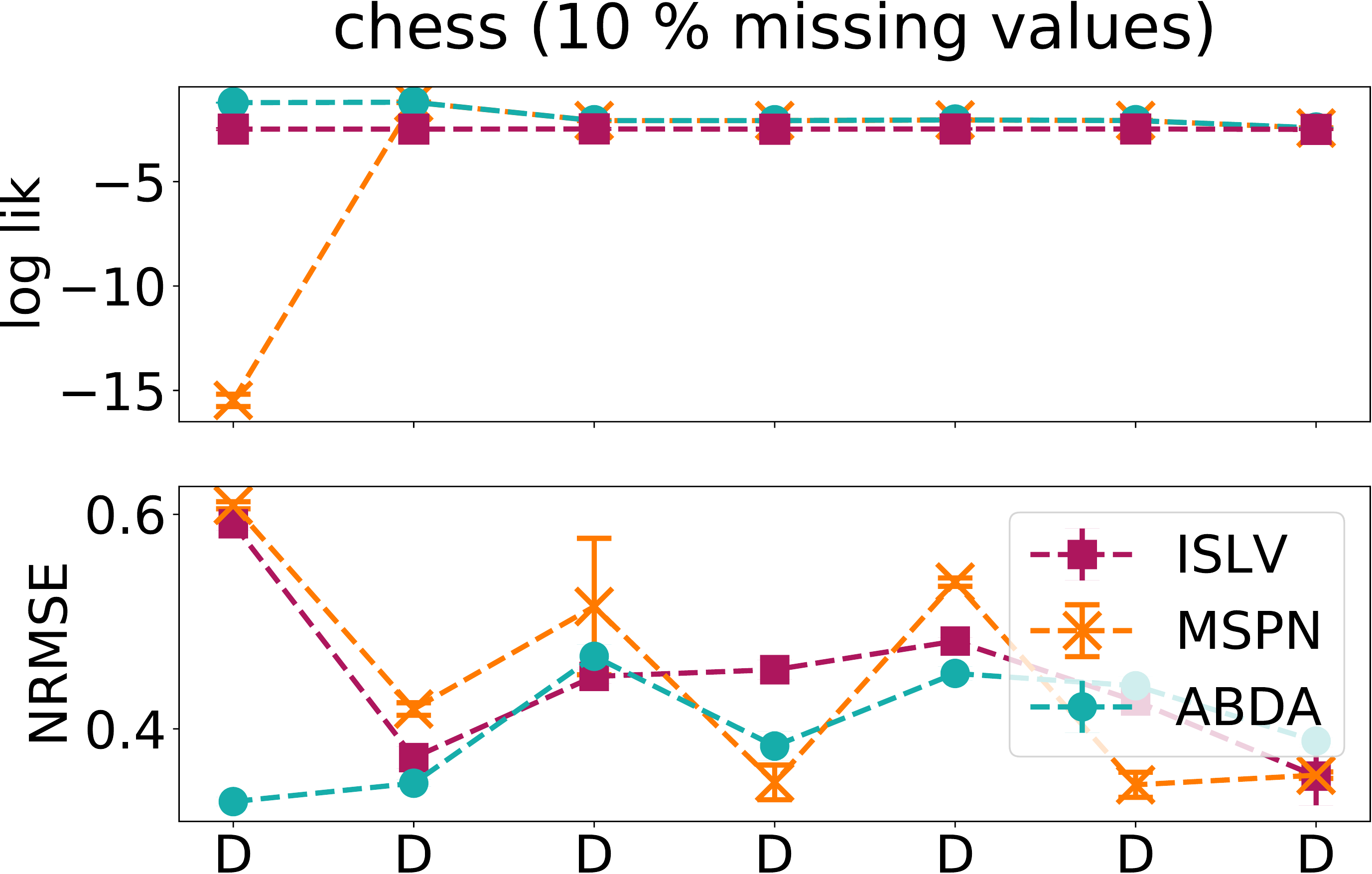}
        \caption{Chess}
        \label{fig:miss:chess}
    \end{subfigure}\hspace{10pt}
    \begin{subfigure}[t]{0.226\columnwidth}
        \includegraphics[width=1\columnwidth]{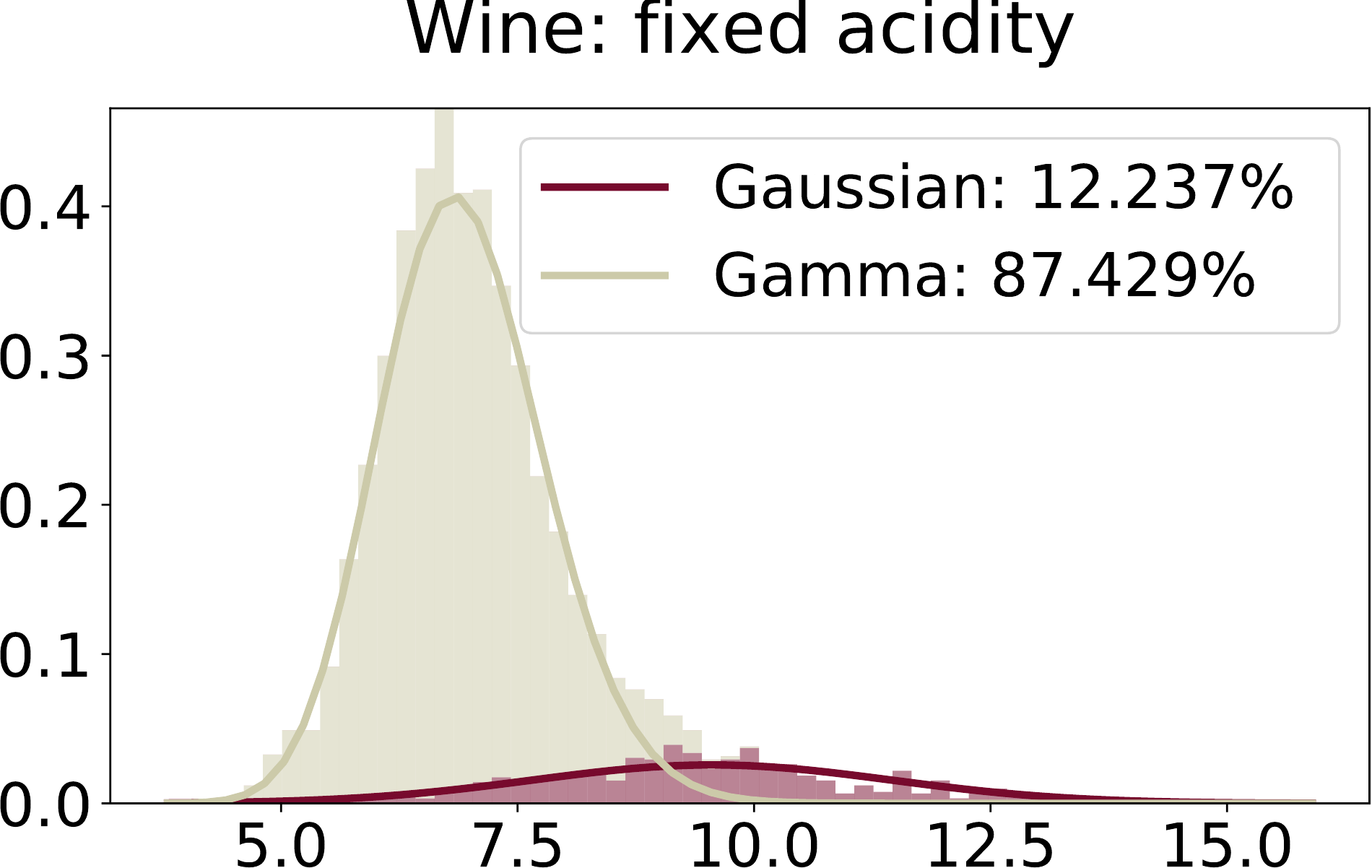}
        \caption{}
        \label{fig:imp1}
    \end{subfigure}\hspace{3pt}
    \begin{subfigure}[t]{0.22\columnwidth}
        \includegraphics[width=1\columnwidth]{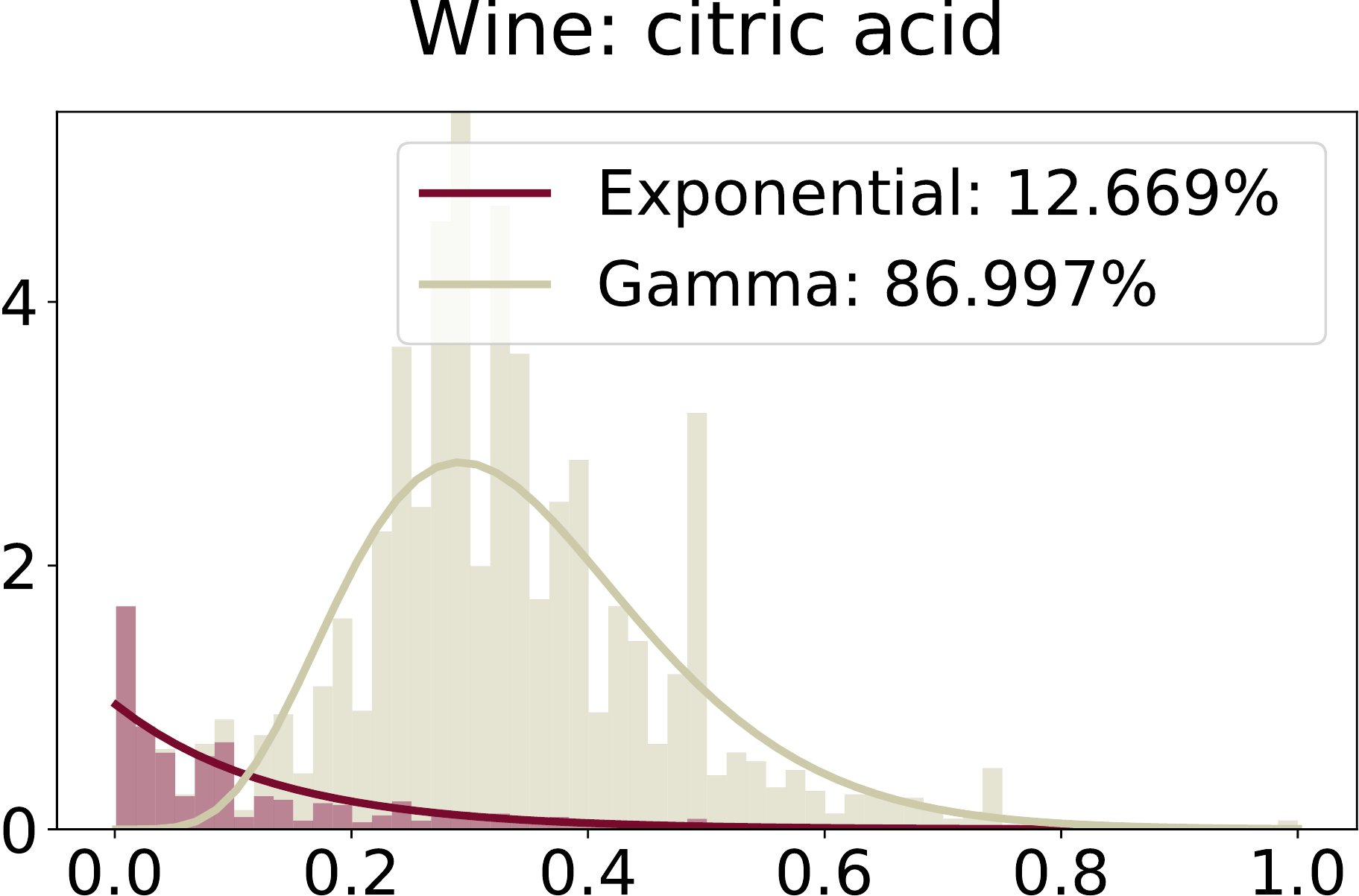}
        \caption{}
        \label{fig:imp2}
    \end{subfigure}  
    \caption{\textbf{Left} (a, b): Missing data estimation. Feature-wise test
      log-likelihood (top) and  NRMSE (bottom) for 10\%  of missing
      values for ABDA (cyan) ISLV (purple) and MSPN (orange). Features
      on the x-axis are labeled as ($\mathsf{D}$)iscrete or
      ($\mathsf{C}$)ontinuous. More plots in Appendix E.\\
    \textbf{Right} (c, d): Data exploration and dependency discovery with ABDA on the Wine quality dataset. 
    ABDA identifies the two modalities in the data induced by red and white wines, and
     extracts the following patterns:
    $5.8 \leq {\color{wwine}\mathsf{FixAcid}} < 8.1\wedge 0.2\leq {\color{wwine}\mathsf{CitAcid}} <
0.5$ and
$7.1 \leq {\color{rwine}\mathsf{FixAcid}} < 12.0\wedge 0.0\leq
{\color{rwine}\mathsf{CitAcid}} < 0.3$ ($\theta=0.9$). 
}
    \label{fig:miss}

\end{figure}

\subsubsection{(Q2) Density estimation and imputation}
We evaluate ABDA both in a \textit{transductive} scenario, where we aim to estimate (or even impute) the missing values in the data used for inference/training; and in an \textit{inductive} scenario, where we aim to estimate (impute) data that was not available during inference/training. 
We compare against ISLV~\cite{Valera2017a}, which directly accounts for data type (but not likelihood model) uncertainty, and MSPNs~\cite{Molina2017b}, to observe the effect of modeling uncertainty over the RV dependency structure via an SPN LV hierarchy.

 From ISLV and MSPN original works we select 12 real-world datasets differing w.r.t. size and feature heterogeneity.
 Appendix C reports detailed dataset information, while Appendix H contains additional experiments in the MSPN original setting.
Specifically, for the transductive setting, we randomly remove either 10\% or 50\% of the data entries,
reserving an additional 2\% as a validation set for hyperparameter tuning (when required), and repeating five times this process for robust evaluation. 
For the inductive scenario, we split the data into train, validation, and test (70\%, 10\%, and 20\% splits).

For ABDA and ISLV, we run 5000 iteratations of Gibbs sampling~\footnote{On the Adult dataset, ISLV did not converge in 72hr.}, discarding the first 4000 for burn-in.
We set for ISLV the number of latent factors to $\floor{D/2}$. 
We learn MSPNs with the same hyper-parameters as for ABDA structure learning, i.e., stopping to grow the network when the data to be split is less than 10\% of the dataset, while employing a grid search in $\{0.3, 0.5, 0.7\}$
for the RDC dependency test threshold.\footnote{In Appendix H we also evaluate ABDA and MSPN robustness to overparametrized structures.}

Tab.~\ref{tab:trans} reports the mean test-log likelihoods--evaluated on missing values in the transductive or on completely unseen test samples in the inductive cases--for all datasets. 
Here, we can see that ABDA outperforms both ISLVs and MSPNs in most cases for both scenarios. %
Moreover, since aggregated evaluations of heterogeneous likelihoods might be dominated by a subset features, we also report in Fig~\ref{fig:miss} the average test log-likelihood and the normalized root mean squared error (NRMSE) of the imputed missing values (normalized by the range of each RV separately) for each feature in the data. 
Here, we observe that ABDA is, in general, more accurate and robust across different features and data types than competitors.
 We finally remark that due to the piecewise approximation of the likelihood adopted by the MSPN, evaluations of the likelihood provided by this approach might be boosted by the fact that it renormalizes an infinite support distribution to a bounded one. 

\subsubsection{(Q3) Anomaly detection.}
We follow the \textit{unsupervised outlier detection} experimental setting in~\cite{Goldstein2016} to evaluate the ability of ABDA to detect anomalous samples on
a set of standard benchmarks.
As a qualitative example, we can observe in Fig~\ref{fig:od} that ABDA either clusters outliers together or relegates them to leaf distribution tails, assigning them low probabilities.
Tab.~\ref{tab:trans} compares, in terms of the mean AUC ROC, ABDA---for which we use the negative log-likelihood as the outlier score---with staple outlier detection methods like one-class SVMs
(1SVM)~\cite{Scholkopf2001},
 local outlier factor (LOF)~\cite{Breunig2000} and 
histogram-based outlier score (HBOS)~\cite{Goldstein2012}.

It is clearly visible that ABDA perform as good as---or even better---in most cases than
methods tailored for outlier detection and not being usable for other data analysis tasks. 
Refer to Appendix F for further experimental details and results.

\begin{figure*}[!t]
    \includegraphics[width=0.35\columnwidth]{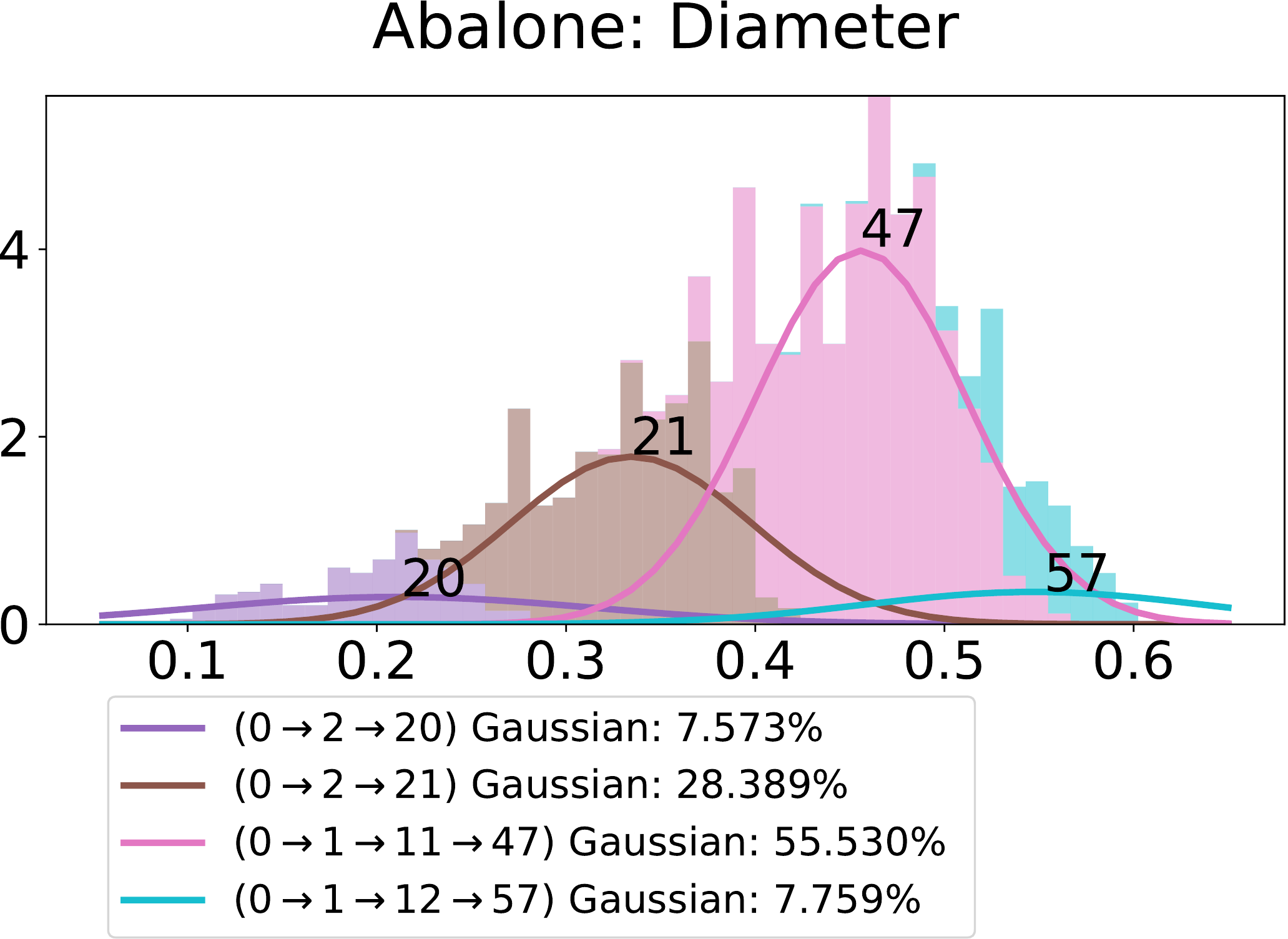}\hspace{15pt}
    \includegraphics[width=0.34\columnwidth]{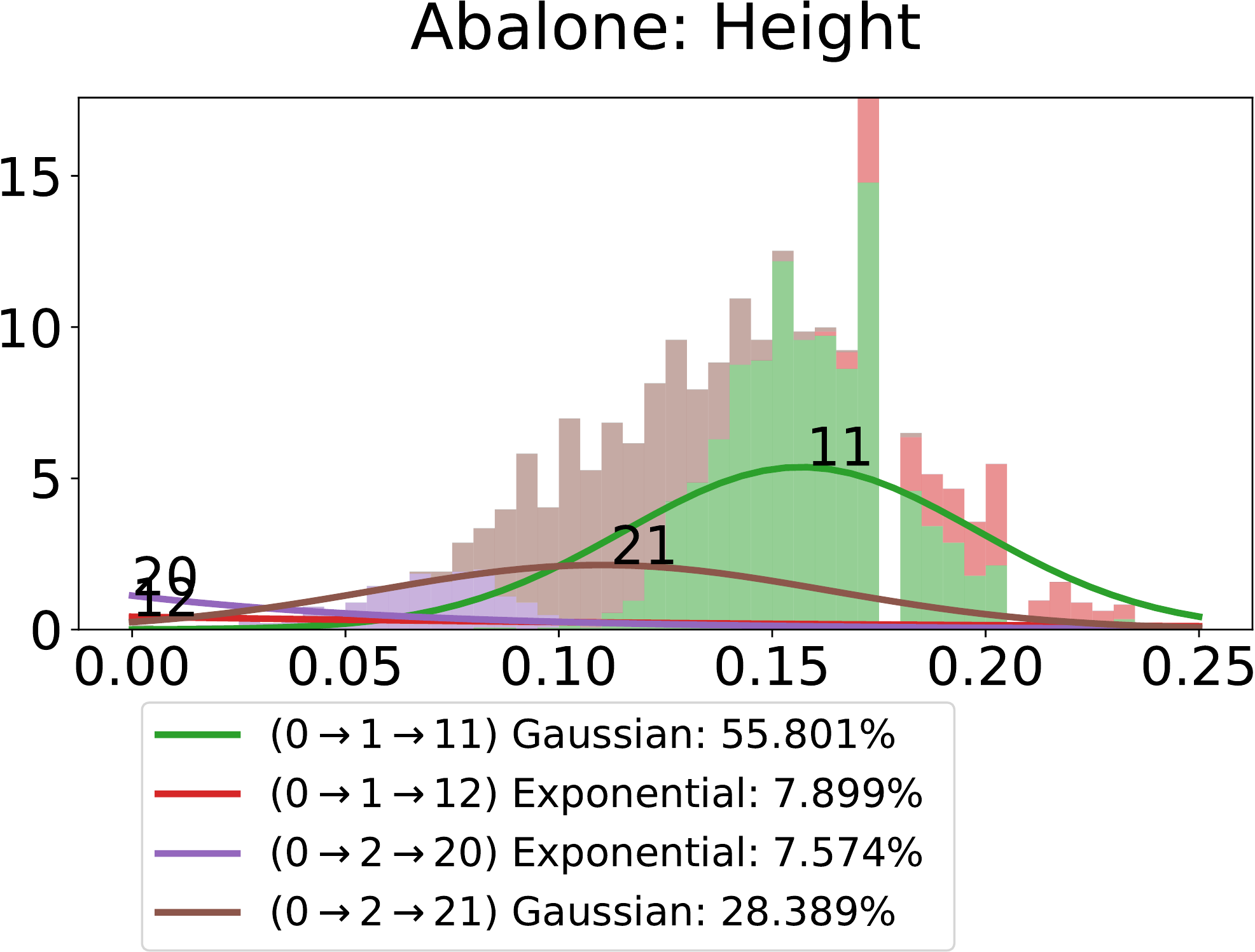}\hspace{8pt}\\[10pt]
    \includegraphics[width=0.36\columnwidth]{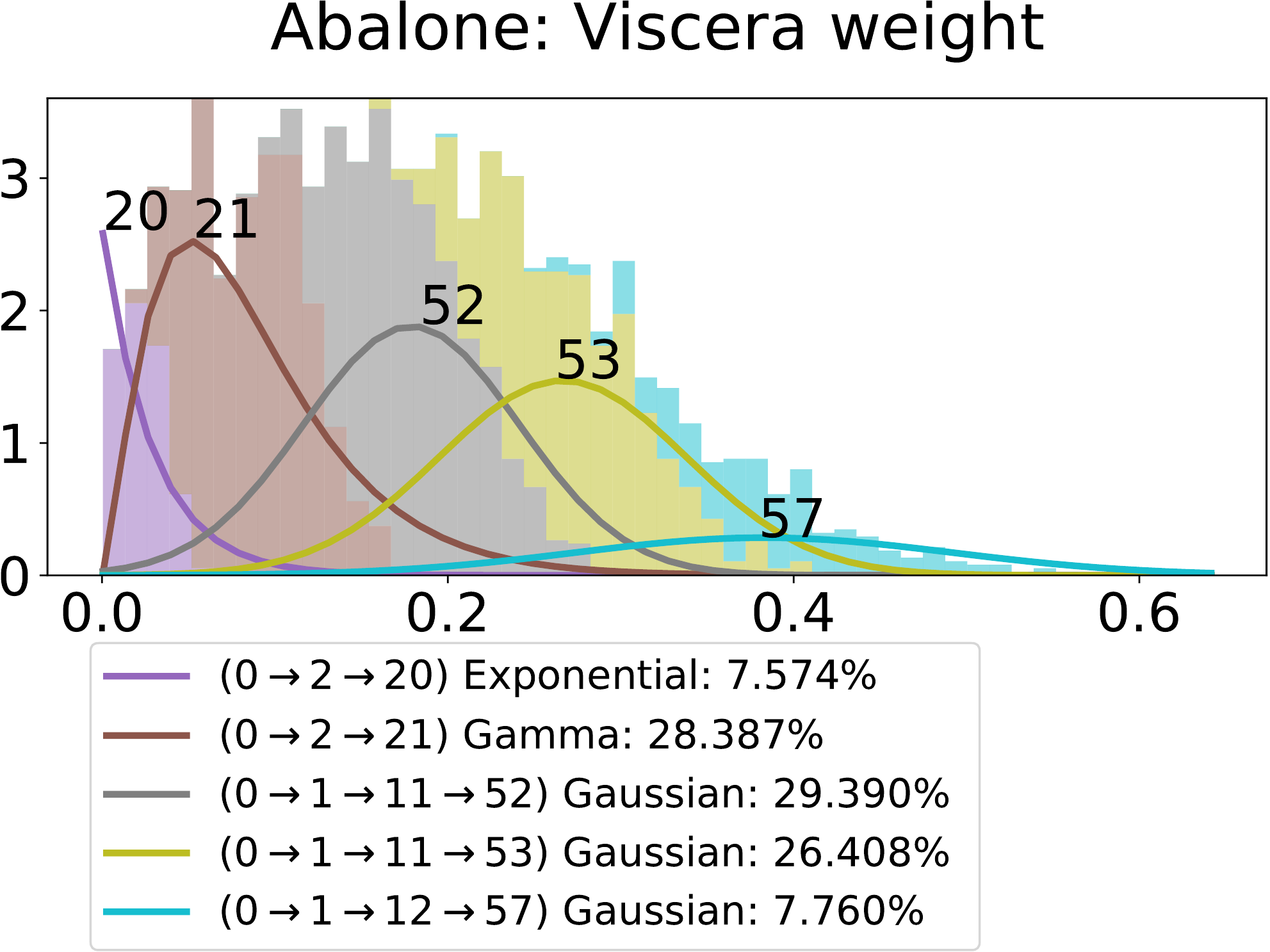}\hspace{17pt}
    \includegraphics[width=0.33\columnwidth]{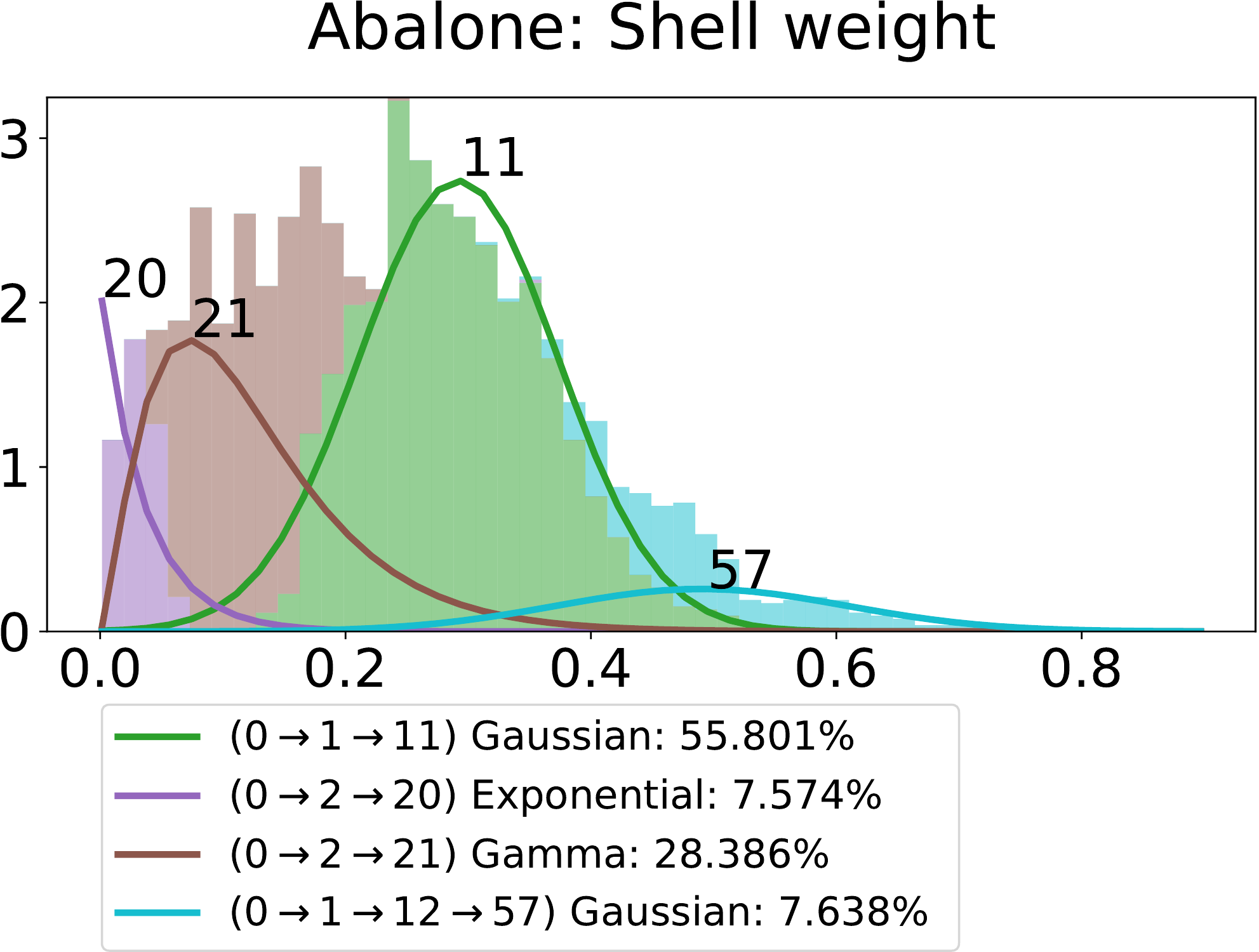}\hspace{5pt}
    \caption{Data exploration and pattern discovery on the Abalone dataset, 
    comprising physical measurements (e.g. diameter, height, weight) of different specimens. 
    Each colored density belongs to a unique partition $\Xb^{\Node}$ discovered by ABDA and here labeled by an integer indicating the corresponding node $\Node$ (0 indicates the root of the SPN). 
    Hierarchies over partitions are shown in the legend as the path connecting the root node to $\Node$ (e.g., $0\rightarrow 1\rightarrow 11\rightarrow 47$ indicates that the partition \#11 (green) is responsible for the green densities and it decomposes into the partitions \#52 (gray), \#53 (yellow) and \#47 (pink), highlighting different correlation patterns across feature intervals).
    For instance, $\mathcal{P}_{1}:\quad 0.088\leq {\color{abagreen}\mathsf{Height}}< 0.224\quad\wedge
    0.158\leq {\color{abagreen}\mathsf{ShellWeight}}< 0.425 \quad(supp(\mathcal{P}_{1})=0.507)$ is a pattern within such a partition, correlating the height and the shell weight features. 
    Additional patterns, and the other estimates for the dataset features, are reported in Appendix G.}
    \label{fig:parm-aba}
\end{figure*}

\subsubsection{(Q4) Dependency discovery.}

Finally, we illustrate how ABDA may be used to find underlying dependency structure in the data on the Wine and Abalone datasets as use cases.
By performing marginal inference for each feature, and by collapsing the resulting deep mixture distribution 
into a shallow one, with ABDA we can recover the data modes and reason about the likelihood distributions associated to them.

As an example, ABDA is able to discern the two modes in the Wine data which correspond to the two types of wine, red and white wines, information not given as input to ABDA (Figs~\ref{fig:imp1}~\ref{fig:imp2}). 
Moreover, ABDA is also able to assign to the two modes accurate and meaningful likelihood models: 
Gamma and Exponential distributions are generally captured for the features fixed and citric acidity, since they are a ratio and indeed follow a positive distribution, while being more skewed and decaying than a Gaussian, employed for the fixed acidity of red wines.  
Note that since ABDA partitions the data into white and red wines
sub-populations, it allows us to reason about statistical dependencies
in  the data in the form of simple conjunctive patterns (see caption of Fig~\ref{fig:miss}), as discussed in the previous Section. 
Here, we observe an anti-correlation between fixed and citric acidity: as the former increases the latter tends to zero.

A more involved analysis is carried on the Abalone dataset and summarized in Fig~\ref{fig:parm-aba}.
There, retrieved data partitions clearly highlight correlations across features and samples of the data.
For instance, it is possible to see how abalone samples differing by weight, height and diameters form neat sub-populations in the data.
See Appendix G in the supplementary for a detailed discussion and more results.

\section{Conclusions}

Towards the goal of fully automating exploratory data analysis via density estimation and probabilistic inference, we introduced Automatic Bayesian Density Analysis (ABDA). 
It automates both data modeling and selection of adequate likelihood models for estimating densities via joint, robust and accurate Bayesian inference. 
We found that the inferred structures are able of accurately analyzing complex data and discovering the data types, the likelihood models and feature interactions. Overall, it outperformed state-of-the-art in different tasks and scenarios in which domain experts would perform exploratory data analysis by hand.

ABDA opens many interesting avenues for future work. 
First, we aim at inferring also prior models in an automatic way; abstracting from inference implementation details by integrating probabilistic programming into ABDA; and casting the LV structure learning as nonparametric Bayesian inference. 
Second, we plan on integrating ABDA in a full pipeline for exploratory data analysis, where probabilistic and logical reasoning can be performed over the extracted densities and patterns to generate human-readable reports, and be treated as input into other ML tasks. 

\scalebox{0.01}{quare id inferam fortasse requiris}

\subsubsection*{Acknowledgements} 
We thank the anonymous reviewers and Wittawat Jitkrittum for their valuable feedback. IV is funded by the MPG Minerva Fast Track Program. This work has benefited from the DFG project CAML (KE 1686/3-1), as part of the SPP 1999, and from the BMBF project MADESI (01IS18043B). 
This project has received funding from the European Union's Horizon 2020 research and innovation programme under the Marie Sk\l{}odowska-Curie Grant Agreement No.~797223.

\appendix

\section{A. Gibbs sampling}

\begin{table}[!h]
\caption{Average time (in seconds) per iteration for the best ABDA and
  ISLV models on the density estimation datasets.}
    \label{tab:gibbs-times}
\vspace{1pt}
\setlength{\tabcolsep}{2pt}
\small
    \centering
    \begin{tabular}{r r r r r r}
    \toprule
      &\textbf{ISLV} &\textbf{ABDA}&&\textbf{ISLV}&\textbf{ABDA}\\
      & (\emph{s/iter})& (\emph{s/iter})&& (\emph{s/iter})& (\emph{s/iter})\\
      \midrule
      \textsf{Abalone} & 5.21 & 0.20 &
      \textsf{Australian} & 0.89& 0.04\\
      \textsf{Autism}& 13.56& 0.25&
      \textsf{Breast} & 1.30&0.05\\
          \textsf{Chess} & 40.6&0.36&
          \textsf{Crx} & 1.10& 0.04\\
            \textsf{Dermatology}  & 2.01 & 0.14&
            \textsf{Diabetes} & 0.75&  0.03\\
               \textsf{German} & 4.05& 0.08&
         \textsf{Student} & 1.81& 0.06\\
          \textsf{Wine}  &7.05& 0.15\\
      \midrule
    \end{tabular}
  \end{table}
  
\paragraph{A1 Implementation and times}
The proposed Gibbs sampling scheme, sketched in Algorithm~\ref{algo:gibbs} is simple and efficient, scaling
linearly in the size of $\Xb$ and the number of sum nodes in $\SPN$.
Note that this time complexity does not depend on the domain size of
discrete RVs, unlike ISLV~\cite{Valera2017a} (which also has to
perform a costly matrix inversion).
Practically, exploiting the  underlying SPN to efficiently condition
on the LV hierarchy---allowing to keep updates for involved leaf
distributions--- \textit{enormously speeds Gibbs sampling} in ABDA,
also on continuous data.
As a reference, see the average times (seconds) per iterations of
ISLV and ABDA in Table~\ref{tab:gibbs-times}. 

Moreover, ABDA lends itself to be implemented parsimoniously\footnote{We will
  publicly release all our code to run the model and reproduce experiments upon acceptance.} if one
has access to sampling routines for the parametric models employed in
the leaf distributions.
See Appendix B for a detailed specification of the likelihood and
posterior parametric forms
involved in the experiments.

\paragraph{A2. Rao-Blackwellised Gibbs sampler}
In all our experiments, we observed it converging quickly to high
likelihood posterior solutions.
We use likelihood traceplots~\cite{Cowles1996} to track the sampler
convergence, noting that at most 3000 samples for the largest
datasets, are generally are needed.

Nevertheless, to further improve convergence, one may devise a Rao-Blackwellised version~\cite{Murphy2012} by collapsing out several parameters in ABDAs. 
For instance, leaf distribution parameters $\boldsymbol{\eta}_{j\ell}^{d}$ can be marginalized when sampling assignments to the LVs $\mathbf{Z}$. Indeed, one can draw them from:

\begin{equation}
p(Z_{n}^{{h}}=c|\mathbf{x}_{n}, \{Z_{n}^p\}_{p\in
  \mathsf{anc}(h)},\boldsymbol\Omega)\propto
\omega_{hc}\mathcal{S}_{c}(\mathbf{x}_{n}|\mathbf{x}_{\setminus n}, \boldsymbol\Omega)\label{eq:samp-z-rao}
\end{equation}

where $\mathcal{S}_{c}(\mathbf{x}_{n}|\mathbf{x}_{\setminus n}, \boldsymbol\Omega)$ denotes the posterior predictive distribution of sample $\mathbf{x}_{n}$ given all remaining samples $\mathbf{x}_{\setminus n}$ through SPN $\SPN$ equipped with weights $\boldsymbol\Omega$. Such a quantity can be computed again in time linear in the size of $\SPN$.

\begin{algorithm}[!t]
\caption{Gibbs sampling inference in ABDA}
\small
\label{algo:gibbs}
\begin{algorithmic}[1]
  \REQUIRE data matrix $\Xb$,  an SPN $\mathcal{S}$,
  $\{\boldsymbol\eta_{j,\ell}^{d}\}_{\Leaf_j\in\mathcal{S},\ell\in\mathcal{L}^{d}},$
  $ \{\mathbf{Z}_{n}\}_{n=1}^{N},
  \boldsymbol\Omega, \{\mathbf{w}_{k}^{d}\}_{L_{k}\in\SPN,d=1}^{D},$$ \{s_{j,n}^{d}\}_{\Leaf_{j}\in\SPN,d=1\dots D, n=1\dots N}$ and $I$ the number of iterations\\[2pt]
  \STATE $\mathcal{D}\leftarrow\emptyset$
  \STATE Initialize $\{\boldsymbol\eta_{j,\ell}^{d}\}_{\Leaf_{j}\in\SPN,\ell\in\mathcal{L}^{d}}, \{\mathbf{Z}_{n}\}_{n=1}^{N},
  \boldsymbol\Omega, \{\mathbf{w}_{j}^{d}\}_{\Leaf_{j}\in\SPN,d=1\dots
    D}, $
  $\{s_{j,n}^{d}\}_{\Leaf_{j}\in\SPN,d=1\dots D, n=1\dots N}$
  \FOR{$it\in 1,\dots,I$}
  \FOR{$n \in 1,\dots,N$}
    \STATE Sample $\mathbf{Z}_{n}$, $\{s_{j,n}^{d}\}_{\Leaf_{j}\in\SPN,d=1\dots D, n=1\dots N}$ given $\mathbf{X}, \boldsymbol\Omega,   \{\boldsymbol\eta_{j,\ell}^{d}\}$
  \ENDFOR
  \FOR{$\Leaf_j \in \SPN$, with scope $d$}
  \FOR{$\ell\in\mathcal{L}^{d}$}
  \STATE Sample $\boldsymbol\eta_{j,\ell}^{d}$ given $\mathbf{X}, 
  \{\mathbf{Z}_{n}\}_{n=1}^{N}, \{s_{j,n}^{d}\}_{\Leaf_{j}\in\SPN,d=1\dots D, n=1\dots N}$  
  \ENDFOR
  \FOR{$d \in 1,\dots,D$}
  \STATE Sample  $\mathbf{w}_{j}^{d}$ given 
  $\{s_{j,n}^{d}\}_{\Leaf_{j}\in\SPN,d=1\dots D, n=1\dots N}$ 
  \ENDFOR
    \ENDFOR
  \STATE Sample $\boldsymbol\Omega$ given $\{\mathbf{Z}_{n}\}_{n=1}^{N}$
  \IF{$it > \mathsf{burn-in}$}
  \STATE $\mathcal{D}\leftarrow\mathcal{D}\cup\{\{\boldsymbol\eta_{j,\ell}^{d}\}_{\Leaf_j\in\mathcal{S},\ell\in\mathcal{L}^{d}}, \{\mathbf{Z}_{n}\}_{n=1}^{N},
  \boldsymbol\Omega, \{\mathbf{w}_{k}^{d}\}_{\Leaf_{k}\in\SPN},$
  $ \{s_{j,n}^{d}\}_{\Leaf_{j}\in\SPN,d=1\dots D, n=1\dots N}\}$
  \ENDIF
  \ENDFOR
 \hspace*{-19pt}
 \textbf{ Output}:  $\mathcal{D}$
\end{algorithmic}
\end{algorithm}

\section{B. Likelihood and prior models}

The parametric forms used in our experiments for the likelihood
models, and their corresponding priors, are shown below, w.r.t. the
statistical data type involved: $\mathsf{REAL}$-valued data,
$\mathsf{POS}$itive real-valued data, $\mathsf{NUM}$erical, 
$\mathsf{NOM}$inal and $\mathsf{BIN}$ary.

Note that, even if in our experiments we focused on likelihood
dictionaries within the exponential family for simplicity,
\textit{ABDA can  be readily extended to any likelihood model}.
E.g., for ordinal data one can just plug the ordinal \emph{probit model}~\cite{Valera2017a} in
in the dictionary.
Of course, introducing a prior distribution that does not allow to
exploit conjugacy, but that could represent valuable prior knowledge about a
distribution, would need to derive an approximate sampling routines
for the involved likelihood model.\bigskip


\noindent\textbf{[\textit{real-valued}]}\\[-15pt]
\begin{description}
\item[\textbf{Gaussian}] $\mathcal{N}(\mu,\sigma^2)$\vspace{-5pt}
\begin{itemize} 
    \item \textbf{prior:} $p(\mu, \sigma^{2}|m_{0}, V_{0},
\alpha_{0},\beta_{0})=\mathcal{N}(\mu|m_{0},\sigma^{2}V_{0})IGamma(\sigma^{2}|\alpha_0,\beta_0)$
    \item \textbf{posterior:} $p(\mu, \sigma^{2}|m_{0}, V_{0},
\alpha_{0},\beta_{0})=\mathcal{N}(\mu|\frac{V_0^{-1}m_{0}+N\bar{x}}{V_0^{-1}+N},V_0^{-1}+N)\times \\ IGamma(\alpha_0+N/2,\beta_0+\frac{1}{2}\sum_{i=1}^{N}(x_{i}-\bar{x})^{2}+\frac{V_0^{-1}N(\bar{x}-m_{0})^{2}}{2(V_0^{-1}+N)}) $
\end{itemize}
\end{description}

\noindent\textbf{[\textit{positive real-valued}]}\\[-15pt] 

\begin{description}
\item[\textbf{Gamma}] with fixed $\alpha$, $Gamma(\alpha,\beta)$\vspace{-5pt}
\begin{itemize}
    \item \textbf{prior:} $p(\beta|\alpha_0,\beta_0)=Gamma(\alpha_0,\beta_0)$
    \item \textbf{posterior:} $p(\beta|\mathbf{x},\alpha_0,\beta_0)=\\Gamma(\alpha_n=\alpha_{0}+N\alpha,\beta_n=\beta_{0}+\sum_{i=1}^{N}x_{i})$
\end{itemize}
\end{description}

\begin{description}
\item[\textbf{Exponential}] $Exponential(\lambda)$\vspace{-5pt}
\begin{itemize}
    \item \textbf{prior:} $p(\lambda|\alpha_0,\beta_0)=Gamma(\alpha_0,\beta_0)$
    \item \textbf{posterior:} $p(\lambda|\mathbf{x},\alpha_0,\beta_0)=\\Gamma(\alpha_n=\alpha_{0}+N\alpha,\beta_n=\beta_{0}+\sum_{i=1}^{N}x_{i})$
\end{itemize}
\end{description}

\noindent\textbf{[\textit{nominal}]}\\[-15pt] 

\begin{description}
\item[\textbf{Categorical}] $Cat(\{\theta_{i}\}_{i=1}^{k})$\vspace{-5pt}
\begin{itemize}
    \item \textbf{prior:} $p(\{\theta_{i}\}_{i=1}^{k}|\boldsymbol\alpha_{i})=Dirichlet(\{\theta_{i}\}_{i=1}^{k}|\boldsymbol\alpha_{i})$
    \item \textbf{posterior:} $p(\{\theta_{i}\}_{i=1}^{k}|\mathbf{x}, \boldsymbol\alpha_{i})=Dirichlet(\{\theta_{i}\}_{i=1}^{k}|\boldsymbol\alpha_{i} + N_{i})$
\end{itemize}
\end{description}

\noindent\textbf{[\textit{numerical}]}\\[-15pt] 

\begin{description}
\item[\textbf{Poisson}]  $Poisson(\lambda)$\vspace{-5pt}
\begin{itemize}
    \item \textbf{prior:} $p(\lambda|\alpha_0,\beta_{0})=Gamma(\lambda|\alpha_{0}, \beta_{0})$
    \item \textbf{posterior:} $p(\lambda|\mathbf{x},\alpha_0,\beta_{0})=Gamma(\lambda|\alpha_{n}=\alpha_{0}+\sum_{i=1}^{N}x_{i}, \beta_{n}=\beta_{0}+N)$
\end{itemize}
\item[\textbf{Geometric}] $Geometric(\theta)$\vspace{-5pt}
\begin{itemize}
    \item \textbf{prior:} $p(\theta|\alpha_{0},\beta_{0})=Beta(\theta|\alpha_{0},\beta_{0})$
    \item \textbf{posterior:} $p(\theta_{i}|\mathbf{x}, \alpha_{0},
      \beta_{0})=Beta(\theta|\alpha_{0} + N,
      \beta_{0} - N + \sum_{i=1}^{N}\mathbf{x}_{i})$
\end{itemize}
\end{description}

\noindent\textbf{[\textit{binary}]}\\[-15pt] 
\begin{description}
\item[\textbf{Bernoulli}]  $Ber(\theta)$\vspace{-5pt}
\begin{itemize}
    \item \textbf{prior:} $p(\theta|\alpha_{0},\beta_{0})=Beta(\theta|\alpha_{0},\beta_{0})$
    \item \textbf{posterior:} $p(\theta_{i}|\mathbf{x}, \alpha_{0},
      \beta_{0})=Beta(\theta|\alpha_{0} + \sum_{i=1}^{N}\mathbf{x}_{i},
      \beta_{0} + N - \sum_{i=1}^{N}\mathbf{x}_{i})$
\end{itemize}
\end{description}


\begin{figure*}[!t]
    \includegraphics[width=0.9\textwidth]{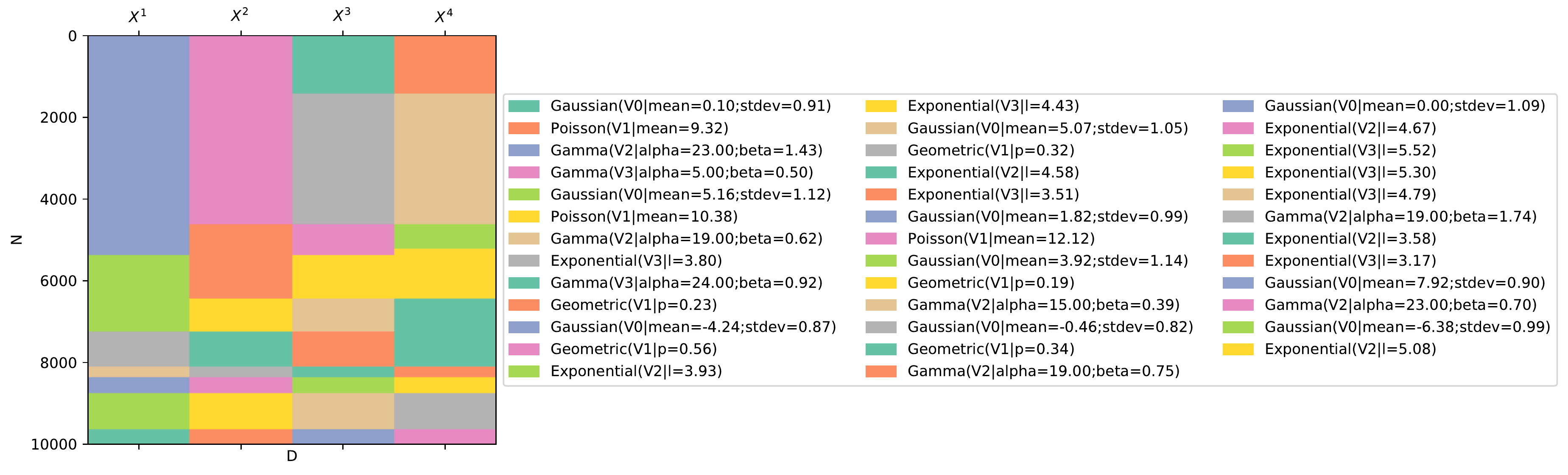}\\
    \includegraphics[width=0.24\textwidth]{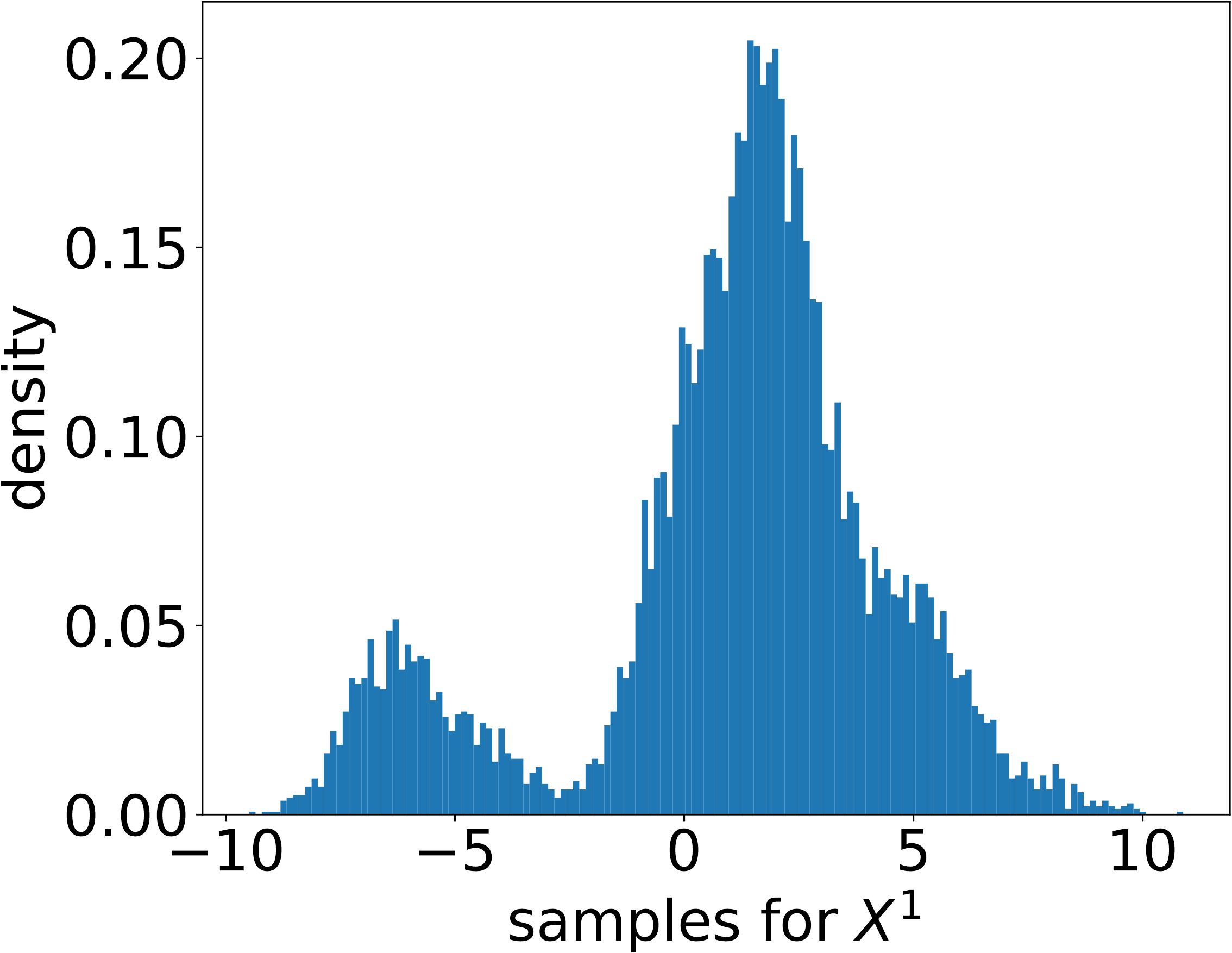}\hfill
    \includegraphics[width=0.24\textwidth]{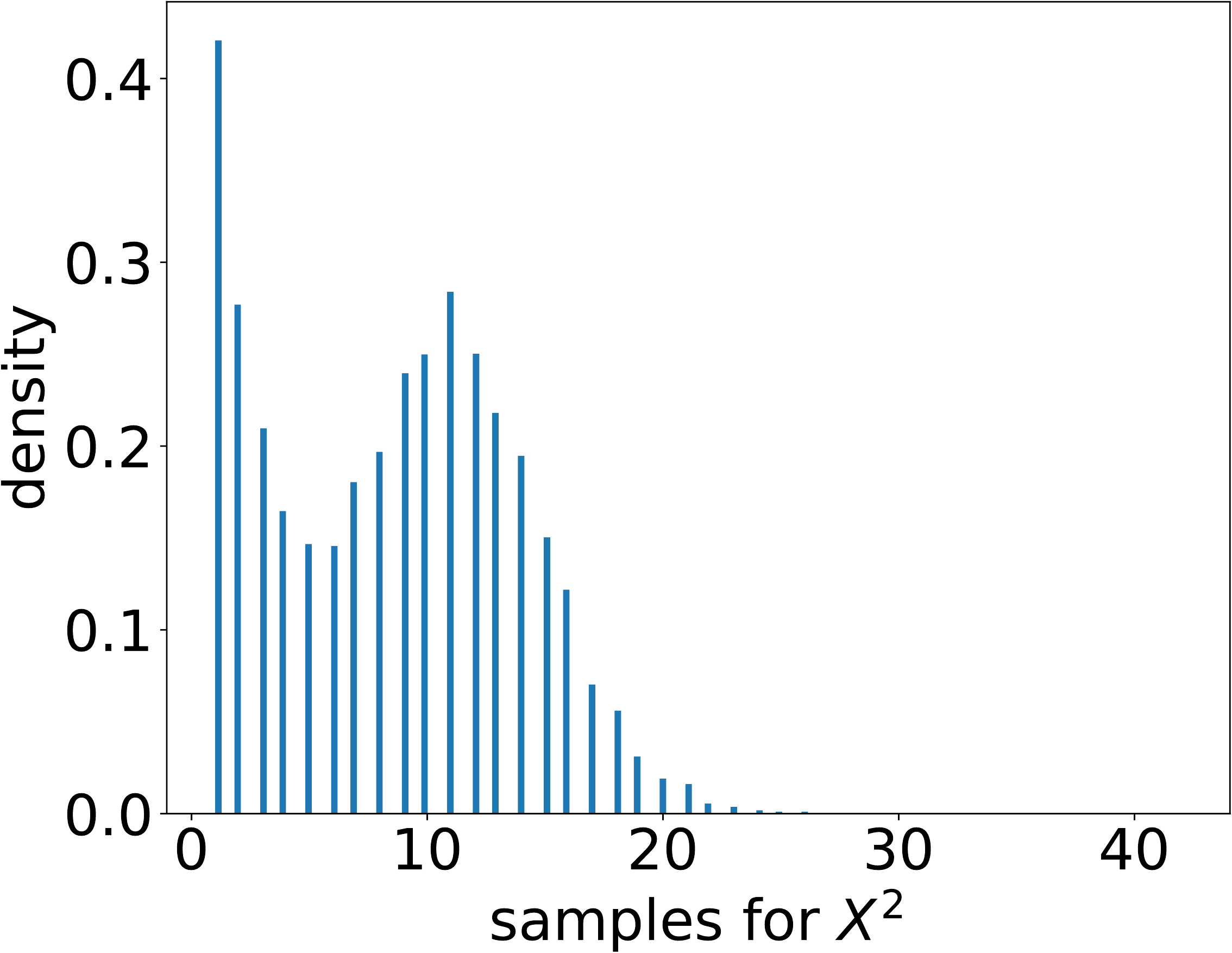}\hfill
    \includegraphics[width=0.24\textwidth]{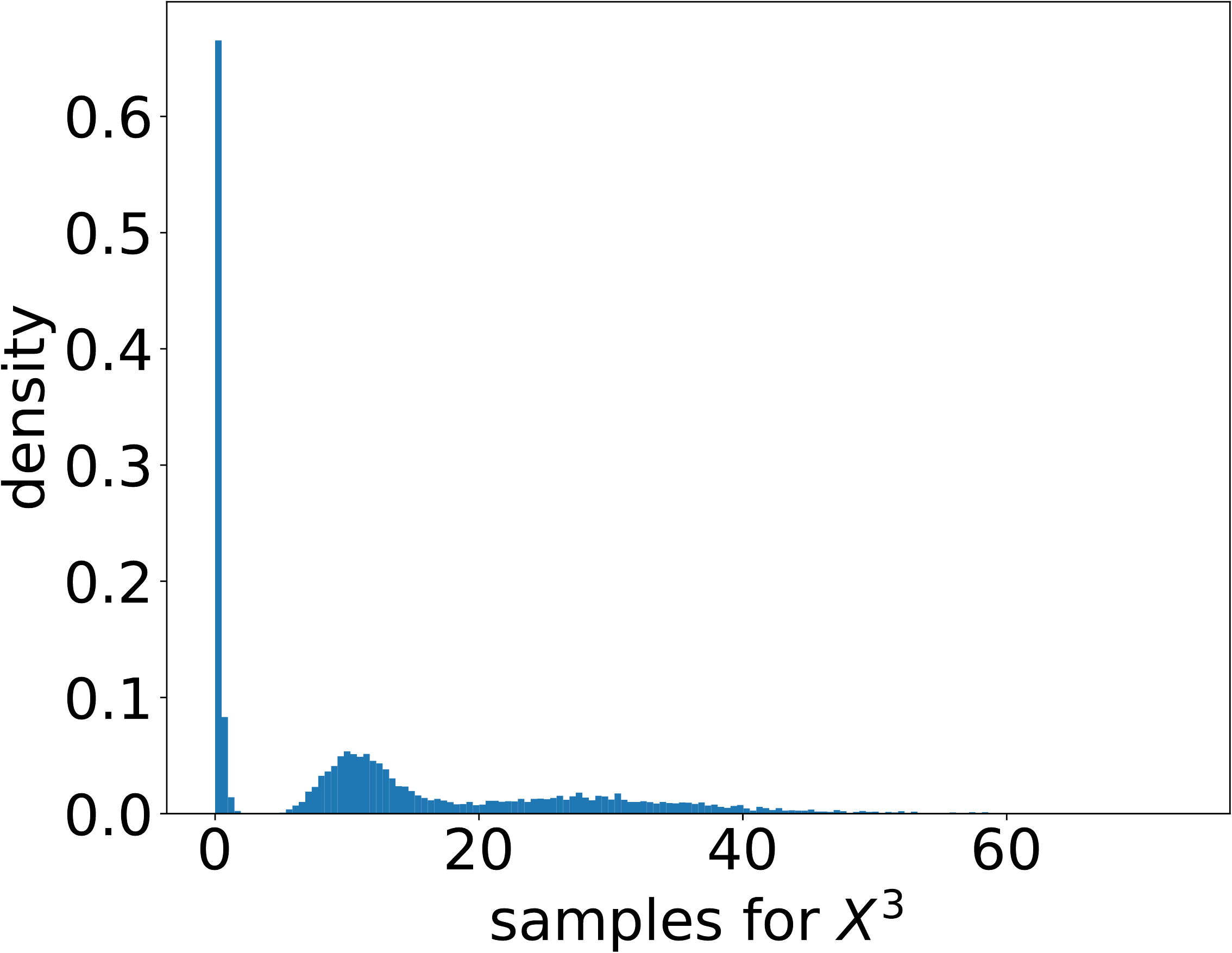}\hfill
    \includegraphics[width=0.24\textwidth]{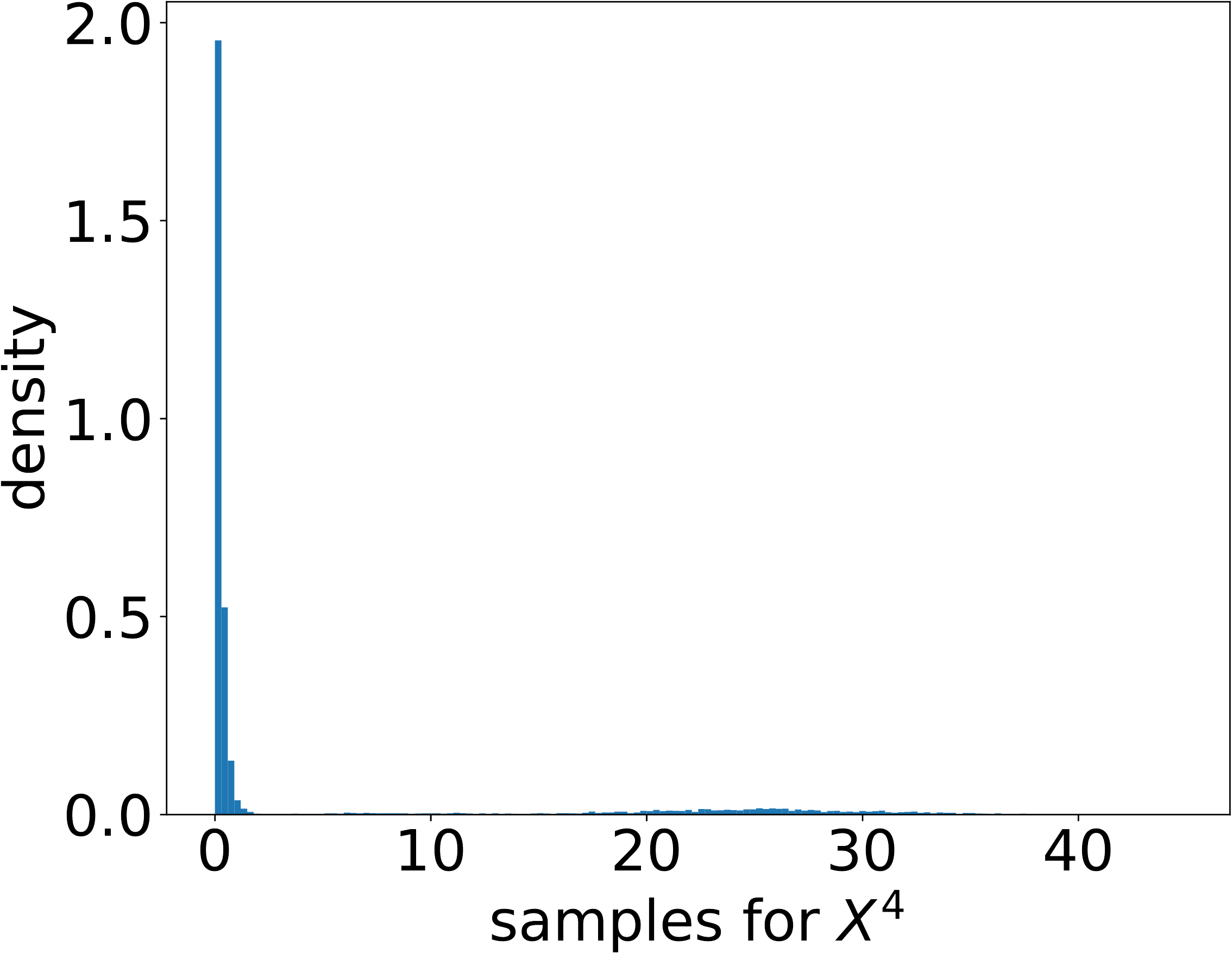}
    \caption{Synthetic data generation. (Top) An example of a partitioning for a synthetic dataset of $N=10000$ samples and $D=4$ features, where samples have been re-ordered to have contiguous partitions for the sake of visualization.
    Each partition is associated with a parametric likelihood model labeled by a color, instantiated with a set of parameters randomly drawn according to a sensible prior (see Appendix B).
    (Bottom) Histogram visualizations for the marginals of the $4$ features generated.}
    \label{fig:partitioning}
\end{figure*}

\section{C. Synthetic data experiments}

\paragraph{C1. Synthetic data generation}
Here we describe in detail the process we adopted for generating the 90 datasets employed in our controlled experiments.
As stated in the main article, we consider an increasing number of samples $N\in\{2000, 5000, 10000\}$ and features $D\in\{4, 8, 16\}$ and for each combination of the two we generate 10 independent datasets.
Later on, we split them into training, validation and test partitions (70\%, 10\%, 20\% splits).

We generate each dataset by mimicing the structure learning process of an SPN.
Please note that in such a way\textit{ we are encompassing highly expressive and complex joint distributions} that might have generated the data (while retaining control over their statistical dependencies, types and likelihood models.
Indeed, SPNs have been demonstrated to capture linear~\cite{Molina2017a} and highly non-linear correlations~\cite{Molina2017b,Vergari2018} in the data, being also able to model constrained random vectors, as ones drawn from the simplex (see. \cite{Molina2017b}).

Each dataset is generated in the following way:
For every feature, we fix a randomly selected type among real, positive, discrete numerical or nominal.
To randomly create a ground truth generative model, i.e., an SPN---denoted as $\SPN'$---we simulate a stochastic guillotine partitioning of a fictitious $N\times D$ data matrix.
Specifically, we follow a LearnSPN-like~\cite{Gens2013,Vergari2015} structure learning scheme in which columns and rows of this matrix are clustered together, in this case in a random fashion.

At each iteration of the algorithm, it decides to try to split the columns or rows proportionally to a probability  $\theta_{\mathsf{split}}=0.8$.
For a tentative column split, the likelihood of a column to be assigned to one of two clusters is drawn from a $\mathsf{Beta}(a=4, b=5)$. In case all columns are assigned to a single cluster, no column split is performed and the process moves to the next iteration.
Concerning row splits, on the other hand, each row is randomly assigned to one of two clusters with a probability drawn from $\mathsf{Beta}(a=4, b=5)$.
We alternate these splitting processes until we reach a matrix partition with less than 10\% of N number of rows. We assign these final partitions to univariate leave nodes in the spn.
For each leaf node, we then randomly select a univariate parametric distribution valid for the type associated with the feature $d$, as listed in the previous Section.

An example of a synthetically generated dataset for $N=10000$ samples and $D=4$ features, and its random partitioning, is shown in Fig.~\ref{fig:partitioning}.
There, samples have been ordered in the data matrix (after performing bi-clustering) to enhance the visualization of contiguous partitions.

We employ the following priors to draw the corresponding parameters for each distribution:
For Gaussian distributions we employ a $\mathsf{Normal-Inverse-Gamma}(\mu=0,V=30,a=10,b=10)$ for the mean and variance,
we draw the shape parameter of the Gamma distributions from a $\mathsf{Uniform}(5, 25)$ involved and their scale parameters from a $\mathsf{Gamma}(a=10, b=10)$ prior, while for Exponential distributions we randomly select the rate from a $\mathsf{Gamma}(a=20, b=5)$.
For the discrete data, we draw the number of categories of a Categorical from a discrete $\mathsf{Uniform}(5, 15)$. We then draw the corresponding probabilities from an equally sized and symmetric $\mathsf{Dirichlet}(\alpha=10)$; the mean parameter of Poissons is drawn from a $\mathsf{Gamma}(a=100, b=10)$.

\paragraph{C2. ABDA inference}
We perform inference on ABDA for each synthetic dataset. 
We learn the LV structure by letting the RDC independence threshold parameter in $\{0.1, 0.3, 0.5\}$ while fixing to $10\%$  of $N$ the minimum number of samples to continue the partitioning process during SPN structure learning.
We select then the best model w.r.t. the log-likelihood scored on a validation set.
Then we run the Gibbs sampler for 3000 iterations discarding the first 2000 samples for burn-in.

For the log-likelihoods in Fig 2a, we estimate them as the mean log-likelihood computed by ABDA over the test samples and averaged across the last 1000 samples of the Gibbs chain.

For recovering the global weights measuring uncertainty over the likelihood models (or statistical data types), we average over the predicted arrays belonging to the last 1000 samples of Gibbs and perform the cosine similarity between these vector and the ground truth one.

For the hard predictions, as shown in the confusion matrixes in Fig.2, we select, from each vector, the likelihood model (or statistical data type) scoring the highest weight in the weight vector.

\newpage
\section{D. Real-world datasets}

\paragraph{D1. Density estimation and imputation datasets.}
For our experiments we use some real-world datasets from the UCI\cite{dheeru2017uci} repository. The number of instances and features are what we used for the models after preprocessing.
We took the datasets as available from~\cite{Molina2017b} and~\cite{Valera2017b} and put them in the same tabular formalism where we labeled each feature to be either continuous or discrete.
We removed binary features from them and we either randomly selected a certain percentage of missing values for the transductive case or we randomly split them into train, validation and test sets for the inductive case (see main text).

\textbf{Abalone} is a dataset of abalone in Tasmania that includes different physical measurements. It contains 4177 instances, 9 features and no missing values. 
\textbf{Adult} is a "Census Income" dataset used to predict low or high income, it contains 32561 instances and 13 features.
\textbf{Australia} contains information about credit card applications in Australia, with 690 and 10 features.
\textbf{Autism} has data about the autistic spectrum disorder screening in adults. It contains 3521 instances and 25 features.
\textbf{Breast} describes 10 attributes about 681 instances of two types of breast cancer.
\textbf{Chess} is a chess endgame database, with 28056 and 7 features.
\textbf{Crx} is a dataset of credit card applications, with 651 instances and 11 attributes.
\textbf{Dermatology} provides 366 instances and 34 about the Eryhemato-Squamous dermatological disease.
\textbf{Diabetes} is a dataset of diabetes patient records, with 768 instances and 8 features.
\textbf{German} is a classification dataset of people described by a set of attributes as good or bad credit risks. It has 1000 instances and 17 attributes.
\textbf{Student} is a dataset of student performance in secondary education. With 395 instances and 20 features.
\textbf{Wine} is a dataset about wine quality based on physicochemical tests. It contains 6497 instances, with 12 features.

\paragraph{D2. Outlier detection datasets.}
The datasets used for anomaly detection come from \cite{Goldstein2016}. 
\textbf{Thyroid} is a dataset of thyroid diseases provided by the Garavan Institute of Sydney, Australia. It contains 6916 instances, 21 features, and 250 outliers.
\textbf{Letter} contains classification data for 26 capital letters of the English alphabet. It contains 1600 instances, 32 features, and 100 outliers.
\textbf{Pen-*} are datasets about pen-based handwritten digit recognition. Pen-global contains 809 instances, 16 features, and 90 outliers. Pen-local contains 6724, 16 features, and 10 outliers.
\textbf{Satellite} is a dataset that contains multi-spectral values of pixels in 3x3 neighborhoods in a satellite image and a label for the central pixel. It contains 5100 instances, 36 features, and 75 outliers. 
\textbf{Shuttle} is a dataset about the space shuttle with 46464 instances, 9 features and 878 outliers.
\textbf{Speech} is a speech recognition dataset with 3686 instances, 400 features and 61 outliers.

\section{E. Missing value estimation}
\paragraph{E1. ABDA can efficiently marginalize over missing values during inference}
In order to be able to deal with missing values at inference time, we first need to provide ABDA with a LV structure that has been induced from the non-missing data entries only.
Secondly, we need to marginalize over these entries while performing Gibbs sampling.
This can be done efficiently by exploiting SPNs' ability to decompose marginal queries into simpler marginalizations at the leaf distributions~\cite{Darwiche2003,Peharz2017}.
When updating the posterior parameters for the leaf likelihood models, we keep track of counts belonging only to non-missing entries.

To this end, we adapted the likelihood-agnostic structure learning introduced by MSPNs~\cite{Molina2017b}.
In a nutshell, in presence of missing entries in the training samples, these are not contributing to the evaluation of the RDC both for the partitioning along the rows of the data matrix (clustering), and for the partitioning along RVs or features (group independence seeking).
Specifically, while transforming the data matrix through the copula transformation (for details, refer to \cite{Molina2017b}) we do not let missing entries contribute to the estimation of the empirical cumulative distribution.
We reserve the same treatment for MSPNs as well.

\paragraph{E2. ABDA can estimate missing values efficiently}
We assume each sample $\x_{n}$ to be in the form $\x_{n}=(\x_{n}^{o},\x_{n}^{m})$, where  $\x_{n}^{o}$  comprises observed values and $\x^{m}_{n}$ missing ones.

To evaluate the effectiveness of a generative model $\mathcal{M}$ in estimating missing values, we employ the probability of seeing the true missing entries $\bar{\x}_{n}^{m}$, i.e., $p_{\mathcal{M}}(\bar{\x}_{n}^{m})$, as a standard approach indicating the ability of $\mathcal{M}$ of selecting the true value for a RV~\cite{Valera2017b}.
This operation requires marginalizing over all observed values $\x_{n}^{o}$, which can be done efficiently with SPN-based models~\cite{Darwiche2003,Peharz2017}.

To impute missing values, on the other hand, we resolve to compute the the \textit{most probable explanation} (MPE)~\cite{Darwiche2009} for the corresponding RVs, that is: $$\tilde{\x}_{n}^{m}=\argmax_{\x_{n}^{m}}p_{\mathcal{M}}(\x_{n}^{m}|\x_{n}^{o})=\argmax_{\x_{n}^{m}}p_{\mathcal{M}}(\x_{n}^{m},\x_{n}^{o})$$
Note that this task is performed by factorization-based models like ISLV efficiently, since each RVs appearing in $\x^{m}_{n}$ is imputed \textit{independently} from others.
Conversely, ABDA and MSPNs can leverage all other RVs while performing imputation. 
The price to pay for this is that, for general SPNs, exact MPE imputation is NP-Hard~\cite{Peharz2017,Choi2017}.
Nevertheless, reasonably accurate approximations to this tasks exist and allow to evaluate an SPN structure in linear time~\cite{Darwiche2003,Chan2006}.
In brief, they employ a bottom-up propagation step for observing $\x_{n}^{o}$, at first, while performing a Viterbi-like decoding top-down traversal of the SPN structure to retrieve the maximally likely states for the RVs in $\x^{m}_{n}$.
Please refer to~\cite{Peharz2017} for more details.

\paragraph{E3. Missing value estimation experiments}

For our missing value estimation experiments in the transductive scenarios we monitor both the mean log-likelihood for the missing data entries in a run and the normalized root mean squared error (NRMSE) w.r.t. the imputation provided by each model (please refer to the previous Section for how to compute them in ABDA).

\begin{figure}[!t]
  \centering
      \includegraphics[width=0.24\columnwidth]{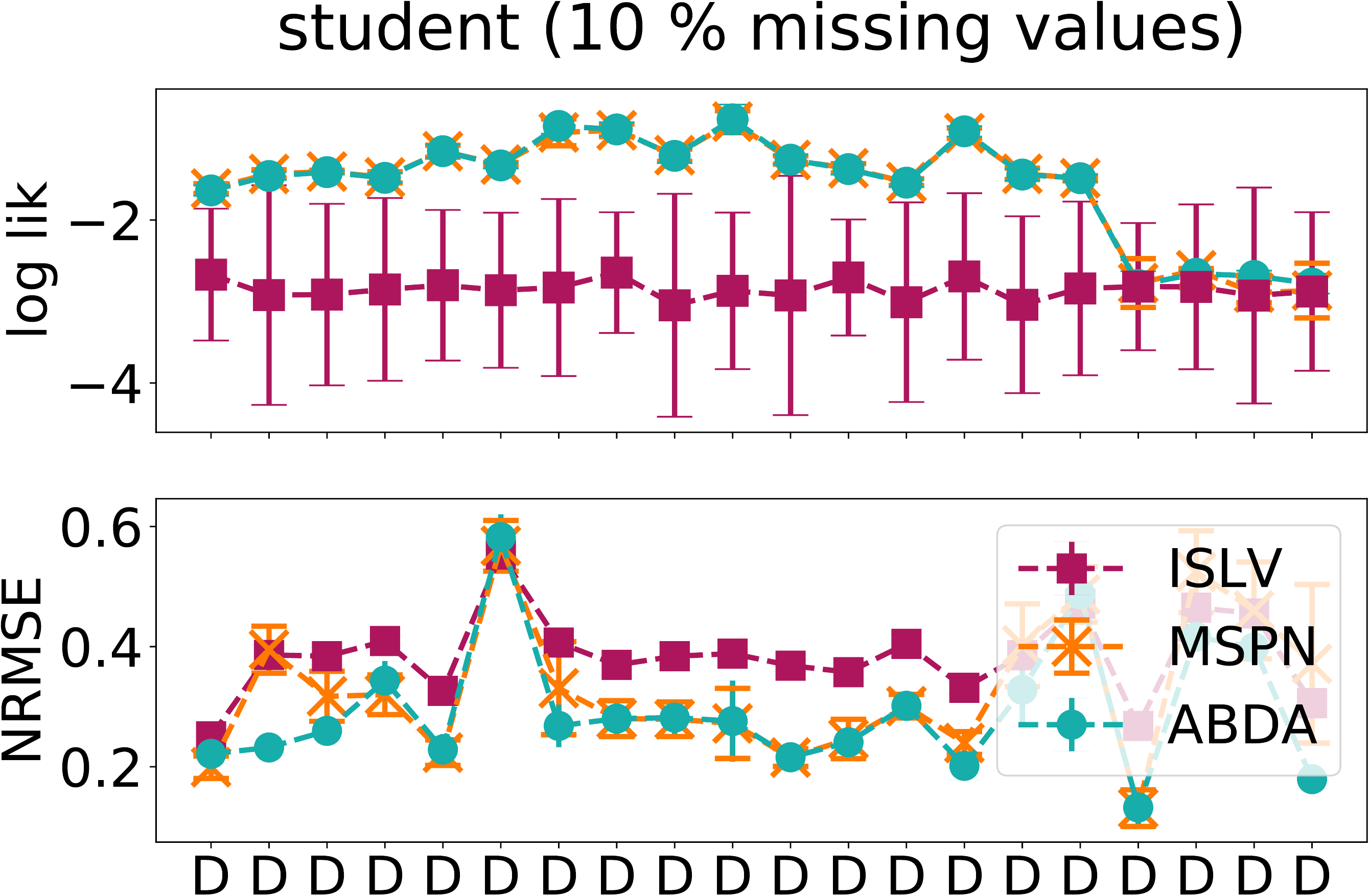}
    \includegraphics[width=0.24\columnwidth]{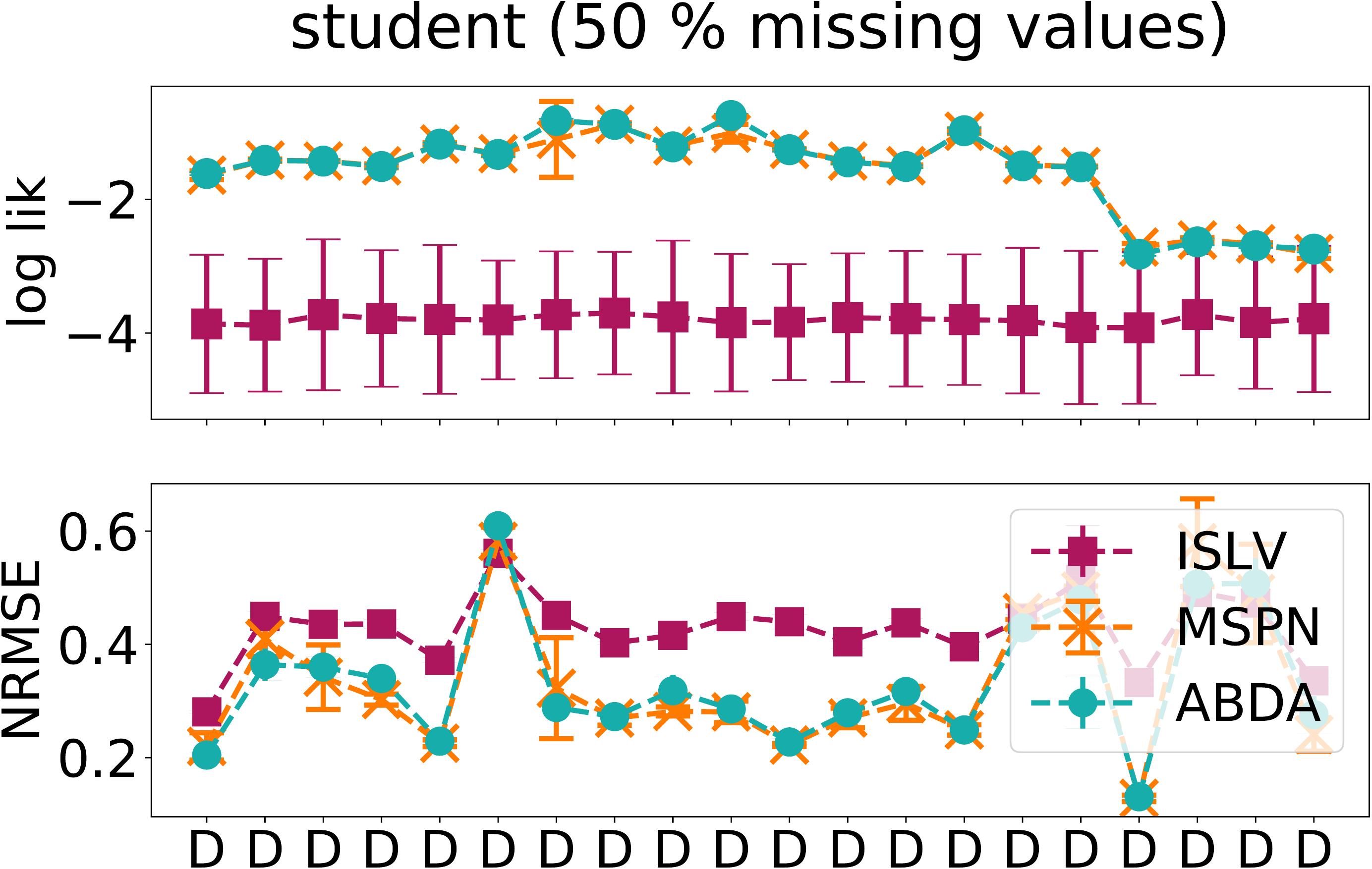}\hspace{5pt}
    \includegraphics[width=0.24\columnwidth]{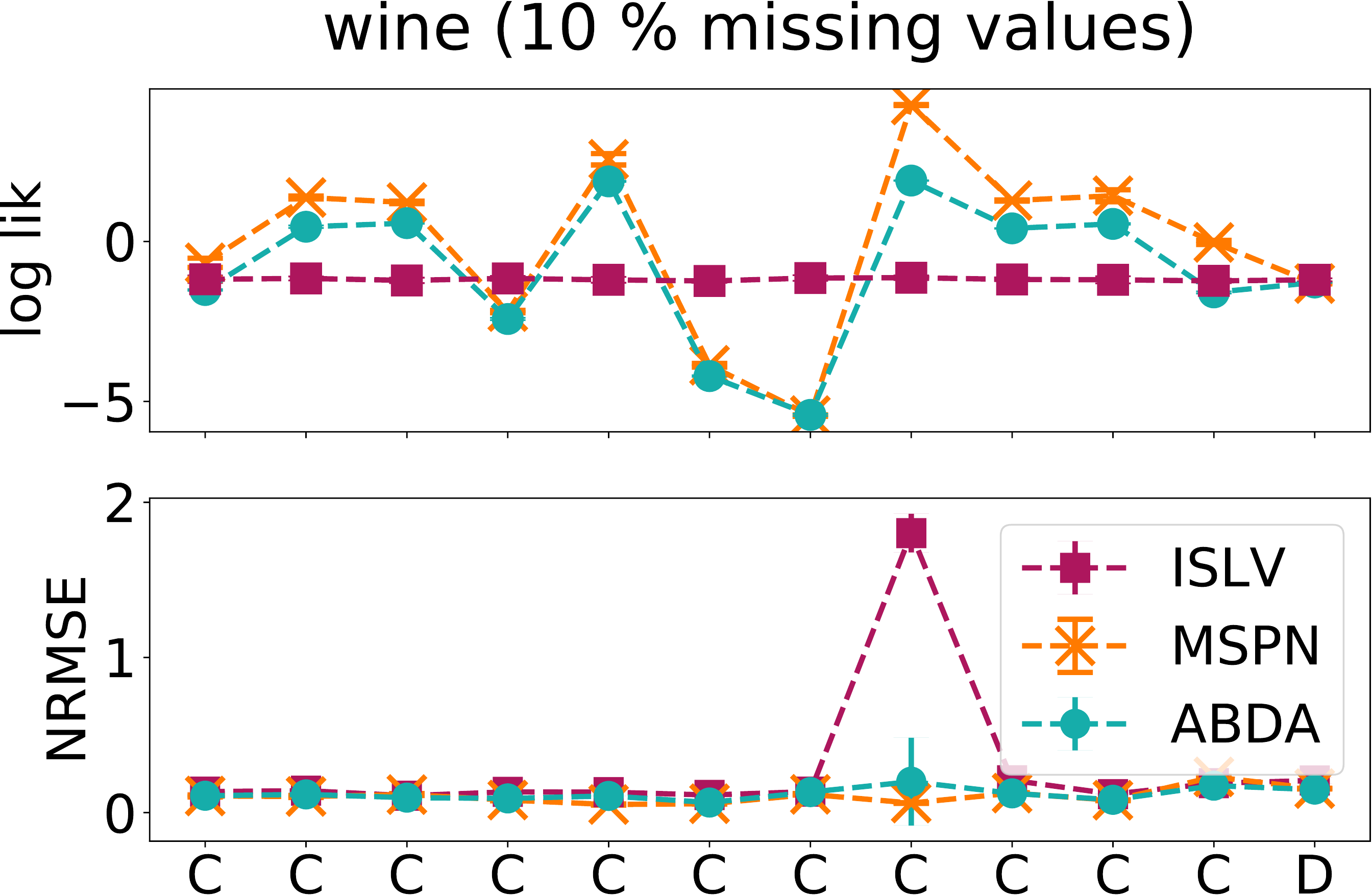}
    \includegraphics[width=0.24\columnwidth]{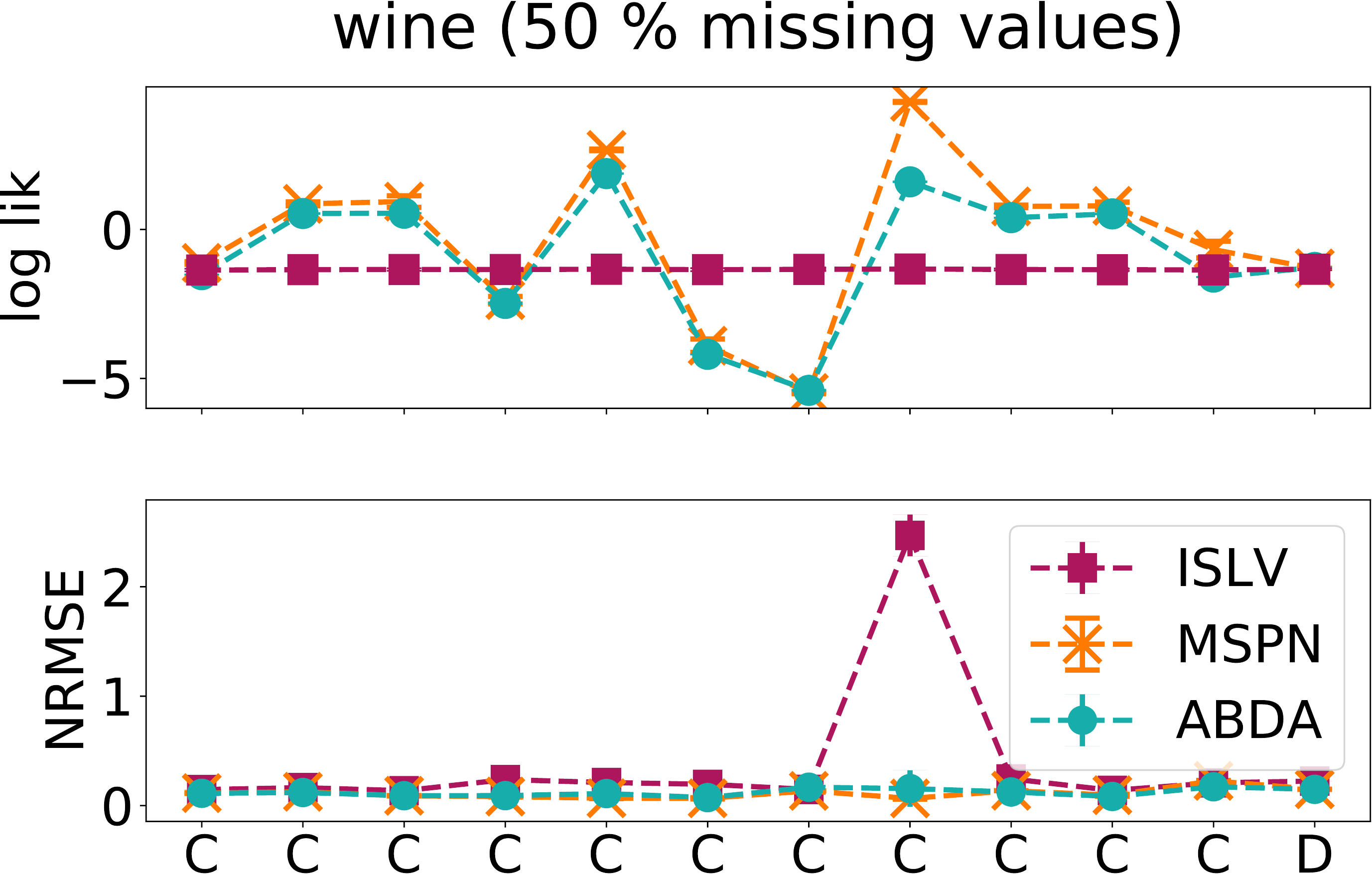}
        \\[5pt]
    \includegraphics[width=0.24\columnwidth]{R-RMSE-err-d-abalonePP-0-1-crop.pdf}
    \includegraphics[width=0.24\columnwidth]{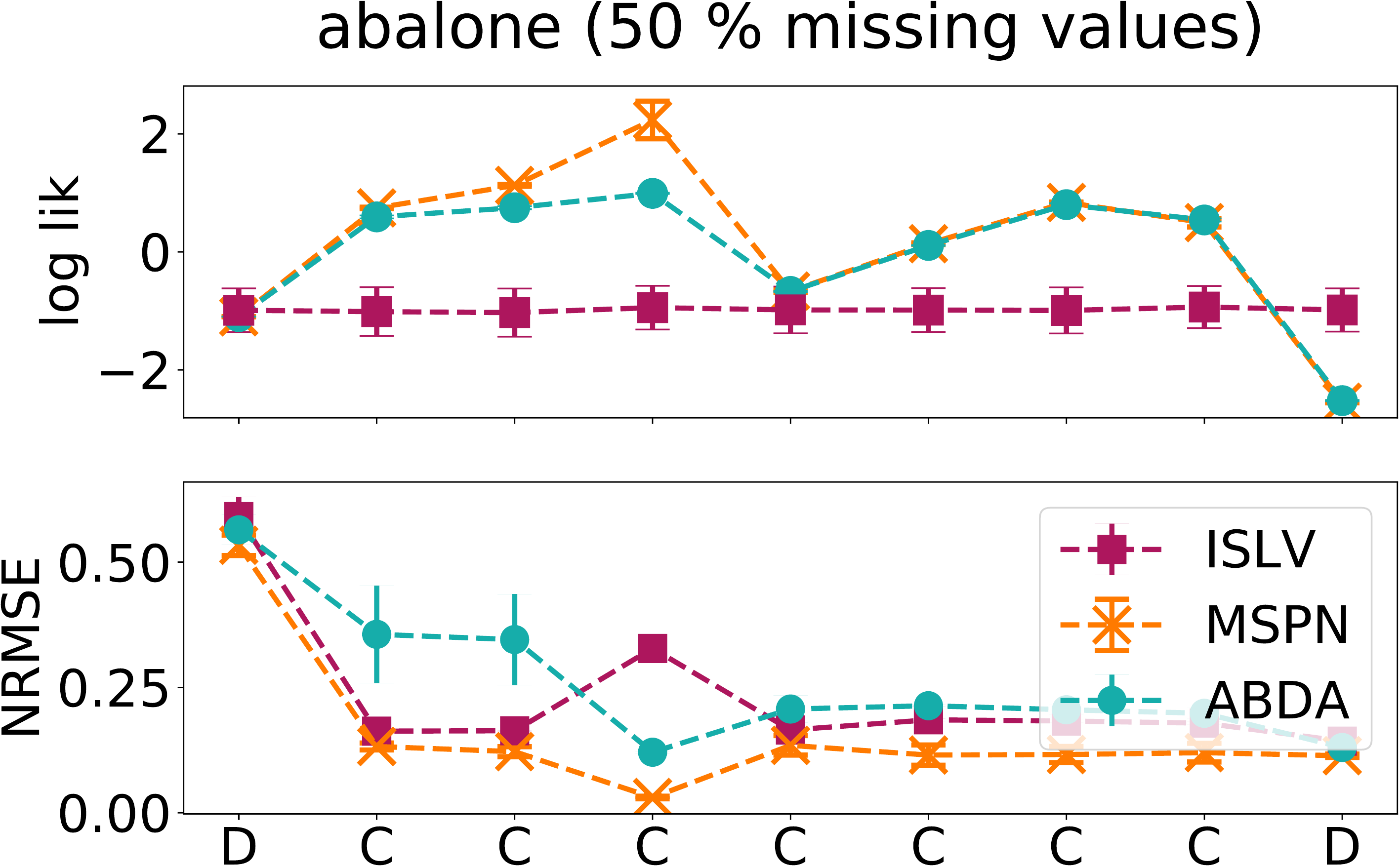}\hspace{5pt}
    \includegraphics[width=0.24\columnwidth]{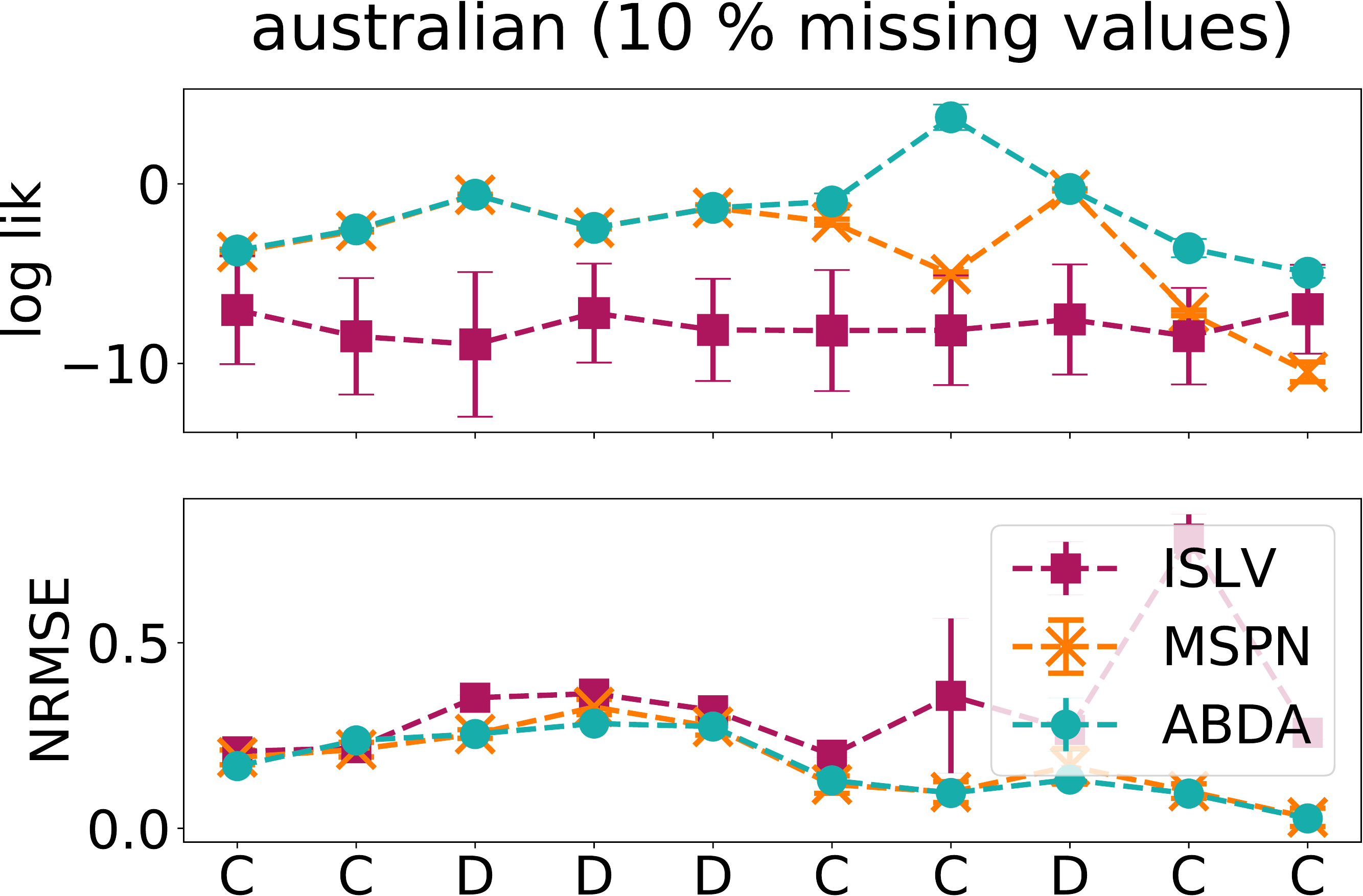}
    \includegraphics[width=0.24\columnwidth]{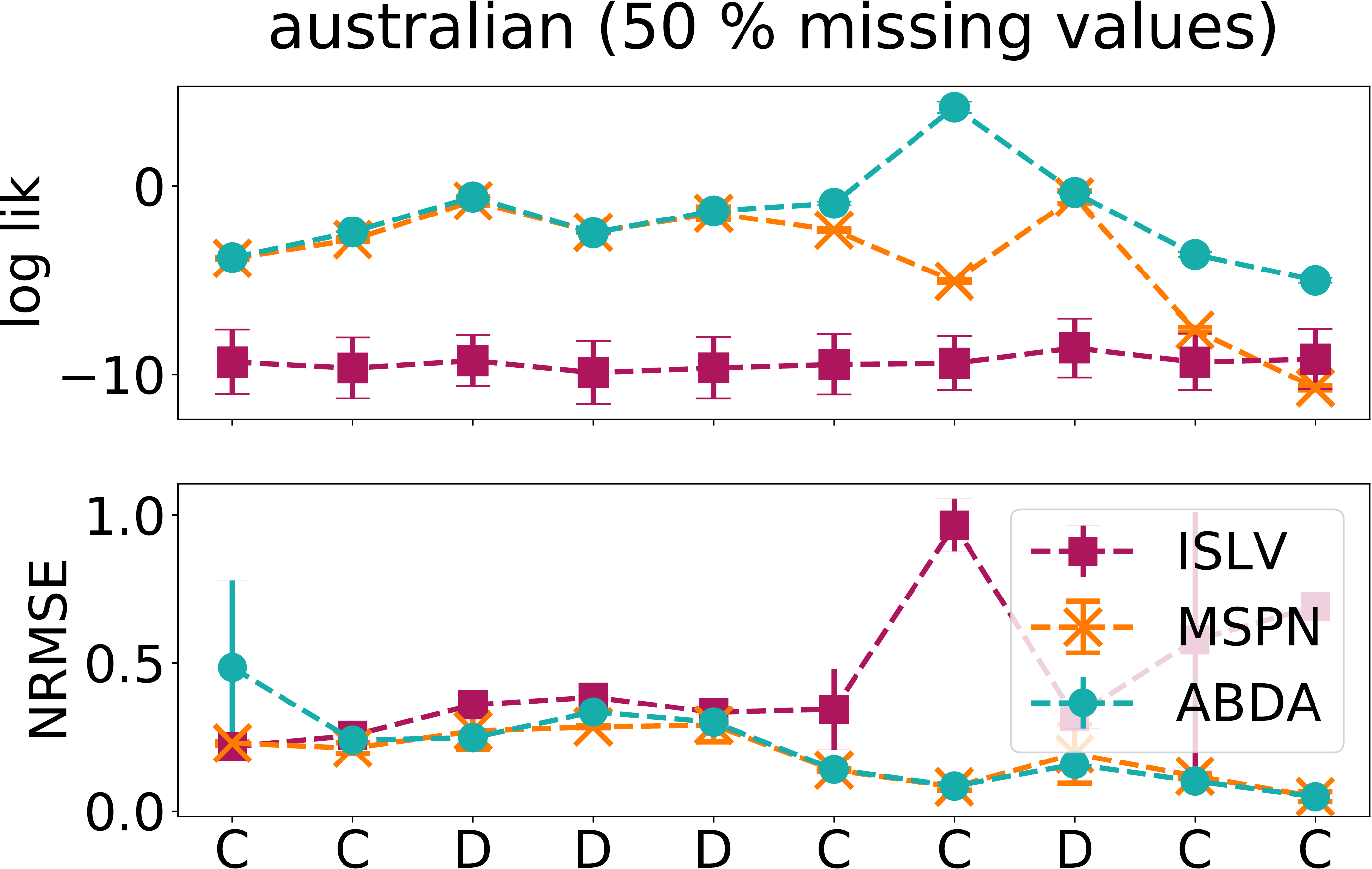}
        \\[15pt]
    \includegraphics[width=0.24\columnwidth]{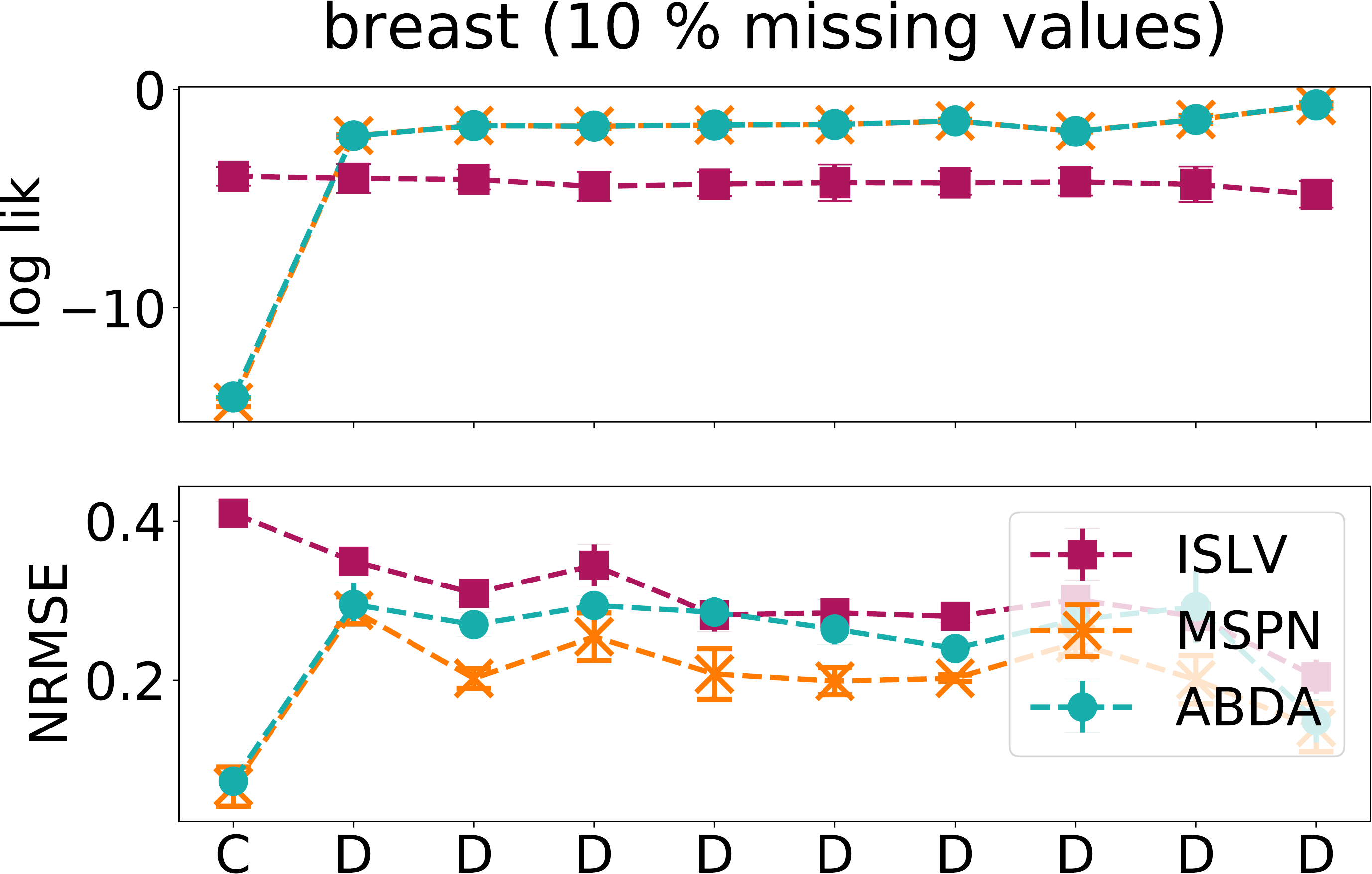}
    \includegraphics[width=0.24\columnwidth]{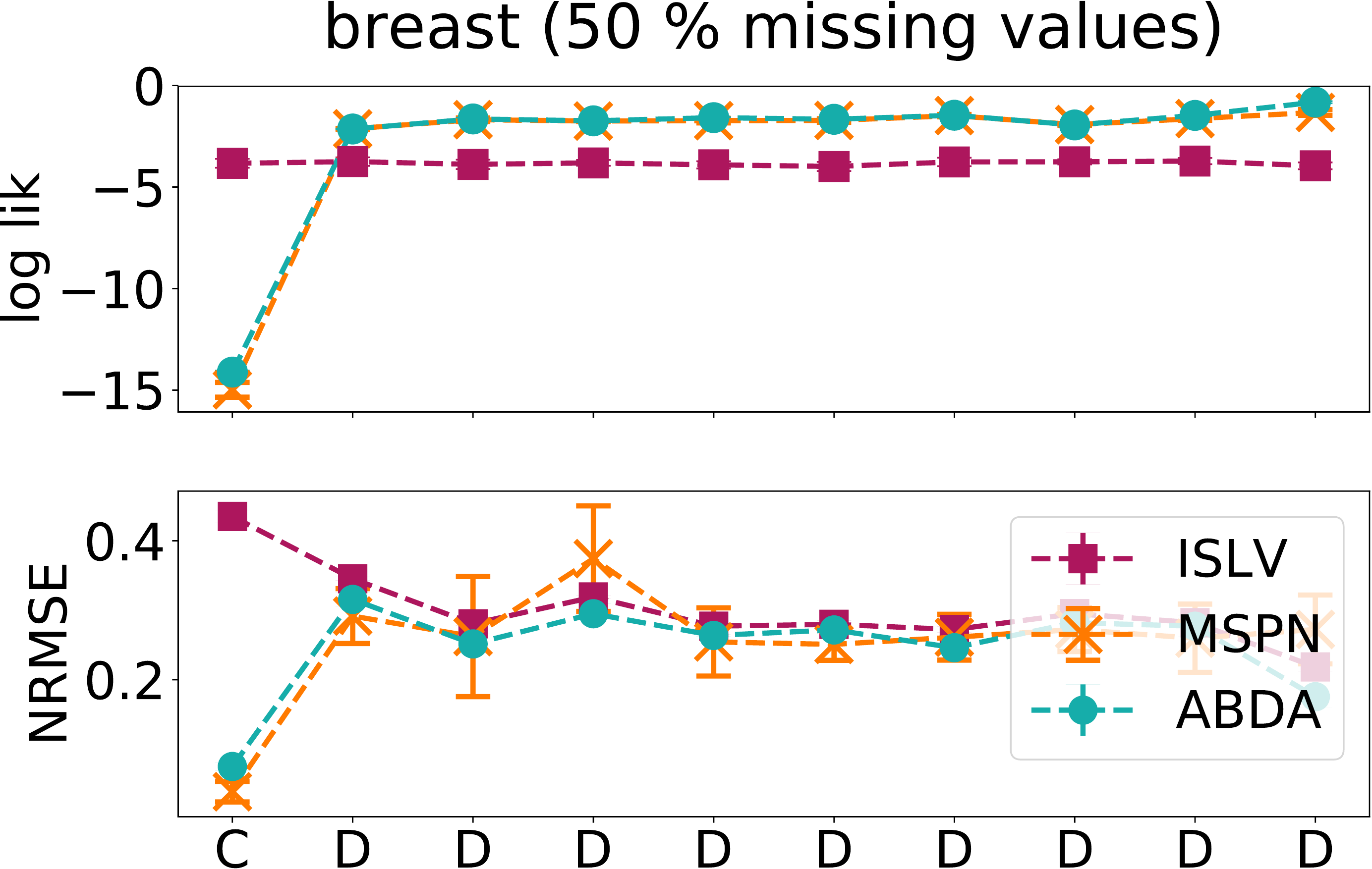}\hspace{5pt}
    \includegraphics[width=0.24\columnwidth]{R-RMSE-err-d-chessPP-0-1-crop.pdf}
    \includegraphics[width=0.24\columnwidth]{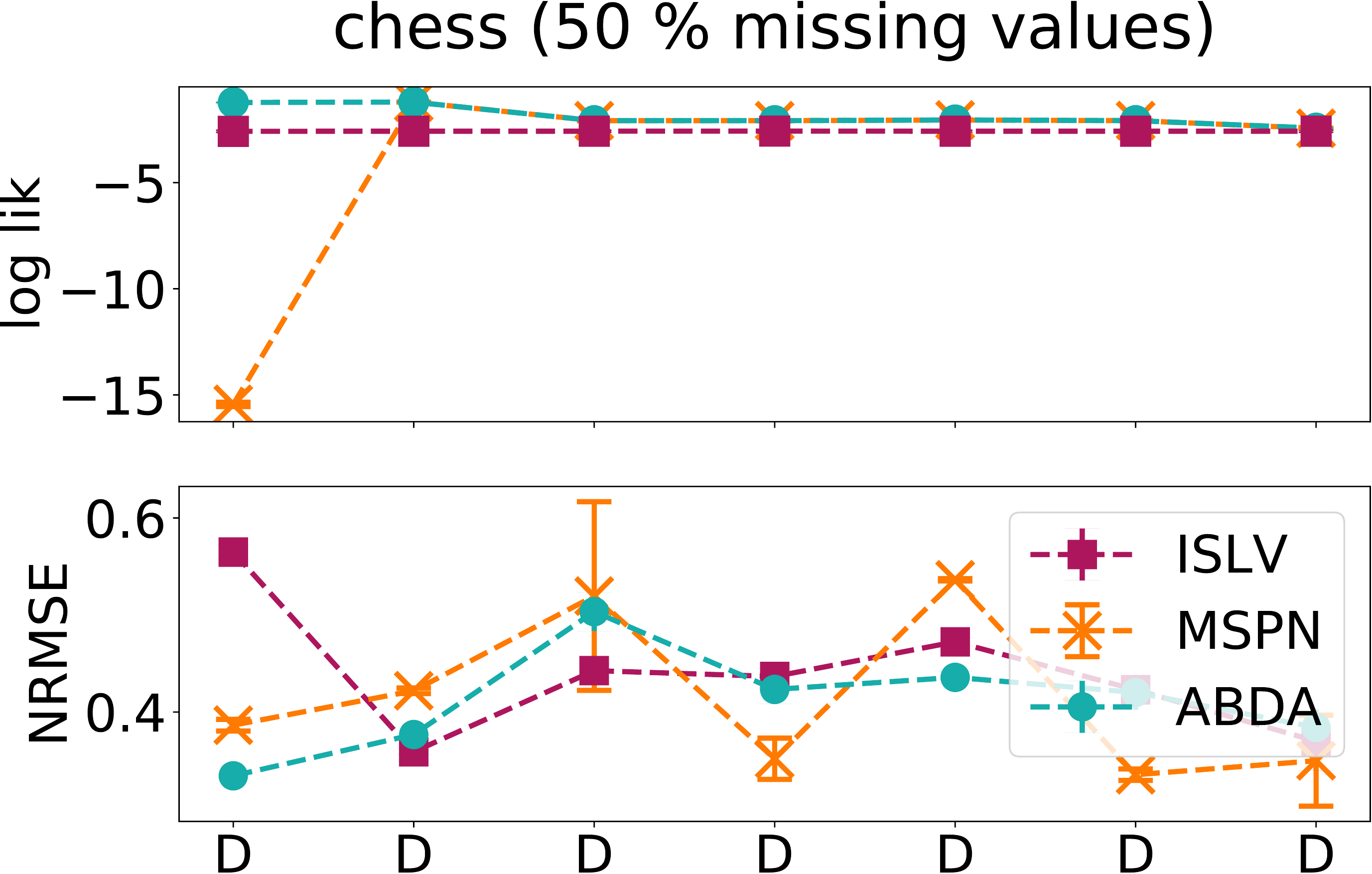}
    \\[15pt]
    \includegraphics[width=0.24\columnwidth]{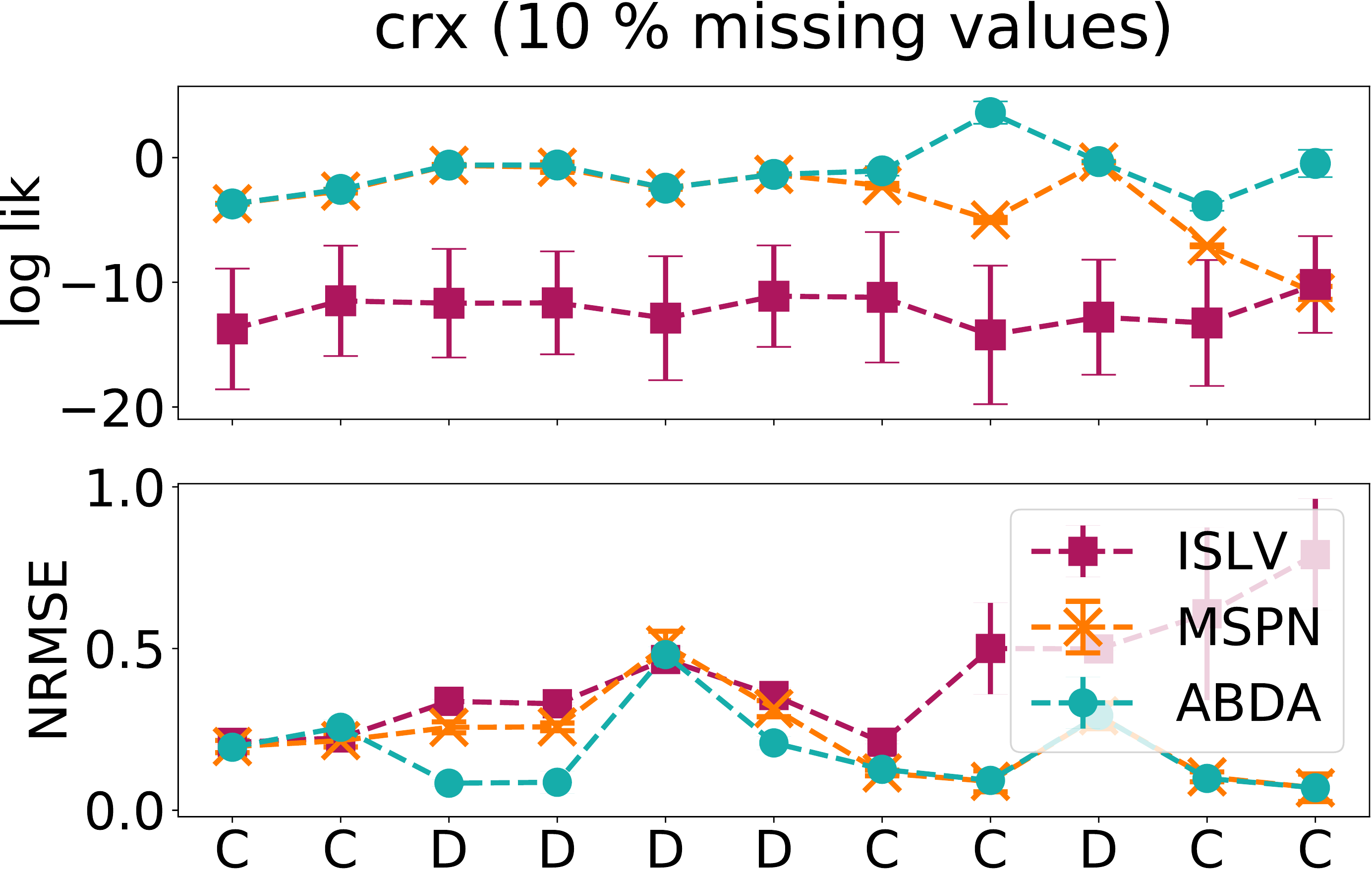}
    \includegraphics[width=0.24\columnwidth]{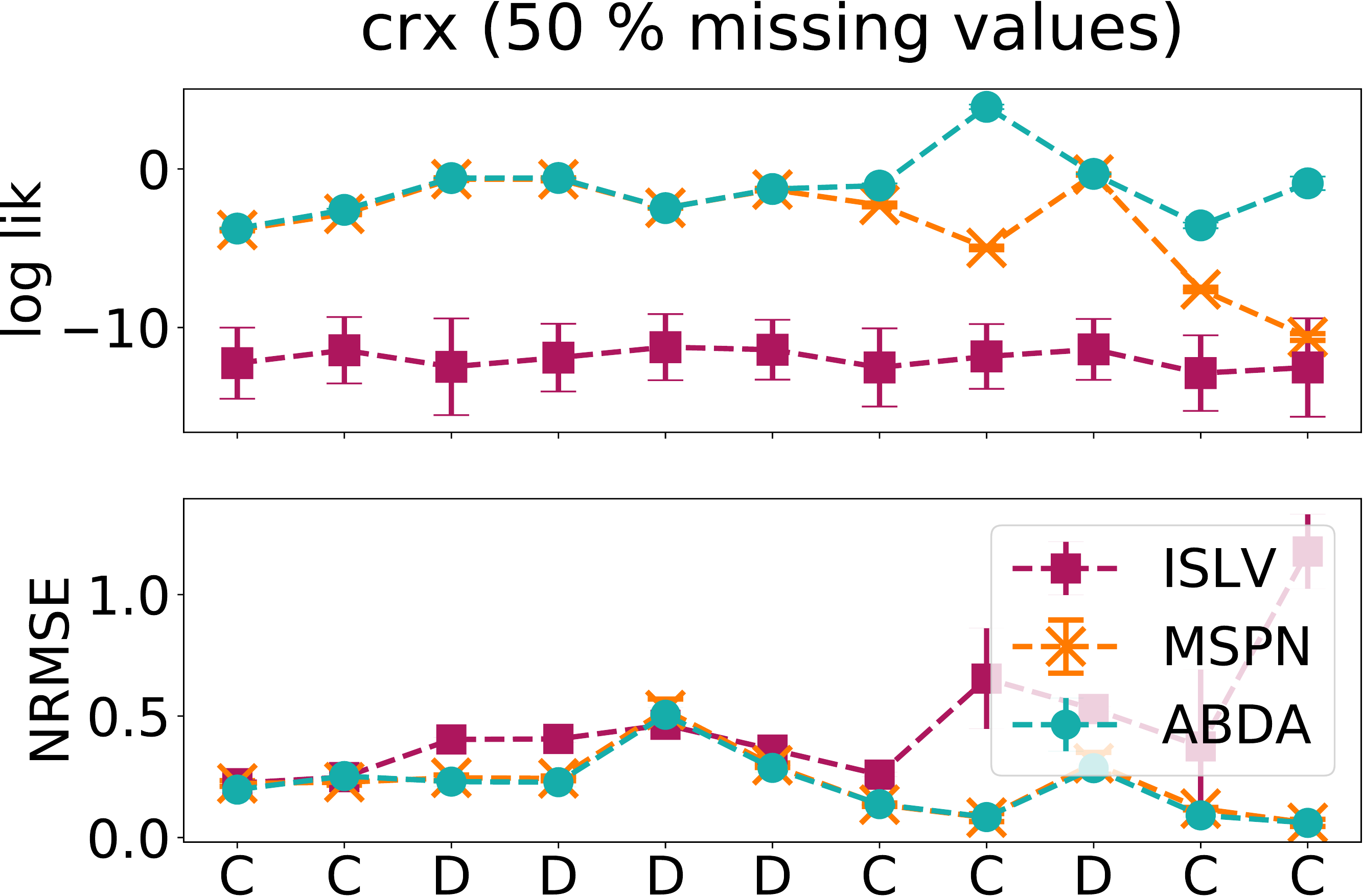}\hspace{5pt}
    \includegraphics[width=0.24\columnwidth]{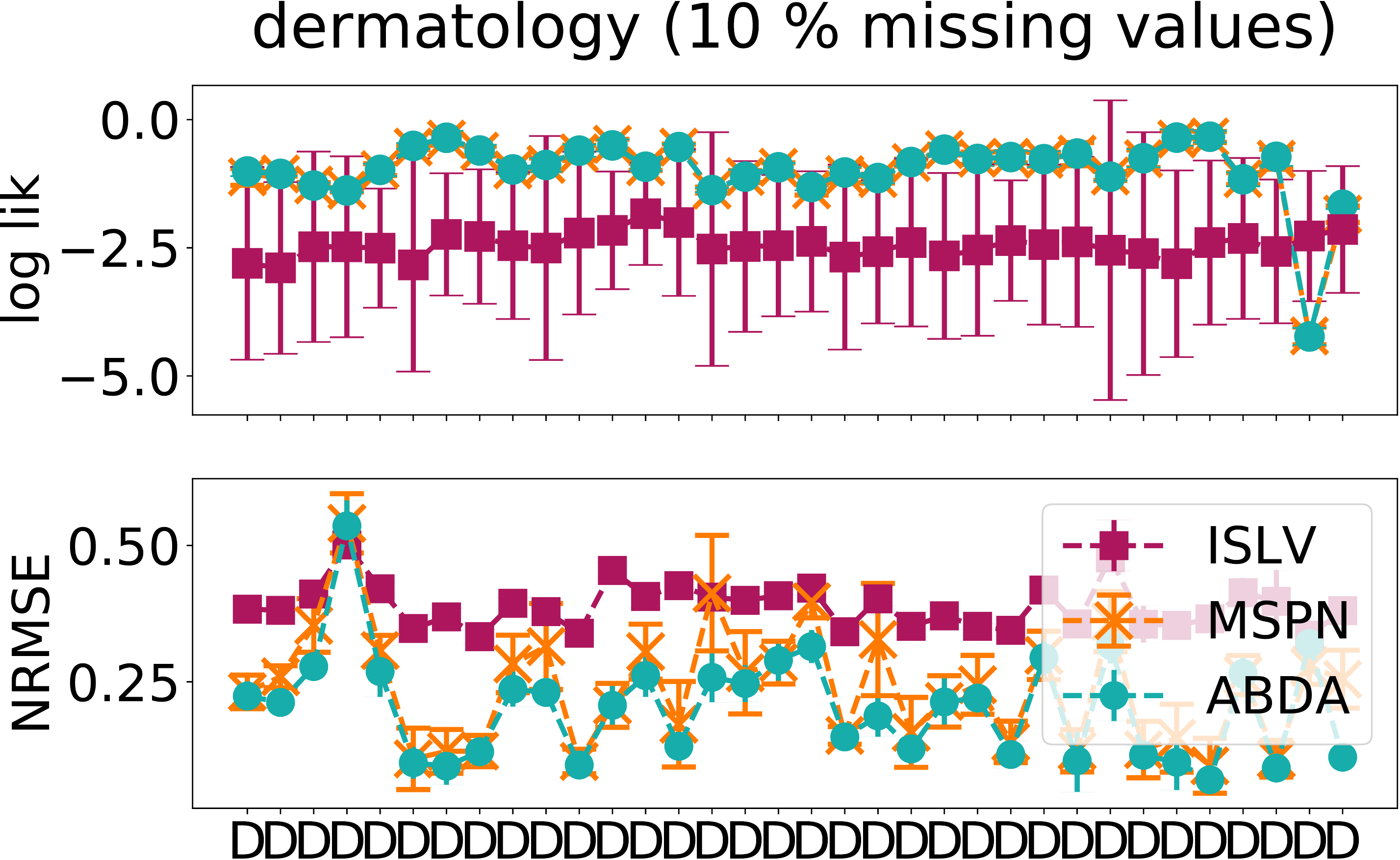}
    \includegraphics[width=0.24\columnwidth]{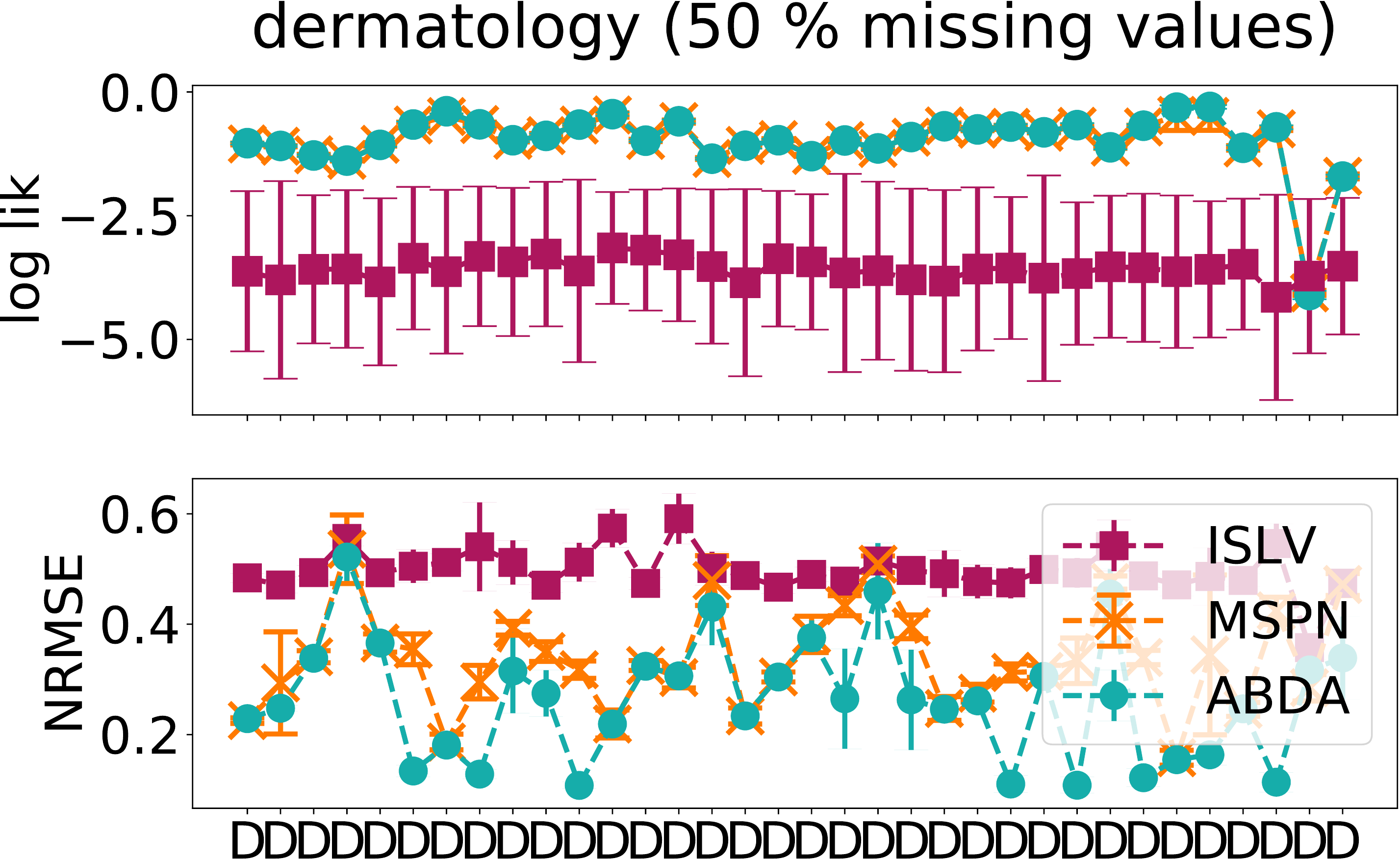}
    \\[15pt]
    \includegraphics[width=0.24\columnwidth]{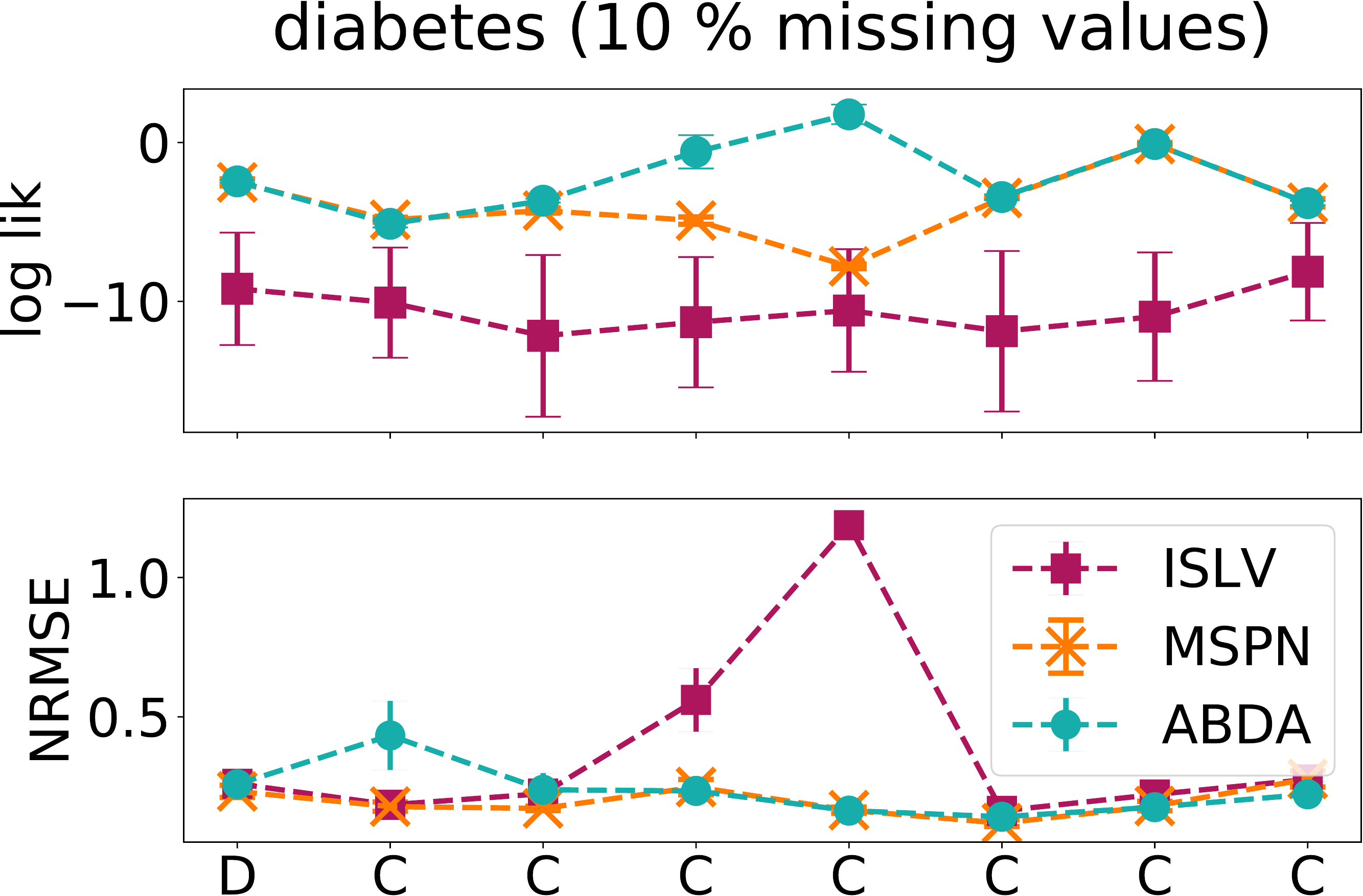}
    \includegraphics[width=0.24\columnwidth]{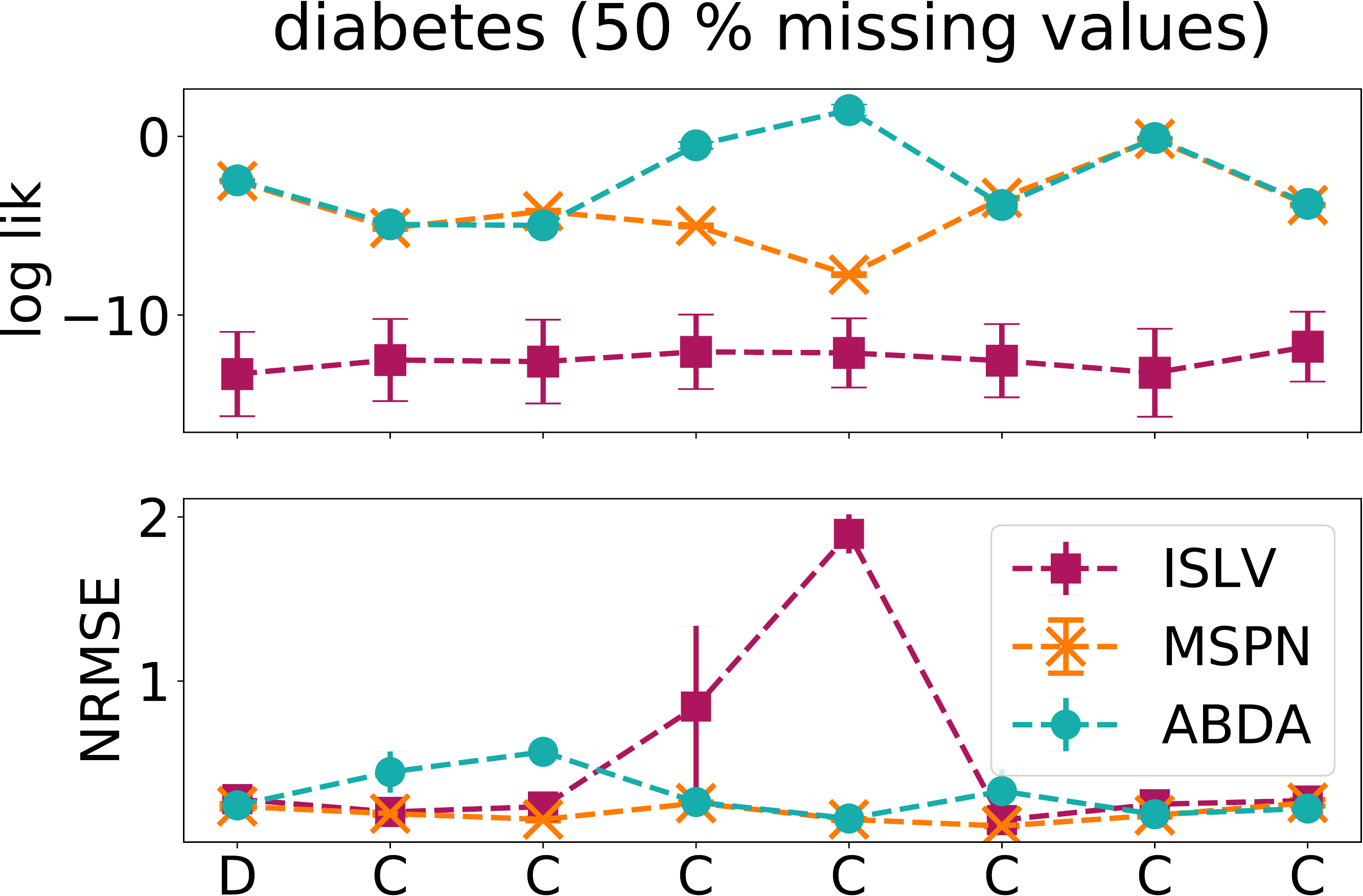}\hspace{5pt}
    \includegraphics[width=0.24\columnwidth]{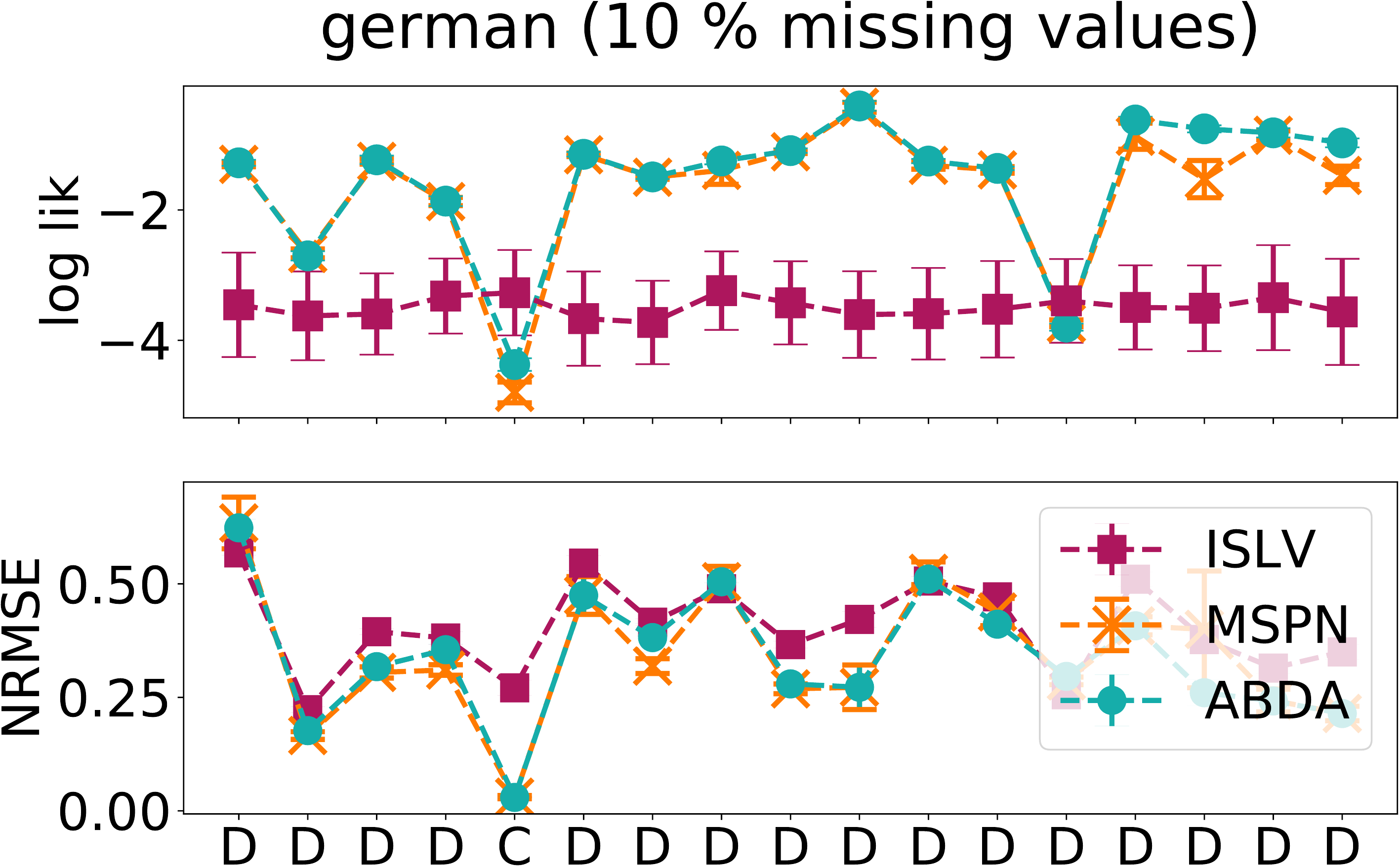}
    \includegraphics[width=0.24\columnwidth]{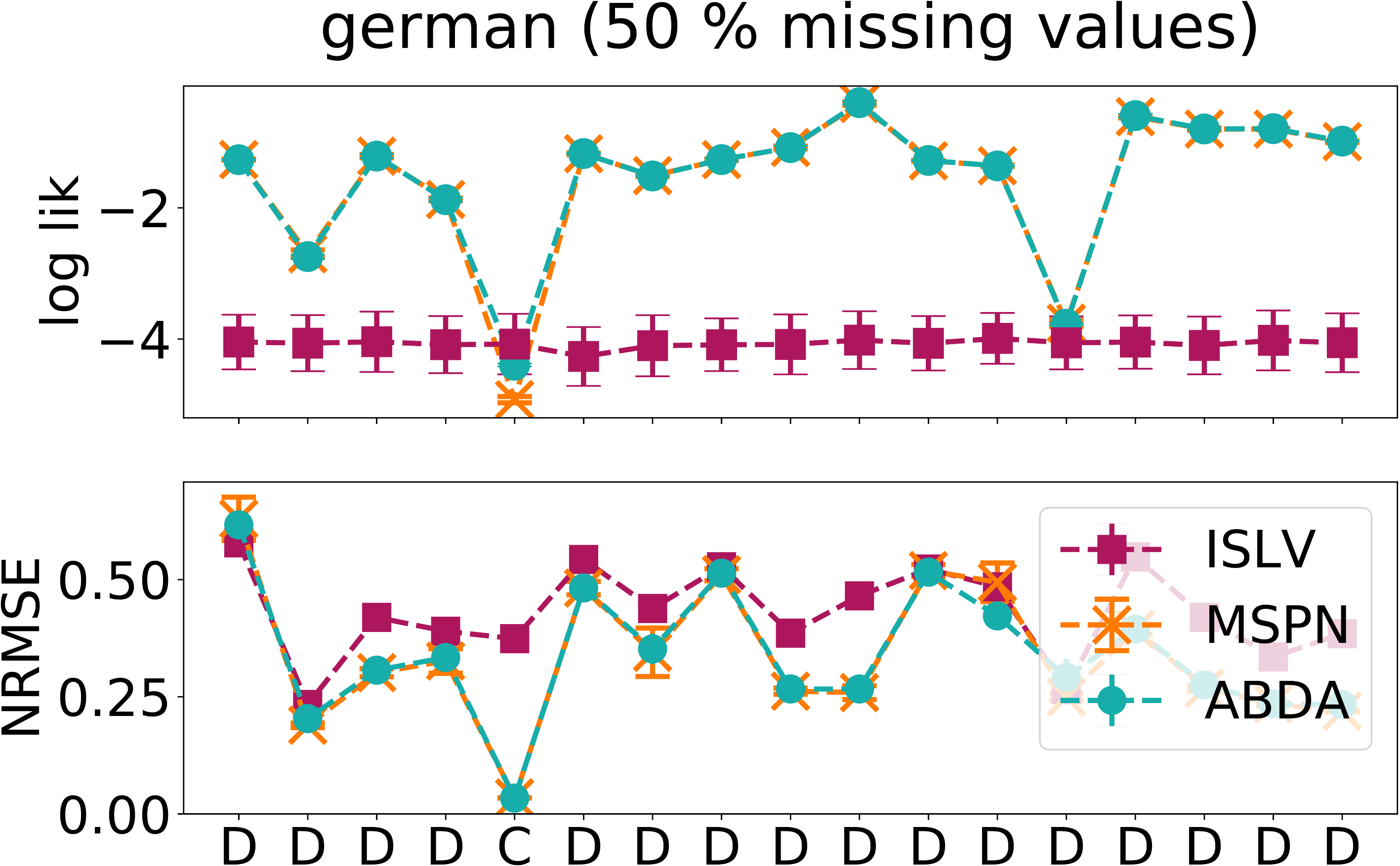}
    \caption{Performance (test log-likelihood, normalized RMSE feature-wise) on real datasets.}
    \label{fig:perf1}
\end{figure}

Figure~\ref{fig:perf1} reports the feature-wise plots for the above mentioned metrics.
As a general trend, one can see how ISLV performs less well than ABDA and MSPNs and how the difference between these two is often significant only on a small subset of the modeled features.
Generally, it seems that ABDA has the advantage of modeling continuous RVs, confirming that the performing discretization as MSPNs do is potentially a problem.
However, note also that in such cases, MSPNs log-likelihood values might be inflated by their adoption of piecewise polynomial approximations as likelihood models.
Indeed, since they are fitted on the small range of values available during learning, their density is renormalized to integrate to 1. On the other hand, since the likelihood distributions used in ABDA generally have infinite tails, their mass is not concentrated only on that observed range.
    

\newpage
\section{F. Anomaly robustness and detection}

\paragraph{F1. ABDA is robust to corrupted data and outliers}
 During inference, ABDA will tend to assign anomalous
 samples  into low-weighted mixture components, i.e., sub-networks of
 the underlying SPN or leaf likelihood models.
 If more that one anomalous sample is relegated to the same partition,
 they will form a \emph{micro-cluster}~\cite{Chandola2009}.

 At test time, ABDA would assign low probability to outliers or novel
 samples.
 The (log-)likelihood of the whole joint distribution can be employed
 as a score for inliers~\cite{Scholkopf2001,Chandola2009} and by
 proper thresholding, one can have a decision rule to detect
 anomalies.
 
 This process can be repeated at each node in the underlying SPN
 structure, thus enabling ABDA to perform \emph{hierarchical anomaly
   detection}
 to decide whether a sample---or just a subset of features in a
 sample (i.e., contextual outliers~\cite{Chandola2009})---is anomalous
 with respect to the distribution induced at that node.
 
As a qualitative example, Figure~\ref{fig:synth} illustrates how ABDA partitioned the Shuttle data, correctly relegating most of the outliers into micro-clusters.
Those not associated to a single micro-cluster are still belonging to the tail of the distribution modeling the time feature (x-axis).

\begin{figure}[!t]
        \includegraphics[width=0.3\columnwidth]{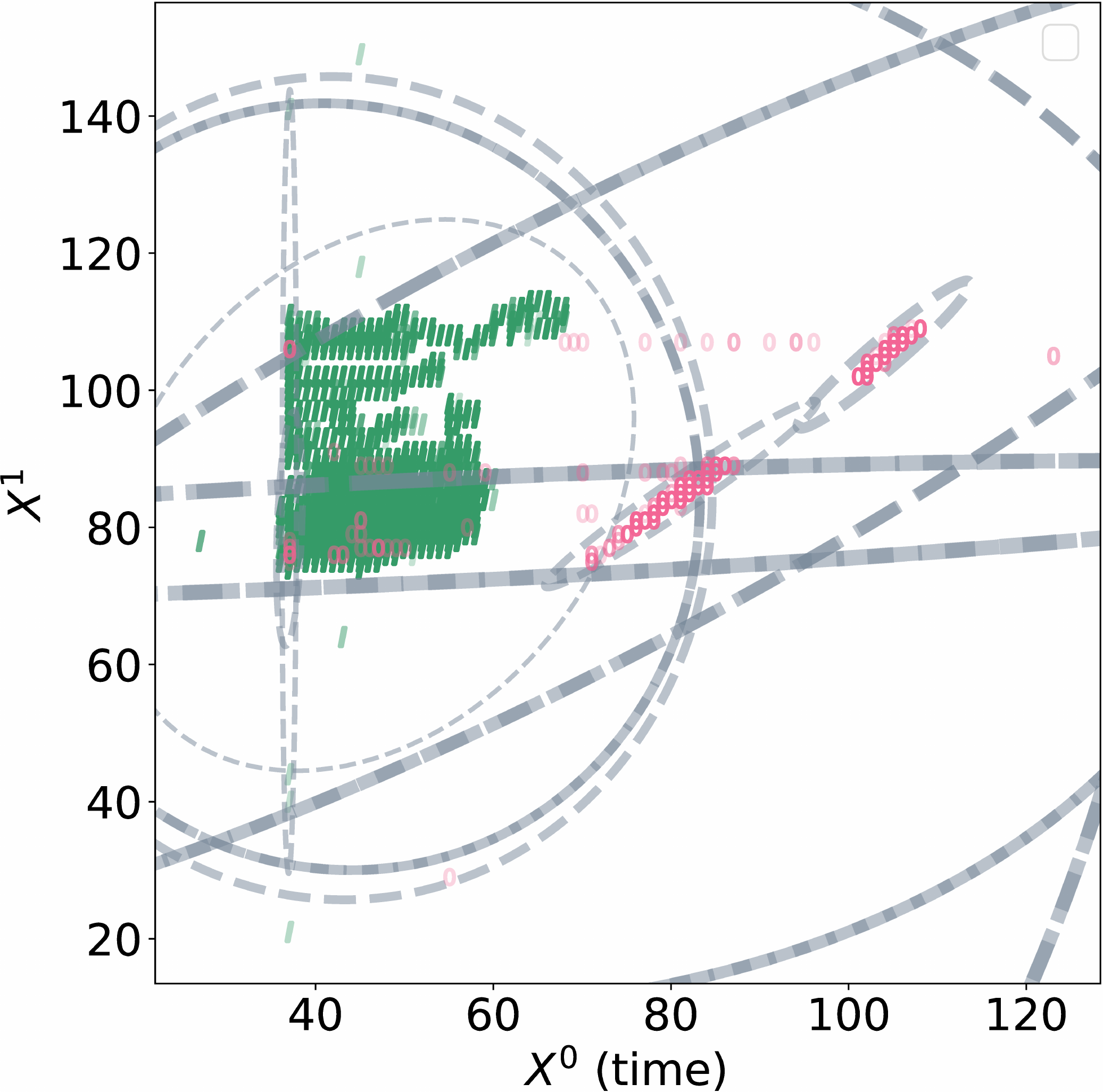}
    \caption{ABDA separates outliers (pink `O') from inliers (green `I') via hierarchical partitioning on \textsf{Shuttle} data. Each ellipse represents a partition induced by the underlying SPN w.r.t. the first two features ($X^{0}$ and $X^{1}$) of the shuttle data.
    The thickness of the ellipses is inversely proportional to the number of samples in the corresponding partition, therefore it also indicates the depth of the partition---finer lines for finer grain clusters.}
    \label{fig:synth}

      \vspace{-6pt}
\end{figure}
\paragraph{F2. Unsupervised outlier detection}
We quantitatively evaluate how ABDA is able to perform \emph{unsupervised pointwise anomaly
  detection}~\cite{Chandola2009}, that is detect outliers available at training
time, without having access to their label.

We follow the experimental setting of~\cite{Goldstein2016}, in which a
set of UCI binary classification datasets have been processed to
include all samples from one class (\emph{inliers}) and a small percentage of
dissimilar samples drawn from the other class (\emph{outliers}).
Please refer to~\cite{Goldstein2016} for detailed dataset statistics.

%
Once trained on these datasets, all models involved are required to output a
score for each training sample, the higher the most confident the
model is about the sample being an outlier.
For ABDA, MSPN and ISLV we employ the sample negative log-likelihood
as the outlier score.

%
As competitors we include one-class support vector machines
(oSVM)~\cite{Scholkopf2001},
the local outlier factor (LOF)~\cite{Breunig2000} and the
histogram-based outlier score (HBOS)~\cite{Goldstein2012}.
%
Accuracy performances for all models are measured in terms of the area under the
receiving operating curve (AUC ROC) computed w.r.t. to the outlier
scores.

%
We follow~\cite{Goldstein2016}, and instead of picking one single model
out of a grid search to optimize its hyperparameters, we average over
all AUC scores from the cross validated models.
We emply an RBF kernel ($\gamma=0.1$) for oSVMs, varying
$\nu\in\{0.2, 0.4, 0.6, 0.8\}$.
For LOF, we run it by evaluating a number of nearest neighbors
$k\in\{20,30,40,50\}$.
For HBOS, instead, we explored these configurations with the Laplacian
smoothing factor $\alpha\in\{0.1, 0.2, 0.5\}$ and number of bins in
$\{10,20,30\}$.
For ABDA and MSPNs we set the minimum number of instances $m$ to $5\%$
of the number of instances and explored the RDC coefficient
$\rho\in\{0.1, 0.3, 0.5\}$.
For ISLV, we run it several times for
$k\in\{\floor*{\frac{D}{3}},\floor*{\frac{D}{2}},
\floor*{\frac{2D}{3}}\}$.

\begin{table}[!h]
  \caption{Unsupervised anomaly detection with ABDA.
  Mean and standard deviation (small) AUC ROC scores for all models and datasets.}
\label{tab:od}
\setlength{\tabcolsep}{1pt}
\scriptsize
    \begin{tabular}{r r r r r r r}
    \toprule
    &\textbf{oSVM}&\textbf{LOF}&\textbf{HBOS}&\textbf{ABDA}&\textbf{MSPN}&\textbf{ISLV}\\
     \midrule
\textsf{Aloi}&51.71&\textbf{74.19}&52.86&47.20&51.86&-\\[-2pt]
      &$\scriptstyle\scriptstyle\pm 0.02$&
       $\scriptstyle\scriptstyle\pm0.70$&$\scriptstyle\scriptstyle\pm
                                          0.53$&$\scriptstyle\scriptstyle\pm 0.02$&$\scriptstyle\scriptstyle\pm
                                                  1.04$&\\
      \textsf{Thyroid}&46.18&62.38&62.77&\textbf{84.88}&77.74&-\\[-2pt]
      &$\scriptstyle\scriptstyle\pm0.39$
                  &$\scriptstyle\scriptstyle\pm1.04$
                               &$\scriptstyle\scriptstyle\pm3.69$
                                             &$\scriptstyle\scriptstyle\pm0.96$
                                                           &$\scriptstyle\scriptstyle\pm0.33$&\\
      \textsf{Breast}&45.77&98.06&94.47&\textbf{98.36}&92.29&82.82\\[-2pt]
      &$\scriptstyle\scriptstyle\pm11.12$
                  &$\scriptstyle\scriptstyle\pm0.70$
                               &$\scriptstyle\scriptstyle\pm0.79$
                                             &$\scriptstyle\scriptstyle\pm0.07$
                                                           &$\scriptstyle\scriptstyle\pm0.99$&$\scriptstyle\scriptstyle\pm1.31$\\
      \textsf{Letter}&63.38&\textbf{86.55}&60.47&70.36&68.61&61.98\\[-2pt]
      &$\scriptstyle\scriptstyle\pm17.60$&$\scriptstyle\scriptstyle\pm2.23$&$\scriptstyle\scriptstyle\pm1.80$&$\scriptstyle\scriptstyle\pm0.01$&$\scriptstyle\scriptstyle\pm0.23$&$\scriptstyle\scriptstyle\pm0.44$\\
      \textsf{Kdd00}&53.40&46.39&87.59&\textbf{99.79}&70.17&-\\[-2pt]
      &$\scriptstyle\scriptstyle\pm3.63$&$\scriptstyle\scriptstyle\pm1.95$&$\scriptstyle\scriptstyle\pm4.70$&$\scriptstyle\scriptstyle\pm0.10$&$\scriptstyle\scriptstyle\pm15.97$&\\
      \textsf{Pen-global}&46.86&87.25&71.93&\textbf{89.87}&75.93&75.54\\[-2pt]
      &$\scriptstyle\scriptstyle\pm1.02$&$\scriptstyle\scriptstyle\pm1.94$&$\scriptstyle\scriptstyle\pm1.68$&$\scriptstyle\scriptstyle\pm2.87$&$\scriptstyle\scriptstyle\pm0.14$&$\scriptstyle\scriptstyle\pm5.40$\\
      \textsf{Pen-local}&44.11&\textbf{98.72}&64.30&90.86&79.25&65.52\\[-2pt]
      &$\scriptstyle\scriptstyle\pm6.07$&$\scriptstyle\scriptstyle\pm0.20$&$\scriptstyle\scriptstyle\pm2.70$&$\scriptstyle\scriptstyle\pm0.79$&$\scriptstyle\scriptstyle\pm5.41$&$\scriptstyle\scriptstyle\pm13.08$\\
      \textsf{Satellite}&52.14&83.51&90.92&94.55&\textbf{95.22}&-\\[-2pt]
      &$\scriptstyle\scriptstyle\pm3.08$&$\scriptstyle\scriptstyle\pm11.98$&$\scriptstyle\scriptstyle\pm0.16$&$\scriptstyle\scriptstyle\pm0.68$&$\scriptstyle\scriptstyle\pm0.45$&\\
      \textsf{Shuttle}&89.37&66.29&\textbf{98.47}&78.61&94.96&-\\[-2pt]
      &$\scriptstyle\scriptstyle\pm5.13$&$\scriptstyle\scriptstyle\pm1.69$&$\scriptstyle\scriptstyle\pm0.24$&$\scriptstyle\scriptstyle\pm0.02$&$\scriptstyle\scriptstyle\pm2.13$&\\
      \textsf{Speech}&45.61&\textbf{49.37}&47.47&46.96&48.24&-\\[-2pt]
      &$\scriptstyle\scriptstyle\pm3.64$&$\scriptstyle\scriptstyle\pm0.87$&$\scriptstyle\scriptstyle\pm0.10$&$\scriptstyle\scriptstyle\pm0.01$&$\scriptstyle\scriptstyle\pm0.67$&\\
      \bottomrule
    \end{tabular}
  \end{table}
  
%
Mean and average AUC scores are reported in Table~\ref{tab:od}.
On the Thyroid dataset we were not able to run ISLV, while on the datasets where there is no value in Table~\ref{tab:od} it either did not converge in 72hrs or in 1000 iterates.
Otherwise, as for ABDA, we report the average AUC ROC w.r.t. the 100 last Gibbs samples after a burn-in of 3000 iterations.
For KDD99, we observed ABDA converging in 1000 iterates (burn-in) and within the 72hrs limit.

It is clearly visible from Table~\ref{tab:od} how ABDA can perform as good as, or even better
in most cases than
standard outlier detection models.
Moreover, this unsupervised task demonstrates
how ABDA is indeed more resilient to outliers and corrupt data
w.r.t. MSPNs.
Indeed, while the greedy structure learning procedure adopted by both
MSPNs and ABDA can get
fooled by some outliers, grouping them to inliers in the same
partition, Bayesian inference in ABDA might be able to re-assign
them to low-probability partitions.

\section{G. Exploratory pattern extraction}

Here we discuss how to employ ABDA to unsupervisedly extract patterns
among RVs---implying dependency among them---in a similar fashion to
what \emph{Association Rule Mining} (ARM) does~\cite{Agrawal1994}.
The aim is, on the one hand, to automatize the way such patterns can
be extracted, filtered and presented (as a ranking) to the user, and
on the other hand, to extract a small set able to explain to the user---in a more
interpretable way---what correlations ABDA has captured.

%
In ARM, given some data over discrete (generally assumed to be
binary for simplicity) RVs $\Xb=\{X^{1},\dots,X^{D}\}$ one is
interested in finding a set of association rules $\{\mathcal{R}_{i}\}$
where each rule $\mathcal{R}:\mathcal{A}\rightarrow\mathcal{C}$ is
composed by an antecedent, $\mathcal{A}=P_{1}\wedge P_{2},\ldots,P_{a}$, and
consequent, $\mathcal{C}=P_{a+1}\wedge P_{a+2},\ldots,P_{a+s}$, part, both of which are conjunctions of
\emph{patterns}, 
i.e. assignments to some RVs in $\Xb$.
For instance one rule might state:
$$P_{1}:X^{1}=0\wedge P_{2}:X^{3}=1\rightarrow P_{3}:X^{2}=1$$.
which would state that whenever the antecedent is observed, i.e.,
$P_{1}$ and $P_{2}$ are satisfied, it is
\emph{likely} to observe $X^{2}$ set to value 1.

%
To quantify the ``importance'' of a rule---thus having a quantitative way to
\emph{filter} and \emph{rank} rules--- one computes for a rule $\mathcal{R}$
measures like its \emph{support} and \emph{confidence}.
The former being defined as:
$$supp(\mathcal{R}:\mathcal{A}\rightarrow\mathcal{C})=\frac{\#\{P_{1},\ldots,P_{a},P_{a+1},\ldots,
  P_{r+s}\}}{N}$$
where the numerator indicates the number of samples for which the
patterns $P_{1},\ldots, P_{a+s}$ are jointly satisfied and $N$ is the
number of samples in the data.
Confidence of a rule, instead is the ratio of its support and that of
the antecedent:
$$conf(\mathcal{R}:\mathcal{A}\rightarrow\mathcal{C})=\frac{supp(\mathcal{A},\mathcal{C})}{supp(\mathcal{A})}
= \frac{\#\{P_{1},\ldots,  P_{a+s}\}}{\#\{P_{1},\ldots, P_{a}\}}$$

%
\paragraph{G1. Probabilistic patterns in ABDA}
Clearly, the notion of support is the
maximum likelihood estimation for the joint probability of its patterns
$$p(P_{1}\wedge \ldots \wedge P_{a}\wedge P_{a+1}\wedge \ldots \wedge P_{a+s})$$

Analogously, the confidence of a rule $\mathcal{R}$ is estimator for the conditional
probability
$$p(P_{a+1}\wedge \ldots\wedge P_{a+s}|P_{1}\wedge \ldots\wedge
P_{a})$$.

By properly defining what a pattern is in ABDA, we might directly
compute the above probabilities efficiently, by exploiting
marginalization over the SPN  structure of ABDA.
We also have to take into consideration how to deal with continuous
RVs natively, since
the classical formulation of ARM patterns would require binarization.

We define an \emph{interval pattern} over RV $X^{i}$ as the event $P:\pi_{low}^{i}\leq
X^{i}<\pi_{high}^{i}$ where $\pi_{low}^{i}<\pi_{high}^{i}$ are two
valid values from the domain of $X^{i}$.
The probability of a single pattern is therefore:
$$p(P:\pi_{low}^{i}\leq X^{i}<\pi_{high}^{i})
=\int_{\pi_{low}^{i}}^{\pi_{high}^{i}}f(X^{i})dX^{i}$$
where $f$ is the density function of $X^{i}$.

Consider now an dependency rule of the form $P_{1},\ldots
P_{a}\rightarrow P_{a+1},\ldots,P_{a+s}$, its support can then be
computed as multivariate integral over $P_{1}\ldots P_{a+s}$
$$\int_{\pi_{low}^{1}}^{\pi_{high}^{1}}\ldots\int_{\pi_{low}^{a+s}}^{\pi_{high}^{a+s}}f(\mathbf{X})dX^{1}\ldots X^{a+s}$$

If the joint density $f$ decomposes as an SPN structure, solving the above
integral would require resolving univariate integrals at the
leaves---which is doable assuming tractable univariate distributions
there as we do in ABDA\footnote{Even if some densities would require
  to approximate it, it would be still doable}---
and propagate the computed probabilities upwards.

By doing so we have a way to exploit the SPN structure in ABDA to
efficiently compute the support of a rule---a collection of dependency
pattern---here extended to continuous RVs.
Therefore we can rank rules by computing their support.
How to extract rules in a (semi-)automatic way?

\paragraph{G2. Automatic pattern mining in ABDA}

The most straightforward approach to extract patterns and rules via
ABDA would mimic \emph{Apriori}, the stereotypical ARM algorithm.
In a nutshell, first patterns of length 1 are mined (i.e., involving a
single RV), the collection of patterns are combined by enumeration
while at the
same time filtering out patterns whose support is less than a
user-specified threshold $\rho$~\cite{Agrawal1994}.

The main issue would be how to determine the atomic patterns in the
form $P:\pi_{low}^{i}\leq
X^{i}<\pi_{high}^{i}$, since we $X^{i}$ can also be continuous, and
hence we can possibly find
an infinite number of intervals $[\pi_{low}^{i}, \pi_{high}^{i})$ from its domain.
The solution comes from ABDA having already applied this partitioning
during inference.
Indeed, SPN leaves in ABDA already describe tractable distributions
\emph{concentrated} on a portion of the whole domain for a feature.
Given a user defined percentile threshold $\lambda\in[0,1]$, we can determine the
interval containing $\lambda$ percentage of the probability mass of
density $f(X^{i})$.
For instance, for $\lambda=90\%$, we might easily find
$[\pi_{low}^{i}, \pi_{high}^{i})$ as the the $5\%$ and
$95\%$percentiles of $f(X^{i})$.

\begin{figure}[!t]
    \includegraphics[width=0.26\columnwidth]{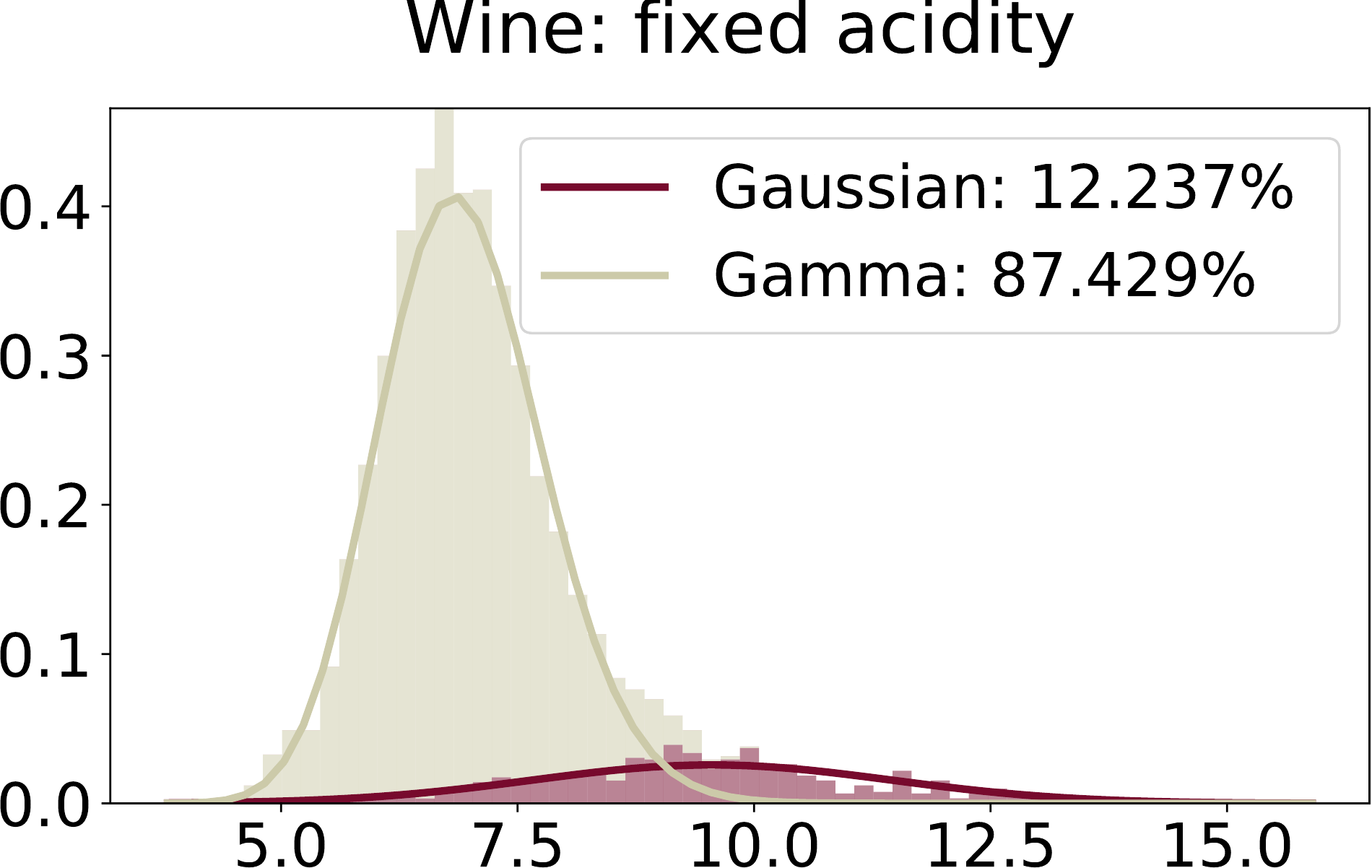}\hspace{10pt}
    \includegraphics[width=0.25\columnwidth]{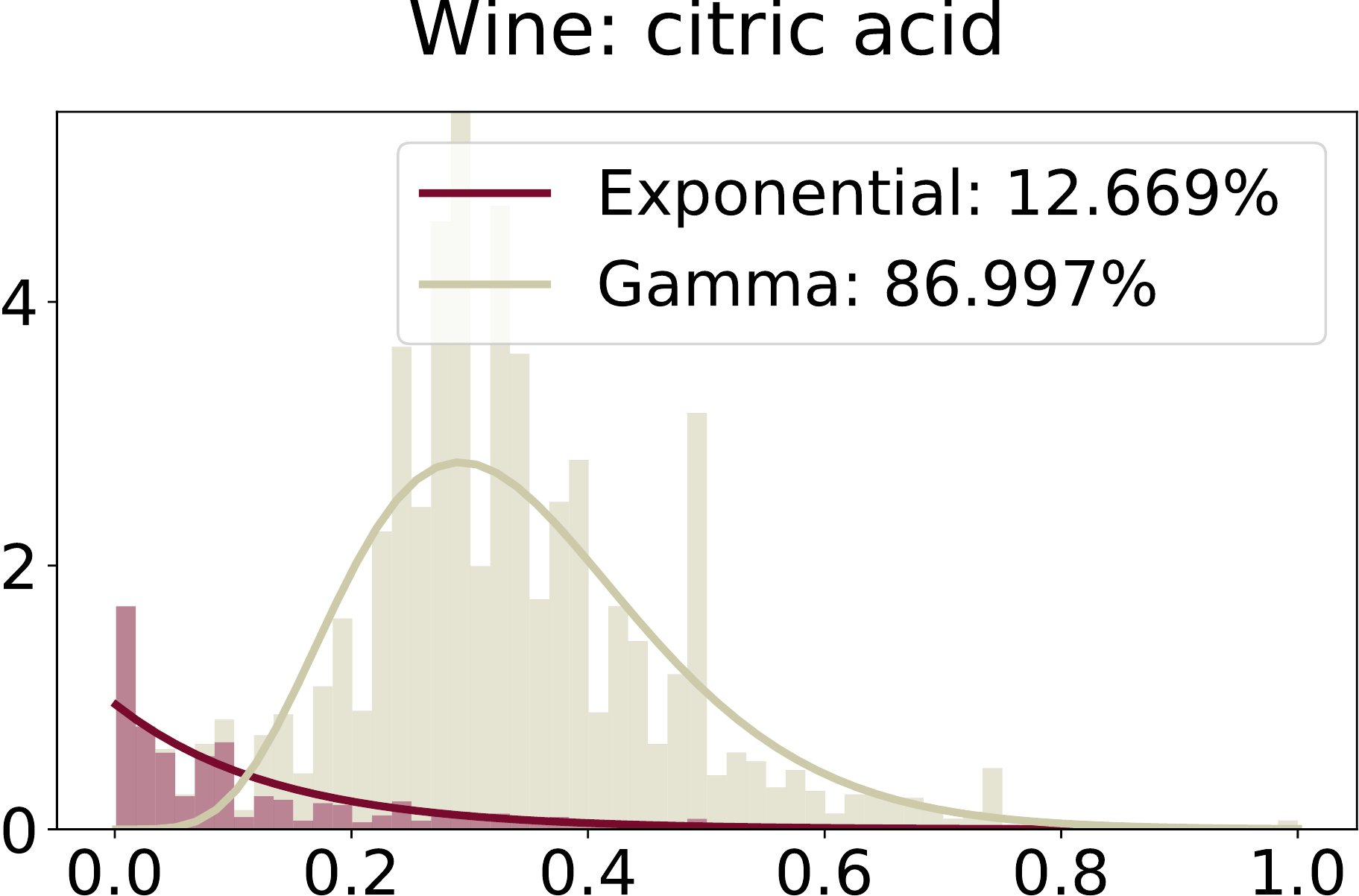}\\
    \caption{Data exploration on Wine data. Density estimation provided by ABDA on the Wine quality dataset. ABDA identifies the two modalities in the data induced by red and white wine over three different features. }
    \label{fig:parm}
\end{figure}

\paragraph{G3. Wine data}
As a concrete example, see Figure~\ref{fig:parm}, depicting two
marginal distributions as learned by ABDA for two features of the Wine
dataset, $X^{1}=\mathsf{FixAcid}$ and $X^{2}=\mathsf{CitAcid}$.
One might extract the following four patterns from them by fixing a threshold of $80\%$ of probability:

$$P_{1}:5.8 \leq {\color{wwine}\mathsf{FixAcid}} < 8.1,\quad
{\color{wwine}\mathsf{FixAcid}}\sim\mathcal{N}$$
$$ P_{2}:0.2\leq {\color{wwine}\mathsf{CitAcid}} <
0.5,\quad {\color{wwine}\mathsf{CitAcid}}\sim Gamma$$
$$P_{3}:7.1 \leq {\color{rwine}\mathsf{FixAcid}} < 12.0,\quad
{\color{rwine}\mathsf{FixAcid}}\sim Gamma$$
$$P_{4}:0.0\leq {\color{rwine}\mathsf{CitAcid}} <
0.3,\quad{\color{rwine}\mathsf{CitAcid}}\sim Exp$$

After these atomic patterns have been extracted from leaves, one can
first determine their support according to the whole SPN $\SPN$
Then, conjunctions of patterns may be mechanically combined as in
Apriori, their support computed and filtered out if it is lower than
the user-defined threshold $\rho$.

Back to the Wine features in Figure~\ref{fig:parm}, one could compose
the following composite pattern via conjunctions, by noting that the extracted patterns are
belonging to product node children and thus referring to samples
belonging the same partition (same color across histograms):

$$P_{1}:5.2 \leq {\color{wwine}\mathsf{FixAcid}} < 8.1\wedge P_{2}:0.2\leq {\color{wwine}\mathsf{CitAcid}} <
0.5$$
$$P_{3}:7.1 \leq {\color{rwine}\mathsf{FixAcid}} < 12.0\wedge P_{4} : 0.0\leq {\color{rwine}\mathsf{CitAcid}} < 0.3$$

\begin{figure*}[!t]
    \includegraphics[width=0.3\columnwidth]{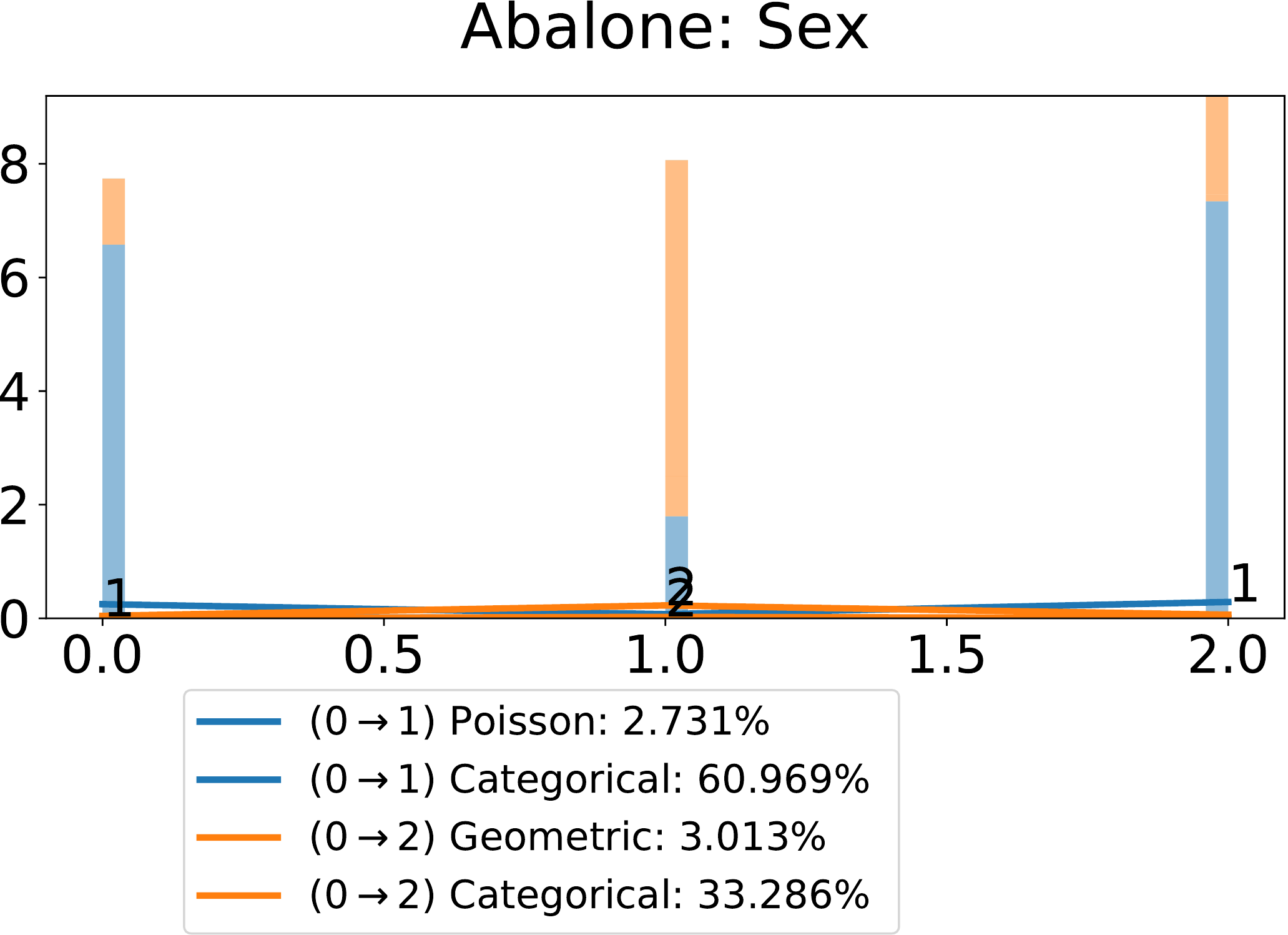}\hspace{10pt}
    \includegraphics[width=0.3\columnwidth]{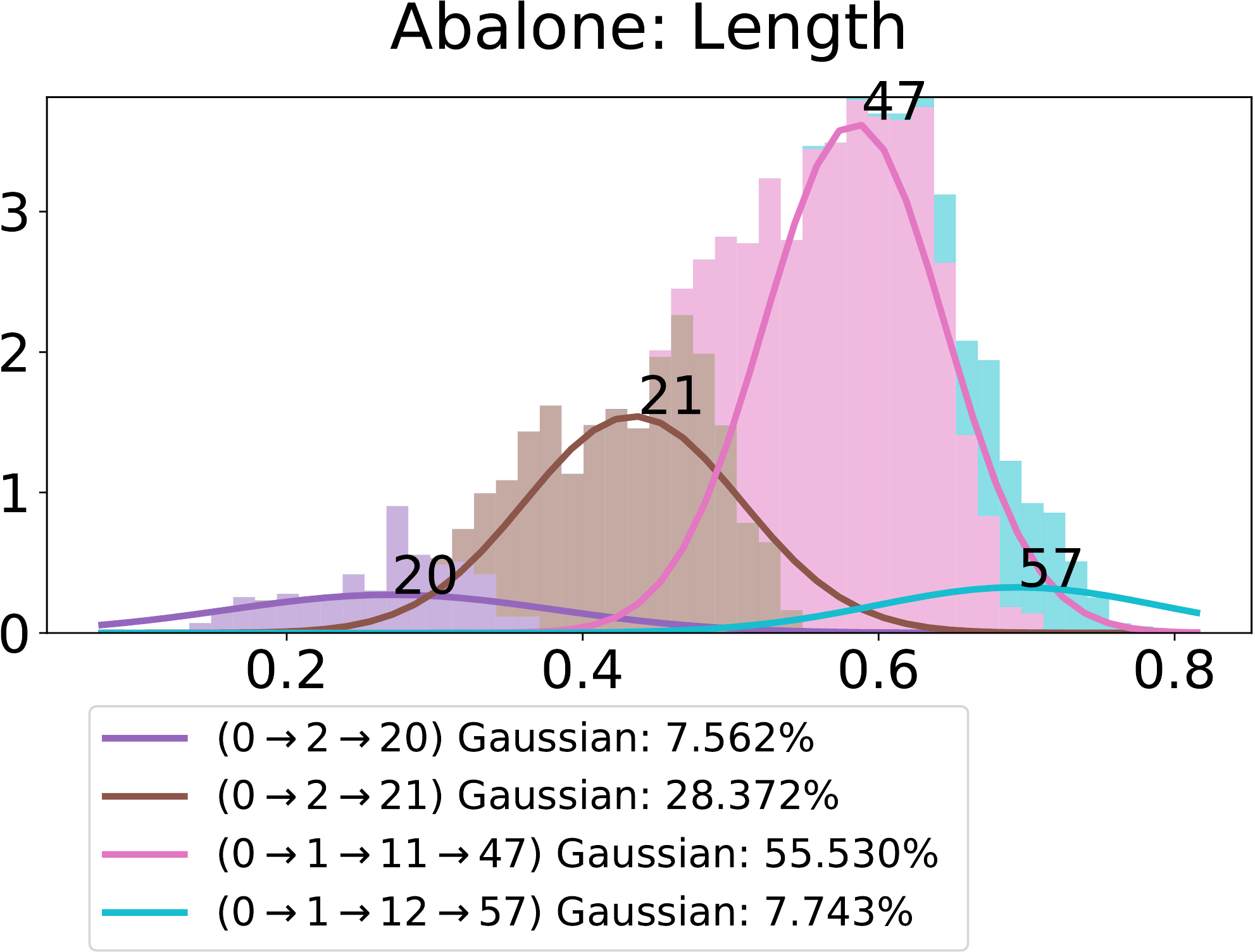}\hspace{10pt}
    \includegraphics[width=0.3\columnwidth]{abalone-d2-fit-crop.pdf}\\[10pt]
    \includegraphics[width=0.3\columnwidth]{abalone-d3-fit-crop.pdf}\hspace{5pt}
    \includegraphics[width=0.31\columnwidth]{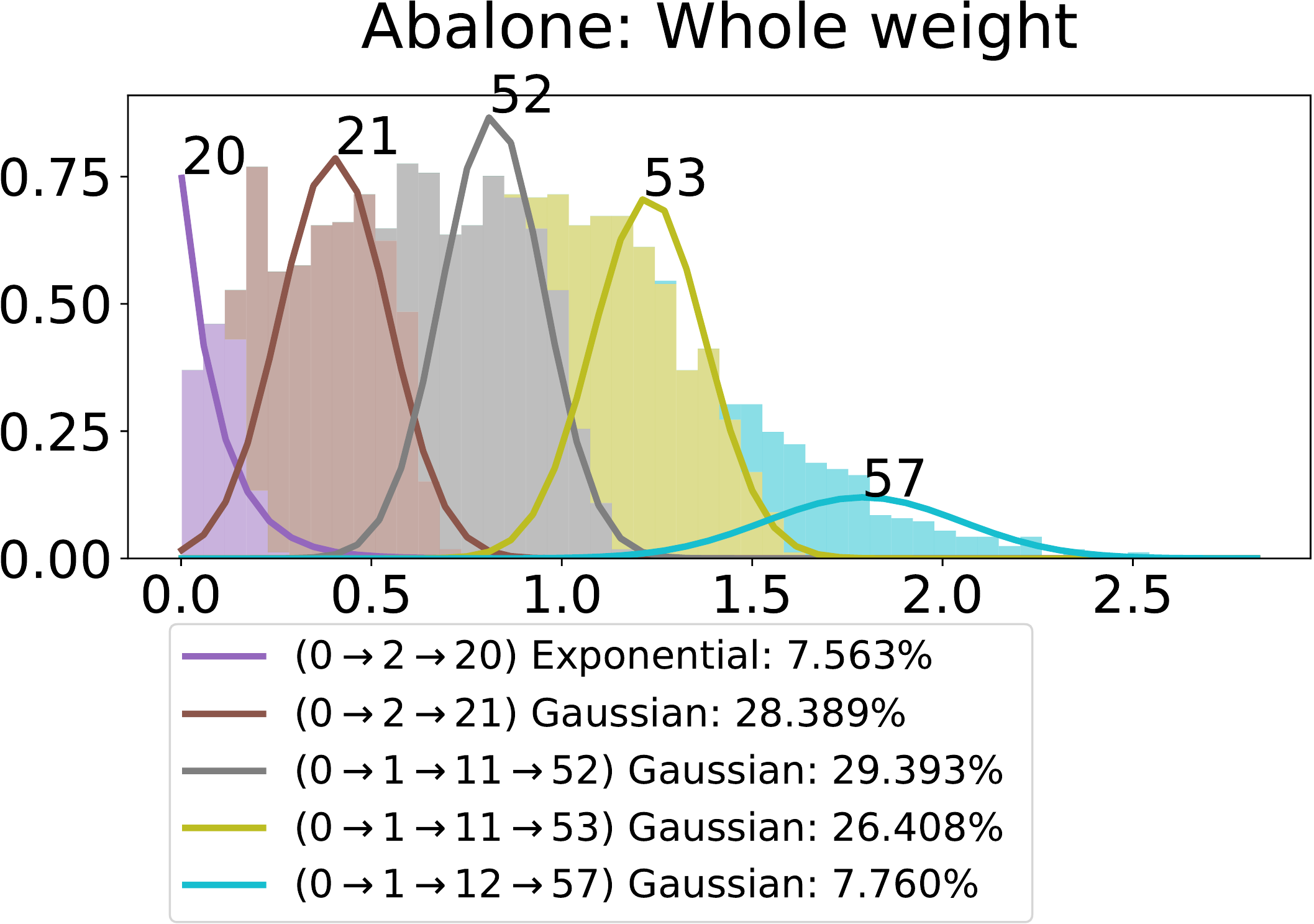}\hspace{7pt}
    \includegraphics[width=0.31\columnwidth]{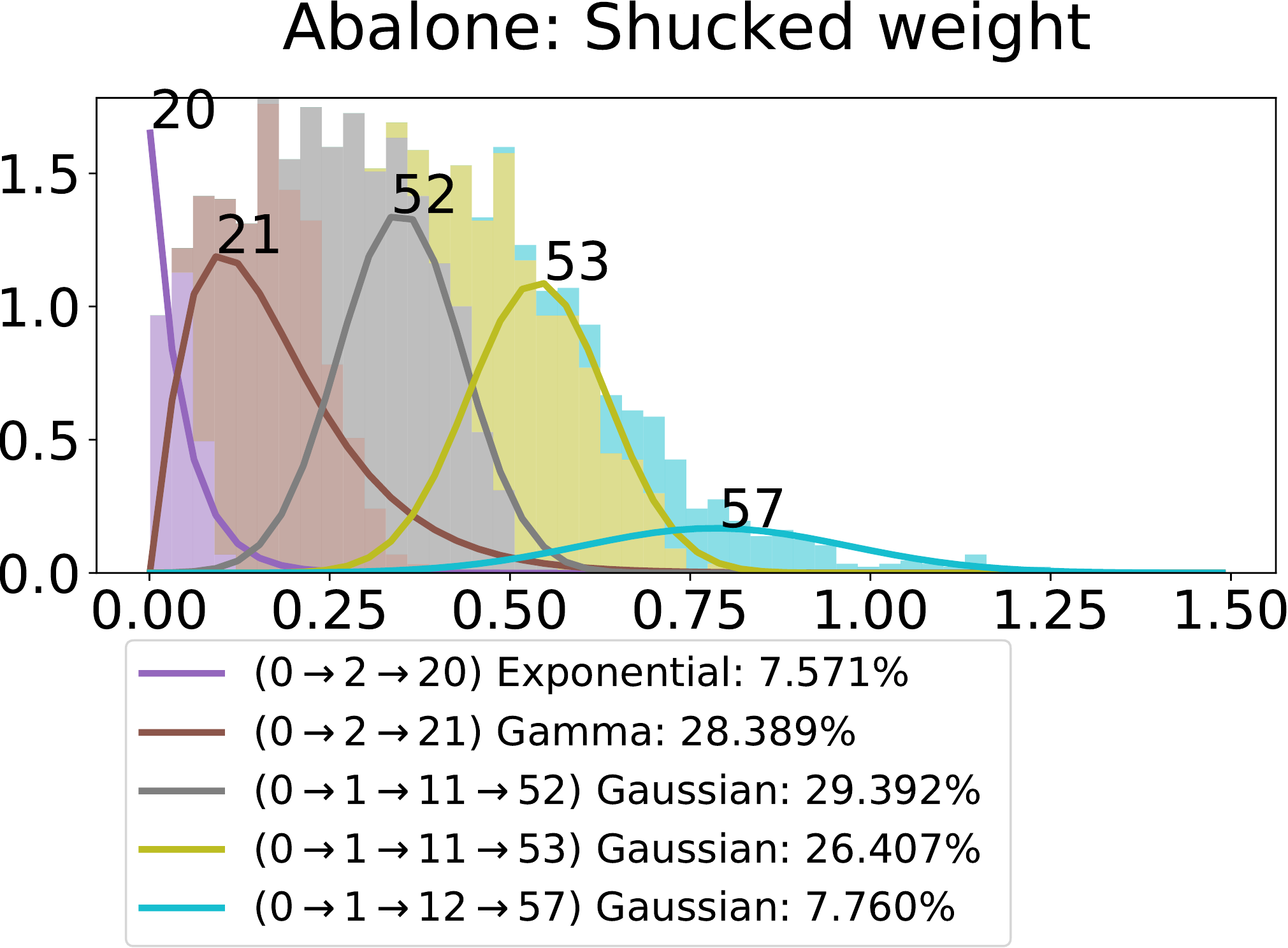}\\[10pt]
    \includegraphics[width=0.3\columnwidth]{abalone-d6-fit-crop.pdf}\hspace{10pt}
    \includegraphics[width=0.3\columnwidth]{abalone-d7-fit-crop.pdf}\hspace{5pt}
    \includegraphics[width=0.3\columnwidth]{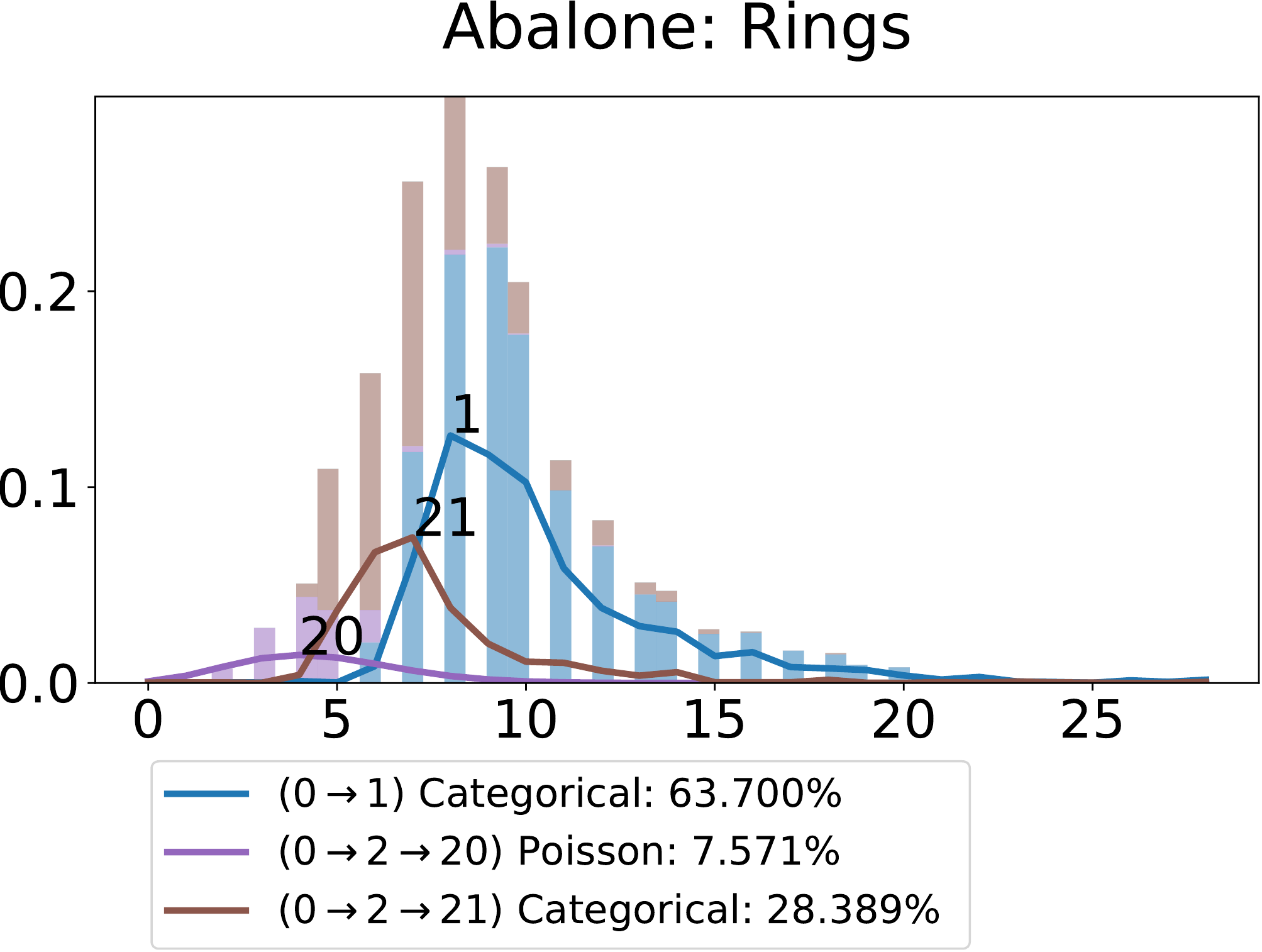}
    \caption{Data exploration on Abalone. Density estimation provided
      by ABDA on the Abalone UCI dataset, showing the extracted
      correlation patterns among sub-populations across different
      features like shell and viscera weights, sex and shell diameter
      and length.
      Equally colored densities model samples assigned by ABDA to the
      same sub-population (partition), while their numerical label
      indicates the position of such partition in the hierarchical
      decomposition discovered (paths are represented in parenthesis
      as sequences of labeled partitions, starting from the initial
      partition, representing the root of the hierarchy.)}
    \label{fig:parm-aba}
\end{figure*}

\paragraph{G4. Abalone data} 
For a more complex example, we employ ABDA on the Abalone dataset (see Appendix D).
%
The dataset contains physical measurements (features) over abalone samples, namely: the Sex of the specimen ('male', 'female' or 'infant'), its Length, Diameter, Height, Whole weight, Shucked Weight and Viscera Weight, its Shell Weight and the number of rings in it.

We employ ABDA on it by running it for 5000 iterates, stopping the SPN LV structure learning until 10\% of the data was reached and employing $0.7$ as the RDC independence threshold.

Figure~\ref{fig:parm-aba} shows the marginal distributions for all features upon which the densities as fit by ABDA. 
Each density is colored by a unique color---shared across all features---indicating the partition $\mathcal{K}$ induced by the SPN $\SPN$.
Each of such partitions is also labeled with an integer, which serves the purpose to indicate the path---appearing in each legend entry---inside $\SPN$ leading from the top partition (numbered as 0, indicating the whole data matrix) to the finer grained one.
For instance, the path $0\rightarrow 2\rightarrow 20$ appearing in the legend of the Length attribute, and associated to the purple partition numbered 20, indicates that such a  partition is contained in the partition number 2 which in turn belongs to the initial partition.

By starting from the Sex feature, one can see that two partitions are discovered (1 and 2) in which the first mostly represents samples belonging to the first and last mode of the categorical features, indicating 'male' and 'infant' abalone samples, while the second comprises mostly 'female' individuals.
From these initial \textit{conditioning} on the sex, all other feature correlations are characterized.

For instance, the green partition (11) is a sub-partition of the 'Male' clustering, and shows a cross correlation across the Height feature and the Weight one, as one would expect.
Additionally, the green partition itself later on splits into two sub-population, the gray (52) and yellow (53) representing correlations among the Shucked and Viscera weights, indicating a mode in the abalone population for specific ranges of these features. 

By looking at all partitions and at each density belonging to them, one can build in a mechanical way  (composite) patterns and rank them by their support.
For this analysis of the Abalone data, these are the top 5 patterns that the ABDA automatically extracts:
\begin{align*}
    \mathcal{P}_{1}:\quad 0.088&\leq {\color{abagreen}\mathsf{Height}}< 0.224\quad\wedge\quad 
    0.158\leq {\color{abagreen}\mathsf{ShellWeight}}< 0.425 \quad(supp(\mathcal{P}_{1})=0.507)\\
    \mathcal{P}_{2}:\quad 0.483&\leq {\color{abapink}\mathsf{Length}}< 0.684\quad\wedge\quad 
    0.364\leq {\color{abapink}\mathsf{Diameter}}< 0.547 \quad(supp(\mathcal{P}_{2})=0.489)\\
    \mathcal{P}_{3}:\quad 0.596&\leq {\color{abagray}\mathsf{WholeWeight}}< 1.040\quad\wedge\quad 
    0.205\leq {\color{abagray}\mathsf{ShuckedWeight}}< 0.491 \quad\wedge\\
    0.076&\leq {\color{abagray}\mathsf{VisceraWeight}}< 0.281 \quad(supp(\mathcal{P}_{3})=0.223)\\
    \mathcal{P}_{4}:\quad 0.596&\leq {\color{abayellow}\mathsf{WholeWeight}}< 1.040\quad\wedge\quad 
    0.205\leq {\color{abayellow}\mathsf{ShuckedWeight}}< 0.491 \quad\wedge\\
    0.076&\leq {\color{abayellow}\mathsf{VisceraWeight}}< 0.281 \quad(supp(\mathcal{P}_{4})=0.202)\\
    \mathcal{P}_{5}:\quad 0.313&\leq {\color{ababrown}\mathsf{Length}}< 0.554\quad\wedge\quad 
     0.230\leq {\color{ababrown}\mathsf{Diameter}}< 0.438\quad\wedge\\ 
    \quad 0.023&\leq {\color{ababrown}\mathsf{Height}}< 0.197\quad\wedge\quad 
     0.165\leq {\color{ababrown}\mathsf{WholeWeight}}< 0.639\quad\wedge\\ 
    0.037&\leq {\color{ababrown}\mathsf{ShuckedWeight}}< 0.408 \quad\wedge\quad
    0.019\leq {\color{ababrown}\mathsf{VisceraWeight}}<0.192 \quad\wedge\\
    0.027&\leq {\color{ababrown}\mathsf{ShellWeight}}< 0.27 \quad\wedge
    \quad 5\leq {\color{ababrown}\mathsf{Rings}}<13\quad(supp(\mathcal{P}_{5})=0.111)\\
\end{align*}

\section{H. ABDA vs MSPN: accuracy and robustness}

In our experimental section, we selected diverse
datasets---w.r.t. size and feature heterogeneity---from both the ISLV
and MSPN original papers with the aim to have a common, fair
experimental setting.
Moreover, we run for ABDA and MSPN a grid search in the
same hyperparameter space, \textit{to block the effect of building the SPN LV
hierarchy}.
Striving for automatic density analysis, such a grid search has been
limited only to the independence test (RDC) threshold parameter, while
original MSPNs were validated over more than five parameters.

For a more straightforward comparison in the original experimental
setting of \cite{Molina2017b}, we cross-validate ABDA also on one additional parameter, $m\in\{20\%, 10\%, 5\%\}$, the minimum number of samples in a partition (as a percentage over the whole number of samples) to stop the learning process to recursively further partition it.
This parameter governs the degree of overparametrization of the learned LV structure.

Table \ref{tab:mspn-head-on} reports the average test log-likelihoods for the best ABDA model on the validation set, while MSPN results, considering unimodal isotonic regression applied to piecewise linear leaf distribution approximations, are directly taken from \cite{Molina2017b}.
As one can see, \textit{ABDA provides competitive results to  MSPNs} even in this experimental setting, {still requiring less parameter tuning}. 
This is due to i) the likelihood mixtures in ABDA better generalizing
than piecewise models---since their support is limited to samples seen
during training and possibly infinite tails of a distribution are not natively captured---and ii)
our Bayesian inference providing indeed robust models.

\begin{table}[!t]
\caption{Density estimation by ABDA vs MSPN over the original 14
  datasets.
  Best average test log-likelihoods in bold.}
\label{tab:mspn-head-on}
\setlength{\tabcolsep}{2pt}
\small
    \begin{tabular}{r r r r r r}
    \toprule
    &\textbf{ABDA}&\textbf{MSPN}&&\textbf{ABDA}&\textbf{MSPN}\\
     \midrule
      \textsf{Anneal-U}&\textbf{-2.65}&-38.31&\textsf{Austr.}&\textbf{-17.70}&-31.02\\
      \textsf{Auto}&-70.62&\textbf{-70.06}&\textsf{Bal-scale}&\textbf{-7.13}&-7.30\\
      \textsf{Breast}&-25.46&\textbf{-24.04}&\textsf{Bre.-cancer}&\textbf{-9.61}&-9.99\\
      \textsf{Cars}&-35.04&\textbf{-30.52}&\textsf{Cleve}&\textbf{-22.60}&-25.44\\
      \textsf{Crx}&\textbf{-15.53}&-31.72&\textsf{Diabete}&\textbf{-17.48}&-27.24\\
      \textsf{German}&\textbf{-32.10}&-32.36&\textsf{Ger.-org}&\textbf{-26.29}&-27.29\\
      \textsf{Heart}&\textbf{-23.39}&-25.90&\textsf{Iris}&-2.96&\textbf{-2.84}\\
      \bottomrule
    \end{tabular}
  \end{table}
  
To get a better understanding of how Bayesian inference in ABDA affects its robustness, we compare ABDA and MSPNs in an inductive scenario when we provide both with increasingly parametrized LV structures.
More specifically, for both we explore deeper and deeper structures by letting the minimum sample percentage parameter $m$ vary in $\{5\%, 1\%, 0.5\%\}$, while we fix the RDC independence threshold to $0.5$.
As one could expect, ABDA is at advantage since it performs an additional parameter learning step after the structure is provided.
Nevertheless, it is worth measuring by \textit{how much more robust} ABDA models are to an overparametrization than MSPNs and, at the same time, evaluate \textit{how much sparser} the SPN structure in ABDA gets, that is, quantifying how ``blindly'' one user can run ABDA with limited or no possibility to cross-validate.

To this end, we measure the relative percentage of decreasing mean test log-likelihoods w.r.t. the mean test log-likelihood achieved by each model in the case in which $m=5\%$.
Moreover, we measure the sparsity of a SPN structure in ABDA as the ratio between the number of \textit{relevant} product nodes and leaf likelihood components over the number of all nodes and likelihood models.
In this case, the relevancy of a node is measured by the fact that at least some samples have been associated to the corresponding node (or likelihood component) partition.

Table~\ref{tab:mspn-rob} reports these values for the 12 datasets involved in our density estimation and imputation experiments.
Firstly, one can see that on the Chess and German datasets, growing a deeper SPN structure is not even possible, in general.
However, for all other datasets, it is clear that MSPN accuracy tends to drop more quickly than ABDA's, whose test predictions are not only more stable for different values of $m$ but sometimes even better.
Lastly, the sparsity of the SPN structure in ABDA is indeed significantly increasing as $m$ becomes smaller\footnote{For a reference, consider that the same structure in an MSPN is never sparse, since a partition in the training data must contain at least one sample. Therefore the sparsity level for $m=5\%$ is already meaningful for our investigation.}.

\begin{table}[!t]
\caption{Robustness of MSPN and ABDA w.r.t. overparametrized structures, i.e. when the minimum percentage of samples ($m$) to split a partition is let vary up to $0.5\%$
  Best relative test log-likelihoods improvement w.r.t. test log likelihoods for $m=5\%$ is reported in bold (in parenthesis).
  The last column reports the sparsity of ABDA structures ($s$) after inference, for each setting.}
\label{tab:mspn-rob}
\setlength{\tabcolsep}{2pt}
\footnotesize
    \begin{tabular}{r r r l r r l r l}
    \toprule
    dataset&$m$&\textsf{MSPN}&&\textsf{ABDA}&&\textsf{s}\\
    \midrule
\multirow{3}{*}{\textsf{abalone}}&$5\%$&10.33&&       4.94&&0.15\\              
&$1\%$&10.16&($\mathbf{-1.68}\%$)&                      4.74&($-4.13\%$)&0.04\\   
&$0.5\%$&9.59&($-7.15\%$)&                       5.00&($\mathbf{+1.30}\%$)&0.03\\
\midrule
\multirow{3}{*}{chess}&$5\%$&-16.21&&        -12.54&&0.66\\            
&$1\%$&-16.21&($\mathbf{-0.00}\%$)&                     -12.54&($\mathbf{-0.00}\%$)&0.69\\ 
&$0.5\%$&-16.21&($\mathbf{-0.00}\%$)&                     -12.54&($-0.00\%$)&0.69\\
\midrule
\multirow{3}{*}{german}&$5\%$&-31.85&&       -25.86&&0.56\\            
&$1\%$&-31.85&($\mathbf{-0.00}\%$)&                     -26.21&($-1.35\%$)&0.45\\ 
&$0.5\%$&-31.85&($\mathbf{-0.00}\%$)&                     -25.87&($\mathbf{-0.05}\%$)&0.59\\
\midrule
\multirow{3}{*}{student}&$5\%$&-36.86&&      -28.93&&0.25\\            
&$1\%$&-44.86&($-21.70\%$)&                    -29.16&($\mathbf{-0.78}\%$)&0.09\\ 
&$0.5\%$&-49.95&($-35.49\%$)&                    -29.32&($\mathbf{-1.35}\%$)&0.05\\
\midrule
\multirow{3}{*}{wine}&$5\%$&-0.10&&          -9.20&&0.14\\             
&$1\%$&-0.31&($-198.38\%$)&                    -8.43&($\mathbf{+8.37}\%$)&0.17\\  
&$0.5\%$&-0.51&($-391.74\%$)&                    -8.73&($\mathbf{+5.05}\%$)&0.07\\
\midrule
\multirow{3}{*}{dermat.}&$5\%$&-35.77&&  -24.96&&0.36\\            
&$1\%$&-46.82&($-30.91\%$)&                    -25.14&($\mathbf{-0.72}\%$)&0.20\\ 
&$0.5\%$&-57.42&($-60.54\%$)&                    -25.16&($\mathbf{-0.78}\%$)&0.14\\
\midrule
\multirow{3}{*}{anneal-U}&$5\%$&-151.09&&    7.68&&0.12\\              
&$1\%$&-156.60&($\mathbf{-3.64}\%$)&                    3.80&($-50.53\%$)&0.14\\  
&$0.5\%$&-166.26&($\mathbf{-10.04}\%$)&                   4.33&($-43.53\%$)&0.08\\
\midrule
\multirow{3}{*}{austral.}&$5\%$&-37.67&&   -15.89&&0.27\\            
&$1\%$&-38.93&($-3.32\%$)&                     -16.27&($\mathbf{-2.38}\%$)&0.21\\ 
&$0.5\%$&-40.08&($-6.37\%$)&                     -16.36&($\mathbf{-2.98}\%$)&0.13\\
\midrule
\multirow{3}{*}{autism}&$5\%$&-41.16&&       -27.69&&0.23\\            
&$1\%$&-42.30&($-2.78\%$)&                    -27.60&($\mathbf{+0.35}\%$)&0.12\\ 
&$0.5\%$&-42.54&($-3.34\%$)&                    -27.33&($\mathbf{+1.29}\%$)&0.10\\
\midrule
\multirow{3}{*}{breast}&$5\%$&-33.42&&       -25.52&&0.23\\            
&$1\%$&-34.87&($-4.32\%$)&                     -26.01&($\mathbf{-1.89}\%$)&0.20\\ 
&$0.5\%$&-35.84&($-7.23\%$)&                     -25.64&($\mathbf{-0.46}\%$)&0.10\\
\midrule
\multirow{3}{*}{crx}&$5\%$&-37.57&&          -12.86&&0.45\\            
&$1\%$&-38.79&($-3.26\%$)&                     -13.03&($\mathbf{-1.37}\%$)&0.18\\ 
&$0.5\%$&-39.39&($-4.87\%$)&                     -13.00&($\mathbf{-1.07}\%$)&0.14\\
\midrule
\multirow{3}{*}{diabetes}&$5\%$&-31.55&&     -16.47&&0.41\\            
&$1\%$&-31.70&($\mathbf{-0.45}\%$)&                     -18.96&($-15.07\%$)&0.31\\
&$0.5\%$&-31.65&($\mathbf{-0.31}\%$)&                     -17.91&($-8.75\%$)&0.27\\
\midrule
\multirow{3}{*}{adult}&$5\%$&-74.98&&    -5.50&&0.28\\             
&$1\%$&-76.15&($-1.55\%$)&                     -5.37&($\mathbf{+2.50}\%$)&0.17\\  
&$0.5\%$&-77.25&($-3.03\%$)&                     -5.36&($\mathbf{+2.62}\%$)&0.13\\  
\bottomrule
    \end{tabular}
  \end{table}

%
\bibliographystyle{plain}
\bibliography{referomnia}

\begin{thebibliography}{10}

\bibitem{Agrawal1994}
Rakesh Agrawal and Ramakrishnan Srikant.
\newblock Fast algorithms for mining association rules.
\newblock In {\em Proceedings of VLDB}, volume 1215, pages 487--499, 1994.

\bibitem{Breunig2000}
Markus~M Breunig, Hans-Peter Kriegel, Raymond~T Ng, and J{\"o}rg Sander.
\newblock {LOF}: identifying density-based local outliers.
\newblock In {\em ACM sigmod record}, volume~29, pages 93--104, 2000.

\bibitem{Chan2006}
Hei Chan and Adnan Darwiche.
\newblock On the robustness of most probable explanations.
\newblock In {\em Proceedings of the Twenty-Second Conference on Uncertainty in
  Artificial Intelligence}, UAI'06, pages 63--71, Arlington, Virginia, United
  States, 2006. AUAI Press.

\bibitem{Chandola2009}
Varun Chandola, Arindam Banerjee, and Vipin Kumar.
\newblock Anomaly detection: A survey.
\newblock {\em ACM Comput. Surv.}, 41(3):15:1--15:58, July 2009.

\bibitem{Choi2017}
Arthur Choi and Adnan Darwiche.
\newblock On relaxing determinism in arithmetic circuits.
\newblock In {\em Proceedings of ICML}, pages 825--833, 2017.

\bibitem{Cowles1996}
Mary~Kathryn Cowles and Bradley~P Carlin.
\newblock Markov chain monte carlo convergence diagnostics: a comparative
  review.
\newblock {\em Journal of the American Statistical Association},
  91(434):883--904, 1996.

\bibitem{Darwiche2003}
Adnan Darwiche.
\newblock A differential approach to inference in {Bayesian} networks.
\newblock {\em Journal of the ACM (JACM)}, 50(3):280--305, 2003.

\bibitem{Darwiche2009}
Adnan Darwiche.
\newblock {\em Modeling and Reasoning with Bayesian Networks}.
\newblock Cambridge, 2009.

\bibitem{Dennis2012}
Aaron Dennis and Dan Ventura.
\newblock Learning the architecture of sum-product networks using clustering on
  variables.
\newblock In {\em Proceedings of NIPS}, pages 2033--2041, 2012.

\bibitem{Dennis2015}
Aaron Dennis and Dan Ventura.
\newblock {Greedy Structure Search for Sum-product Networks}.
\newblock In {\em IJCAI'15}, pages 932--938. AAAI Press, 2015.

\bibitem{dheeru2017uci}
Dua Dheeru and Efi~Karra Taniskidou.
\newblock Uci machine learning repository.
\newblock {\em University of California, Irvine, School of Information and
  Computer Sciences}, 2017.

\bibitem{duvenaudLGTG13}
David~K. Duvenaud, James~Robert Lloyd, Roger~B. Grosse, Joshua~B. Tenenbaum,
  and Zoubin Ghahramani.
\newblock Structure discovery in nonparametric regression through compositional
  kernel search.
\newblock In {\em Proceedings of ICML}, pages 1166--1174, 2013.

\bibitem{Gens2012}
Robert Gens and Pedro Domingos.
\newblock Discriminative learning of sum-product networks.
\newblock In {\em Proceedings of NIPS}, pages 3239--3247, 2012.

\bibitem{Gens2013}
Robert Gens and Pedro Domingos.
\newblock Learning the structure of sum-product networks.
\newblock In {\em Proceedings of ICML}, pages 873--880, 2013.

\bibitem{Ghahramani2000}
Zoubin Ghahramani and Matthew~J Beal.
\newblock Variational inference for {Bayesian} mixtures of factor analysers.
\newblock In {\em Proceedings of NIPS}, pages 449--455, 2000.

\bibitem{Goldstein2012}
Markus Goldstein and Andreas Dengel.
\newblock Histogram-based outlier score ({HBOS}): A fast unsupervised anomaly
  detection algorithm.
\newblock {\em KI-2012}, pages 59--63, 2012.

\bibitem{Goldstein2016}
Markus Goldstein and Seiichi Uchida.
\newblock A comparative evaluation of unsupervised anomaly detection algorithms
  for multivariate data.
\newblock {\em PloS one}, 11(4), 2016.

\bibitem{Guyon2016}
I.~Guyon, I.~Chaabane, J.~H. Escalante, and S.~Escalera.
\newblock A brief review of the chalearn automl challenge: Any-time any-dataset
  learning without human intervention.
\newblock In {\em ICML Workshop on AutoML}, 2016.

\bibitem{lloydDGTG14}
James~Robert Lloyd, David~K. Duvenaud, Roger~B. Grosse, Joshua~B. Tenenbaum,
  and Zoubin Ghahramani.
\newblock Automatic construction and natural-language description of
  nonparametric regression models.
\newblock In {\em AAAI}, 2014.

\bibitem{LopezPaz2013}
David Lopez-Paz, Philipp Hennig, and Prof.~Bernhard Sch\"{o}lkopf.
\newblock The randomized dependence coefficient.
\newblock In {\em NIPS}, 2013.

\bibitem{Mansinghka2016}
V.~Mansinghka, P.~Shafto, E.~Jonas, C.~Petschulat, M.~Gasner, and J.~B
  Tenenbaum.
\newblock Crosscat: A fully bayesian nonparametric method for analyzing
  heterogeneous, high dimensional data.
\newblock {\em JMLR}, 2016.

\bibitem{Molina2017a}
Alejandro Molina, Sriraam Natarajan, and Kristian Kersting.
\newblock Poisson sum-product networks: {A} deep architecture for tractable
  multivariate poisson distributions.
\newblock In {\em AAAI}, 2017.

\bibitem{Molina2017b}
Alejandro Molina, Antonio Vergari, Nicola {Di Mauro}, Sriraam Natarajan,
  Floriana Esposito, and Kristian Kersting.
\newblock Mixed sum-product networks: A deep architecture for hybrid domains.
\newblock In {\em AAAI}, 2018.

\bibitem{Murphy2012}
Kevin~P. Murphy.
\newblock {\em Machine Learning: A Probabilistic Perspective}.
\newblock MIT Press, 2012.

\bibitem{Peharz2017}
Robert Peharz, Robert Gens, Franz Pernkopf, and Pedro Domingos.
\newblock On the latent variable interpretation in sum-product networks.
\newblock {\em IEEE Transactions on Pattern Analysis and Machine Intelligence},
  39(10):2030--2044, 2017.

\bibitem{Peharz2015a}
Robert Peharz, Sebastian Tschiatschek, Franz Pernkopf, and Pedro Domingos.
\newblock On theoretical properties of sum-product networks.
\newblock {\em AISTATS}, 2015.

\bibitem{Poon2011}
Hoifung Poon and Pedro Domingos.
\newblock {Sum-Product Networks: a New Deep Architecture}.
\newblock {\em UAI 2011}, 2011.

\bibitem{Pronobis2017}
Andrzej Pronobis, Francesco Riccio, and Rajesh~PN Rao.
\newblock Deep spatial affordance hierarchy: Spatial knowledge representation
  for planning in large-scale environments.
\newblock In {\em ICAPS 2017 Workshop}, 2017.

\bibitem{Rooshenas2014}
Amirmohammad Rooshenas and Daniel Lowd.
\newblock {Learning Sum-Product Networks with Direct and Indirect Variable
  Interactions}.
\newblock In {\em ICML}, 2014.

\bibitem{Scholkopf2001}
Bernhard Sch\"{o}lkopf, John~C. Platt, John~C. Shawe-Taylor, Alex~J. Smola, and
  Robert~C. Williamson.
\newblock Estimating the support of a high-dimensional distribution.
\newblock {\em Neural Comput.}, 13(7):1443--1471, July 2001.

\bibitem{Trapp2017}
Martin Trapp, Tamas Madl, Robert Peharz, Franz Pernkopf, and Robert Trappl.
\newblock Safe semi-supervised learning of sum-product networks.
\newblock {\em UAI}, 2017.

\bibitem{Valera2017b}
I.~{Valera}, M.~F. {Pradier}, and Z.~{Ghahramani}.
\newblock {General Latent Feature Models for Heterogeneous Datasets}.
\newblock {\em ArXiv e-prints}, June 2017.

\bibitem{Valera2017a}
Isabel Valera and Zoubin Ghahramani.
\newblock Automatic discovery of the statistical types of variables in a
  dataset.
\newblock In {\em ICML}, pages 3521--3529, 06--11 Aug 2017.

\bibitem{Vergari2015}
Antonio Vergari, Nicola {Di Mauro}, and Floriana Esposito.
\newblock {Simplifying, Regularizing and Strengthening Sum-Product Network
  Structure Learning}.
\newblock In {\em ECML-PKDD 2015}, 2015.

\bibitem{Vergari2018}
Antonio Vergari, Nicola {Di Mauro}, and Floriana Esposito.
\newblock Visualizing and understanding sum-product networks.
\newblock {\em MLJ}, 2018.

\bibitem{Vergari2017}
Antonio Vergari, Robert Peharz, Nicola {Di Mauro}, Alejandro Molina, Kristian
  Kersting, and Floriana Esposito.
\newblock Sum-product autoencoding: Encoding and decoding representations using
  sum-product networks.
\newblock In {\em AAAI}, 2018.

\bibitem{Zhao2016b}
Han Zhao, Pascal Poupart, and Geoffrey~J Gordon.
\newblock A unified approach for learning the parameters of sum-product
  networks.
\newblock In {\em NIPS}, pages 433--441. Curran Associates, Inc., 2016.

\end{thebibliography}

\end{document}